\documentclass[twoside]{article}

\usepackage[accepted]{aistats2023}

\usepackage[T1]{fontenc} 
\usepackage{amsfonts} 
\usepackage{nicefrac} 
\usepackage{microtype} 
\usepackage{xcolor} 

\usepackage{url} 
\usepackage{lipsum}		 

\usepackage[textsize=tiny]{todonotes}

\usepackage[utf8]{inputenc}
\usepackage{csquotes}
\usepackage[english]{babel}
\usepackage{comment}

\usepackage[
backend=biber,
style=alphabetic,
sorting=ynt
]{biblatex}
\usepackage{color}
\usepackage{graphicx}
\usepackage{algorithm}
\usepackage{algorithmic}
\usepackage{latexsym}
\usepackage{multirow}
\usepackage{amssymb}
\usepackage{amsmath,amssymb}
\usepackage{mathtools}
\usepackage{fancyhdr}
\usepackage{booktabs}
\usepackage{bm}


\usepackage{hyperref}
\hypersetup{
 	colorlinks=true, 
 	bookmarksnumbered=true,
 	pdfborder={0 0 0},
 	citecolor=cyan,
 	linkcolor=red,
 	urlcolor=green,
 	bookmarkstype=toc
}

\usepackage{caption}
\usepackage{subcaption}
\newtheorem{thm}{Theorem}[section]

\newtheorem{lem}{Lemma}[section]

\newtheorem{prob}{Problem}[section]

\newtheorem{assum}{Assumption}[section]

\usepackage{lscape}
\usepackage{booktabs}
\usepackage{wrapfig}
\usepackage{soul}

\definecolor{mygray}{gray}{0.6}

\usepackage{scrwfile}
\TOCclone[\contentsname~(\appendixname)]{toc}{atoc}

\AfterTOCHead[toc]{%
}
\AfterTOCHead[atoc]{%
 \edef\maintocdepth{\the\value{tocdepth}}%
 \value{tocdepth}=-10000\relax%
}

\usepackage{blindtext}
\usepackage[page]{appendix} 

\begin{document}

\twocolumn[

\aistatstitle{Conjugate Gradient Method for Generative Adversarial Networks}
\aistatsauthor{Hiroki Naganuma\And Hideaki Iiduka}
\aistatsaddress{Mila, Université de Montréal \And Meiji University } ]

\begin{abstract}
One of the training strategies of generative models is to minimize the Jensen–Shannon divergence between the model distribution and the data distribution. Since data distribution is unknown, generative adversarial networks (GANs) formulate this problem as a game between two models, a generator and a discriminator. The training can be formulated in the context of game theory and the local Nash equilibrium (LNE). It does not seem feasible to derive guarantees of stability or optimality for the existing methods. This optimization problem is far more challenging than the single objective setting. Here, we use the conjugate gradient method to reliably and efficiently solve the LNE problem in GANs. We give a proof and convergence analysis under mild assumptions showing that the proposed method converges to {a} LNE with three different learning rate update rules, including a constant learning rate. {Finally, we demonstrate that the proposed method outperforms stochastic gradient descent (SGD) and momentum SGD in terms of best Fréchet inception distance (FID) score and outperforms Adam on average.} The code is available at \url{https://github.com/Hiroki11x/ConjugateGradient_GAN}.
\end{abstract}

\begin{table*}[t]
\caption{Convergence rates of our algorithms with constant and diminishing learning rates.}
\label{table:1}
\scalebox{1}{ 
\begin{tabular}{ll|c|c|c}
\bottomrule
\multirow{2}{*}{} & \multirow{2}{*}{} & \multirow{2}{*}{Constant learning rate} & \multirow{2}{*}{Diminishing learning rate} & \multirow{2}{*}{Diminishing learning rate} \\ 
 & & & & \\
 & & ($a_n = a$, $b_n = b$) & ($a_n = n^{-1/2}$, $b_n = n^{-1/2}$) & ($a_n = n^{-\eta_a}$, $b_n = n^{-\eta_b}$) \\ \hline\hline
\multirow{2}{*}{SGD} & Generator & $\mathcal{O}(N^{-1}) + C_1^G a$& $\mathcal{O}(N^{-1/2})$ & $\mathcal{O}(N^{-\min\{\eta_a, 1 - \eta_a\}})$ \\
 & Discriminator & $\mathcal{O}(N^{-1}) + C_1^D b$& $\mathcal{O}(N^{-1/2})$ & $\mathcal{O}(N^{-\min\{\eta_b, 1 - \eta_b\}})$\\ \hline
\multirow{2}{*}{Momentum} & Generator & $\mathcal{O}(N^{-1}) + C_1^G a + C_2^G \beta^G$&$\mathcal{O}(N^{-1/2})$ & $\mathcal{O}(N^{-\min\{\eta_a, 1 - \eta_a\}})$ \\
 & Discriminator & $\mathcal{O}(N^{-1}) + C_1^D b + C_2^D \beta^D$ & $\mathcal{O}(N^{-1/2})$ & $\mathcal{O}(N^{-\min\{\eta_b, 1 - \eta_b\}})$ \\ \hline
\multirow{2}{*}{CG} & Generator & $\mathcal{O}(N^{-1}) + C_1^G a + C_2^G \beta^G$& $\mathcal{O}(N^{-1/2})$ & $\mathcal{O}(N^{-\min\{\eta_a, 1 - \eta_a\}})$ \\
 & Discriminator & $\mathcal{O}(N^{-1}) + C_1^D b + C_2^D \beta^D$ & $\mathcal{O}(N^{-1/2})$ & $\mathcal{O}(N^{-\min\{\eta_b, 1 - \eta_b\}})$ \\ \bottomrule 
\end{tabular}
} 
\\
$C_i^G$ and $C_i^D$ ($i=1,2$) are positive constants independent of learning rates $a$ and $b$ and number of iterations $N$. $\beta^G$ and $\beta^D$ are upper bounds of the CG parameter $\beta_n^G$ and $\beta_n^D$, respectively (see Assumption \ref{const}(C2) for details). $\eta_a$ and $\eta_b$ satisfy $1/2 < \eta_b < \eta_a < 1$. The convergence rate is measured as the expectation of the variational inequality (see, e.g., \eqref{sum_g} and \eqref{sum_d}). See Theorems \ref{thm:1} and \ref{thm:2} for detailed convergence analyses. 

\end{table*}

\section{Introduction}
Generative models that estimate the observed data's probability distribution play important roles in machine learning \cite{hong2019generative}. Generative model training involves minimizing the Jensen–Shannon divergence between the data's density function and the generative model's density function. However, as this is computationally intractable, generative adversarial networks (GANs) \cite{goodfellow2014generative} have been proposed to cast the problem as a two-player game between a discriminator network and a generator network.

Since the advent of GANs in 2014, a number of studies have successfully adapted them to a large variety of applications, such as image recognition \cite{radford2015unsupervised}, style conversion \cite{zhu2017unpaired} \cite{karras2019style}, resolution improvement \cite{brock2018large}, text-to-image conversion \cite{zhang2017stackgan}, image inpainting \cite{pathak2016context}, video frame completion \cite{meyer2015phase}, font generation \cite{hayashi2019glyphgan}, speech \cite{donahue2018adversarial}, text generation \cite{fedus2018maskgan}, and many more. Besides these applications, GANs have also been used to generate data for expanding and augmenting datasets used in medical image segmentation tasks \cite{sandfort2019data}, and GANs-based methods have been adapted for domain generalization \cite{sandfort2019data} to coordinate models across different domains.

One way of training GANs is to solve a Nash equilibrium problem \cite{10.2307/1969529} with two players, a discriminator that minimizes the synthetic-real discrimination error, and a generator that maximizes this error. The focus of this paper is based on two algorithms presented in \cite{Heusel2017} that use two time-scale update rules (TTUR) for finding a local Nash equilibrium (LNE). The first is based on stochastic gradient descent (SGD), while the other relies on adaptive moment estimation (Adam). The two algorithms converge almost surely to stationary LNEs when they use {\em diminishing} learning rates \cite[(A2)]{Heusel2017}, \cite[Theorems 1 and 2]{Heusel2017}. Meanwhile, numerical results \cite{Heusel2017} have shown that the two algorithms also perform well when they use {\em constant} learning rates. 

\subsection{Motivation}
Our main motivations are related to the results in \cite{Heusel2017}.

\begin{itemize}
\item We first seek to theoretically verify whether optimization algorithms based on TTUR with {\em constant} learning rates can be applied to LNE problems in GANs. Such a verification would help to bridge the gap between theory and practice regarding the use of TTUR \cite{Heusel2017} in optimization.

\item The second motivation is to show whether {\em conjugate gradient} (CG)-type algorithms, which use CG directions to search for the minima of the observed loss functions, can be applied to LNE problems in theory and practice. 
\end{itemize}

Generally, SGD and its variants generate search directions by using the gradients of loss functions at the current approximation point. One method to accelerate the steepest descent method is the conjugate gradient (CG) method \cite[Chapter 5]{NoceWrig06}, \cite{Hager2006}. The CG direction is defined by the current gradient and the past search direction (see Section \ref{cgm} for the definition of the CG method). CG-type algorithms were studied to solve minimization problems in neural networks \cite{MOLLER1993525,le2011optimization}. Since the LNE problems considered in this paper are more complicated than the minimization problems in \cite{MOLLER1993525,le2011optimization}, we propose to investigate their use for solving LNEs, which is a more challenging setting. We aim to determine whether CG-type algorithms favorably compare to state-of-the-art methods for finding stationary LNEs in GANs.

\begin{table*}[htbp]
\begin{center}
\caption{Notation List}
\scalebox{0.9}{ 
\begin{tabular}{c||l} \hline
Notation & Description\\ \hline
$\mathbb{N}$ & 
The set of all positive integers and zero \\
$[N]$ & 
$[N] := \{1,2,\ldots,N\}$ ($N \in \mathbb{N}\backslash \{ 0 \}$)\\

$|A|$ &
The number of elements of a set $A$ \\

$\mathbb{R}^d$ & 
A $d$-dimensional Euclidean space with inner product $\langle \cdot, \cdot \rangle$, which induces the norm $\| \cdot \|$ \\
$\mathbb{R}_+^d$ & 
$\mathbb{R}_+^d := \{ \bm{x} \in \mathbb{R}^d \colon x_i \geq 0 \text{ } (i\in [d]) \}$ \\
$\mathbb{R}_{++}^d$ & 
$\mathbb{R}_{++}^d := \{ \bm{x} \in \mathbb{R}^d \colon x_i > 0 \text{ } (i\in [d]) \}$ \\
$\mathbb{E}[X]$ & 
The expectation of a random variable $X$ \\
$\mathcal{R}$ &
The set of real-world samples $\bm{x}^{(i)}$ \\
$\mathcal{S}$ &
The set of synthetic samples $\bm{z}^{(i)}$ \\
$\mathcal{R}_n$ &
Mini-batch of $m$ real-world samples $\bm{x}^{(i)}$ at time $n$, obtained from the training set, which is shuffled and aligned. \\
$\mathcal{S}_n$ &
Mini-batch of $m$ synthetic samples $\bm{z}^{(i)}$ at time $n$, obtained from the training set, which is shuffled and aligned. \\
\hline 
$\mathcal{L}_D^{(i)}(\bm{\theta}, \cdot)$ &
A loss function of discriminator for a fixed $\bm{\theta} \in \mathbb{R}^{\Theta}$ and real-world sample $\bm{x}^{(i)}$ \\
$\mathcal{L}_D(\bm{\theta}, \cdot)$ &
A loss function of discriminator for a fixed $\bm{\theta} \in \mathbb{R}^{\Theta}$, i.e., 
$\mathcal{L}_D(\bm{\theta}, \cdot) := |\mathcal{R}|^{-1}\sum_{i\in \mathcal{R}} \mathcal{L}_D^{(i)}(\bm{\theta}, \cdot)$ \\
$\mathcal{D}_{\bm{\theta}}(\bm{w})$ &
The gradient of $\mathcal{L}_D^{(i)}(\bm{\theta}, \cdot)$
for mini-batch $\mathcal{R}_n$, i.e., 
$\mathcal{D}_{\bm{\theta}}(\bm{w}) := \sum_{i \in \mathcal{R}_n} \nabla_{\bm{w}} \mathcal{L}_D^{(i)}(\bm{\theta}, \bm{w})$
\\
$\mathcal{D}(\bm{\theta}, \bm{w})$ &
The stochastic gradient of $\mathcal{L}_D^{(i)}(\bm{\theta}, \cdot)$
for mini-batch $\mathcal{R}_n$, i.e., $\mathcal{D}(\bm{\theta}, \bm{w}) := m^{-1} \mathcal{D}_{\bm{\theta}}(\bm{w})$
\\ \hline
$\mathcal{L}_G^{(i)}(\cdot, \bm{w})$ &
A loss function of generator for a fixed $\bm{w} \in \mathbb{R}^{W}$ and synthetic sample $\bm{z}^{(i)}$ \\
$\mathcal{L}_G(\cdot, \bm{w})$ &
A loss function of generator for a fixed $\bm{w} \in \mathbb{R}^{W}$, i.e., 
$\mathcal{L}_G(\cdot, \bm{w}) := |\mathcal{S}|^{-1} \sum_{i\in \mathcal{S}} \mathcal{L}_G^{(i)}(\cdot, \bm{w})$ \\
$\mathcal{G}_{\bm{w}}(\bm{\theta})$ &
The gradient of $\mathcal{L}_G^{(i)}(\cdot, \bm{w})$
for mini-batch $\mathcal{S}_n$, i.e., 
$\mathcal{G}_{\bm{w}}(\bm{\theta}) := \sum_{i \in \mathcal{S}_n} \nabla_{\bm{\theta}} \mathcal{L}_G^{(i)}(\bm{\theta}, \bm{w})$
\\
$\mathcal{G}(\bm{\theta}, \bm{w})$ &
The stochastic gradient of $\mathcal{L}_G^{(i)}(\cdot, \bm{w})$
for mini-batch $\mathcal{S}_n$ as unbiased estimation, i.e., $\mathcal{G}(\bm{\theta}, \bm{w}) := m^{-1} \mathcal{G}_{\bm{w}}(\bm{\theta})$
\\ \hline
$\mathrm{LNE}(\mathcal{L}_D, \mathcal{L}_G)$ &
The set of stationary LNEs for Nash equilibrium problem for 
$\mathcal{L}_D$ and $\mathcal{L}_G$
\\ \hline
\end{tabular}\label{notation}
} 
\end{center}
\end{table*}

\subsection{Contributions of this paper} 
Our contributions are as follows:

\begin{itemize}
\item We propose a CG-type algorithm (Algorithm \ref{algo:1}) for solving LNE problems in GANs. Our method leverages efficient CG parameters, such as the Fletcher--Reeves (FR) \cite{Fletcher1964}, Polak--Ribi\`ere--Polyak (PRP) \cite{M2AN_1969__3_1_35_0,POLYAK196994}, Hestenes--Stiefel (HS) \cite{Hestenes1952MethodsOC}, and Dai--Yuan (DY) \cite{doi:10.1137/S1052623497318992} formulas, to generate the CG direction.

\item We provide a convergence guarantee and convergence rate analyses of the proposed algorithm with a constant learning rate rule (Theorem \ref{thm:1}) and a diminishing learning rate rule (Theorem \ref{thm:2}), which we summarize in Table \ref{table:1}.

\item We provide an extensive empirical study on the convergence to a LNE for SGD, momentum SGD, and CG-type algorithms. Notably, we observe that our CG-type algorithm outperforms SGD, momentum SGD, and Adam for the training of GANs over an extensive hyperparameter search range (Figure \ref{fig:fig2}, Table \ref{table:exp-fid-const}).

\end{itemize}

The proposed algorithm uses the CG direction determined from both the current search direction and past search direction, in contrast to the SGD-type and Adam-type algorithms in \cite{Heusel2017} that use only stochastic gradient directions. Thanks to the previous results \cite{NoceWrig06}, \cite{Hager2006}, CG methods can quickly solve non-convex smooth optimization problems. Hence, we expect the proposed algorithm to perform better than those using stochastic gradient directions. We should also note that the proposed algorithm includes the SGD-type and momentum-type algorithms (see Table \ref{table:1}). 

We would like to emphasize that the main theoretical contribution is showing convergence as well as the convergence rate of the proposed algorithm with {\em constant} learning rates (Theorem \ref{thm:1} and Table \ref{table:1}), which is in contrast to the previous results for algorithms with diminishing learning rates (see \cite{Heusel2017} for training GANs and \cite{DBLP:journals/corr/KingmaB14,DBLP:conf/iclr/ReddiKK18} for training deep neural networks). The results indicate that the proposed algorithm using a small constant learning rate achieves approximately an $\mathcal{O}(N^{-1})$ convergence rate (Table \ref{table:1} and \eqref{n}), where $N$ denotes the number of iterations. This implies that optimization algorithms with constant learning rates perform well, as evidenced in \cite{Heusel2017}.

We also analyze the convergence and the convergence rate of the proposed algorithm with {\em diminishing} learning rates (Theorem \ref{thm:2} and Table \ref{table:1}). The results indicate that the proposed algorithm using diminishing learning rates $a_n = b_n = n^{-1/2}$ achieves an $\mathcal{O}(N^{-1/2})$ convergence rate (Table \ref{table:1} and \eqref{improve}). To guarantee that the algorithm converges almost surely to a stationary LNE, we need to use diminishing learning rates $a_n = n^{-\eta_a}$ and $b_n = n^{-\eta_b}$, where $1/2 < \eta_b < \eta_a < 1$. Accordingly, the algorithm for the generator achieves an $\mathcal{O}(N^{-\min\{\eta_a, 1 - \eta_a\}})$ convergence rate, while the one for the discriminator achieves an $\mathcal{O}(N^{-\min\{\eta_b, 1 - \eta_b\}})$ convergence rate (Table \ref{table:1} and \eqref{rate}). {See Theorem \ref{thm:2}(ii) for other convergence analyses of the proposed algorithm.}

\section{Mathematical Preliminaries}
\label{sec:mp}
\subsection{Problem formulation}
\begin{assum}\label{assum:1}
We assume the following conditions \cite[(A1)]{Heusel2017} (The notation used in this paper is summarized in Table \ref{notation}):

\begin{enumerate}
\item[{\em (A1)}] $\mathcal{L}_D^{(i)} \colon \mathbb{R}^\Theta \times \mathbb{R}^W \to \mathbb{R}$ ($i\in \mathcal{R}$) and $\mathcal{L}_G^{(i)}\colon \mathbb{R}^\Theta \times \mathbb{R}^W \to \mathbb{R}$ ($i\in \mathcal{S}$) are continuously differentiable;
\item[{\em (A2)}] For $\bm{w} \in \mathbb{R}^W$, $\mathcal{G}_{\bm{w}} \colon \mathbb{R}^\Theta \to \mathbb{R}^\Theta$ is $L_1$-Lipschitz continuous.\footnote{$A \colon \mathbb{R}^d \to \mathbb{R}^d$ is said to be Lipschitz continuous with a Lipschitz constant $L$ ($L$-Lipschitz continuous) if $\| A(\bm{x}) - A(\bm{y})\| \leq L \|\bm{x} - \bm{y} \|$ for all $\bm{x},\bm{y} \in \mathbb{R}^d$.} For $\bm{\theta} \in \mathbb{R}^\Theta$, $\mathcal{D}_{\bm{\theta}} \colon \mathbb{R}^W \to \mathbb{R}^W$ is $L_2$-Lipschitz continuous.
\end{enumerate}

\end{assum}

This paper considers the following LNE problem with two players, a discriminator and a generator \cite{Heusel2017}: 

\begin{prob}\label{prob:1}
Find a pair $(\bm{\theta}^*, \bm{w}^*) \in \mathbb{R}^\Theta \times \mathbb{R}^W$ satisfying 
\begin{align}\label{lne}
\nabla_{\bm{w}} \mathcal{L}_D (\bm{\theta}^*, \bm{w}^*) = \bm{0} 
\text{ and }
\nabla_{\bm{\theta}} \mathcal{L}_G (\bm{\theta}^*, \bm{w}^*) = \bm{0}.
\end{align}
\end{prob}

A Nash equilibrium $(\bm{\theta}^*, \bm{w}^*) \in \mathbb{R}^\Theta \times \mathbb{R}^W$ \cite{10.2307/1969529} defined by 
\begin{equation}
\begin{split}\label{lne2}
\mathcal{L}_D (\bm{\theta}^*, \bm{w}^*) \leq \mathcal{L}_D (\bm{\theta}^*, \bm{w})
\text{ for all } \bm{w} \in \mathbb{R}^W
\\
\text{ and }
\mathcal{L}_G (\bm{\theta}^*, \bm{w}^*) \leq \mathcal{L}_G (\bm{\theta}, \bm{w}^*)
\text{ for all } \bm{\theta} \in \mathbb{R}^\Theta
\end{split}
\end{equation}
satisfies \eqref{lne}. Here, $(\bm{\theta}^*, \bm{w}^*) \in \mathbb{R}^\Theta \times \mathbb{R}^W$ {satisfying \eqref{lne} is called a {\em stationary LNE}.} 

\subsection{Conjugate gradient methods}
\label{cgm}
Let us consider a stationary point problem associated with unconstrained nonconvex optimization, 
\begin{align}\label{nonconvex}
\text{find a point } \bm{x}^* \in \mathbb{R}^d \text{ such that } 
\nabla f (\bm{x}^*) = \bm{0}, 
\end{align}
where $f \colon \mathbb{R}^d \to \mathbb{R}$ is continuously differentiable. There are many optimization methods \cite[Chapters 3, 5, and 6]{NoceWrig06} for solving problem \eqref{nonconvex}, such as the steepest descent method, Newton method, quasi-Newton methods, and CG method. The CG method \cite[Chapter 5]{NoceWrig06}, \cite{Hager2006} is defined as follows: given $\bm{x}_0 \in \mathbb{R}^d$ and $\bm{d}_0 := - \nabla f (\bm{x}_0)$, 
\begin{align}\label{cg}
\bm{x}_{n+1} := \bm{x}_n + \alpha_n \bm{d}_n,
\text{ }
\bm{d}_{n+1} := - \nabla f (\bm{x}_{n+1}) + \beta_{n+1} \bm{d}_n,
\end{align}
where $(\alpha_n)_{n\in\mathbb{N}}$ is the sequence of step sizes (referred to as learning rates in the machine learning field), $\beta_{n+1} \in \mathbb{R}_+$, and $\bm{d}_n$ denotes the search direction called the CG direction. The CG direction $\bm{d}_{n+1}$ at time $n+1$ is computed from not only the current gradient $\nabla f(\bm{x}_{n+1})$ but also the past direction $\bm{d}_n$. Since algorithm \eqref{cg} does not use any inverses of matrices, the method requires little memory. Overall, CG methods are divided into the linear kind and the nonlinear kind. 

The linear kind can solve a linear system of equations $A \bm{x} = \bm{b}$ with a positive-definite matrix $A$ and $\bm{b} \in \mathbb{R}^d$, which is equivalent to minimizing $f(\bm{x}) = (1/2) \langle \bm{x}, A \bm{x} \rangle - \langle \bm{b}, \bm{x} \rangle$ over $\mathbb{R}^d$. When the eigenvalues of $A$ consist of $m$ large values, with $d-m$ smaller eigenvalues, the linear method will terminate at a solution after only $m+1$ steps \cite[Chapter 5.1]{NoceWrig06}.

The nonlinear kind has been widely studied (see also the following parameters $\beta_n$ used in \eqref{cg}) and has proved to be quite successful in practice \cite[Chapter 5.2]{NoceWrig06}. This implies that the nonlinear kind can be applied to a large-scale stationary point problem \eqref{nonconvex} with a general nonlinear function $f$.

Well-known parameters $\beta_{n}$ for the nonlinear CG method \eqref{cg} include the FR \cite{Fletcher1964}, PRP \cite{M2AN_1969__3_1_35_0,POLYAK196994}, HS \cite{Hestenes1952MethodsOC}, and DY \cite{doi:10.1137/S1052623497318992} formulas defined as follows: 
\begin{align*}
&\beta_n^{\mathrm{FR}} = \frac{\| \nabla f(\bm{x}_n) \|^2}{\| \nabla f(\bm{x}_{n-1}) \|^2},
\text{ }\\
&\beta_n^{\mathrm{PRP}} = \frac{\langle \nabla f (\bm{x}_n), \nabla f(\bm{x}_n)-\nabla f(\bm{x}_{n-1}) \rangle}{\| \nabla f(\bm{x}_{n-1}) \|^2},\\
&\beta_n^{\mathrm{HS}} = \frac{\langle \nabla f (\bm{x}_n), \nabla f(\bm{x}_n)-\nabla f(\bm{x}_{n-1}) \rangle}{\langle \bm{d}_{n-1}, \nabla f(\bm{x}_n)-\nabla f(\bm{x}_{n-1}) \rangle}, 
\text{ } \\
&\beta_n^{\mathrm{DY}} = \frac{\| \nabla f(\bm{x}_n) \|^2}{\langle \bm{d}_{n-1}, \nabla f(\bm{x}_n)-\nabla f(\bm{x}_{n-1}) \rangle}.
\end{align*}
The Hager--Zhang (HZ) \cite{Hager2005} formula is a modification of the HS formula; it is defined as follows: 
\begin{align*}
 \beta_n^{\mathrm{HZ}} = \beta_n^{\mathrm{HS}} - \mu \frac{\|\nabla f(\bm{x}_n)-\nabla f(\bm{x}_{n-1})\|^2 \langle \nabla f(\bm{x}_n), \bm{d}_{n-1} \rangle}{\langle \bm{d}_{n-1}, \nabla f(\bm{x}_n)-\nabla f(\bm{x}_{n-1}) \rangle^2}, 
\end{align*}
where $\mu > 1/4$. The hybrid conjugate gradient method \cite{Dai_Yuan_2001} combining the HS and DY methods uses 
\begin{align*}
 \beta_n = \max \left\{ 0, \min \left\{ \beta_n^{\mathrm{HS}}, \beta_n^{\mathrm{DY}} \right\} \right\},
\end{align*}
while the hybrid conjugate gradient method \cite{Hu_Storey_1991} combining the FR and PRP methods uses
\begin{align*}
 \beta_n = \max \left\{ 0, \min \left\{ \beta_n^{\mathrm{FR}}, \beta_n^{\mathrm{PRP}} \right\} \right\}.
\end{align*}
The global convergence and convergence rate of the nonlinear CG methods with the above parameters $\beta_n$ are described in \cite[Chapter 5.2]{NoceWrig06}, \cite{Hu_Storey_1991}, \cite{Dai_Yuan_2001}, and \cite{Hager2005}.

\subsection{Relationship and difference between conjugate gradient methods and momentum method}
The momentum method (momentum SGD) \cite[(9)]{polyak1964}, \cite[Section 2]{sut2013} is defined as follows: given $\bm{x}_0 \in \mathbb{R}^d$ and $\bm{m}_{-1} = \bm{0}$,
\begin{align}\label{momentum}
{\bm{m}_n = \nabla f(\bm{x}_n) + \mu \bm{m}_{n-1},}
\text{ }
{\bm{x}_{n+1} = \bm{x}_n - \epsilon \bm{m}_n,}
\end{align}
where $\epsilon > 0$ is the learning rate and $\mu \in [0,1]$ is the momentum coefficient. The momentum method \eqref{momentum} generates a sequence defined by 
\begin{align*}
{\bm{x}_{n+1} := \bm{x}_n - \epsilon \nabla f (\bm{x}_n) - \epsilon \mu \bm{m}_{n-1},}
\end{align*}
while the CG method \eqref{cg} generates a sequence defined by 
\begin{align*}
\bm{x}_{n+1} := \bm{x}_n - \alpha_n \nabla f (\bm{x}_{n}) + \alpha_n \beta_{n} \bm{d}_{n-1},
\end{align*}
where $\bm{d}_{-1} = \bm{0}$ {and $\beta_n \in \mathbb{R}_+$}. Accordingly, the CG method \eqref{cg} is a momentum method with a learning rate $\alpha_n$ and momentum coefficient ${-\beta_{n}}$. While the momentum method \eqref{momentum} uses the momentum coefficient $\mu$, the CG method \eqref{cg} uses a {negative} momentum coefficient ${-\beta_{n}}$ dependent of $n$ through the CG parameters $\beta_n$ listed in Subsection \ref{cgm}. The difference in these momentum coefficients can significantly change the learning dynamics with the CG method and momentum method \cite{sun2021training}.

\section{Conjugate Gradient Method for Local Nash Equilibrium Problem}
Notably, the training in the recent studies on GANs (which addresses the LNE problem) has been conducted by setting the momentum to zero \cite{brock2018large,karras2019style} or a negative value \cite{gidel2019negative}. This trend motivates the use of CG methods that do not use a fixed positive constant as the momentum coefficient.

\label{sec:cgm}
The following algorithm for solving Problem \ref{prob:1} is based on the CG method. Algorithm \ref{algo:1} with $\beta_n^D = \beta_n^G = 0$ coincides with the SGD type of algorithm \cite[(1)]{Heusel2017}, i.e., 
\begin{align*}
 &\bm{w}_{n+1} = \bm{w}_n - b_n \mathcal{D}(\bm{\theta}_n, \bm{w}_n),\\ 
 &\bm{\theta}_{n+1} = \bm{\theta}_n - a_n \mathcal{G}(\bm{\theta}_n, \bm{w}_n).
\end{align*}

\begin{algorithm} 
\caption{Conjugate Gradient Method for Problem \ref{prob:1}} 

\label{algo:1} 
\begin{algorithmic}[1] 
\REQUIRE
$(a_n), (b_n) \subset \mathbb{R}_{++}$, $(\beta_n^{D}), (\beta_n^{G}) \subset \mathbb{R}_+$
\STATE
$(\bm{\theta}_0, \bm{w}_0) \in \mathbb{R}^\Theta \times \mathbb{R}^W$, 
$\bm{d}_{-1}^{D} \in \mathbb{R}^W$, 
$\bm{d}_{-1}^{G} \in \mathbb{R}^\Theta$
\FOR{$n = 0$, $n \gets n+1$} 
\STATE 
$\bm{d}_n^{D} := - \mathcal{D}(\bm{\theta}_n, \bm{w}_n) + \beta_n^{D} \bm{d}_{n-1}^{D}$
\STATE 
$\bm{w}_{n+1} := \bm{w}_n + b_n \bm{d}_n^{D}$
\STATE 
$\bm{d}_n^{G} := - \mathcal{G}(\bm{\theta}_n, \bm{w}_n) + \beta_n^{G} \bm{d}_{n-1}^{G}$
\STATE 
$\bm{\theta}_{n+1} := \bm{\theta}_n + a_n \bm{d}_n^{G}$
\ENDFOR 
\end{algorithmic}
\end{algorithm}

\subsection{Constant learning rate rule}
\label{subsec:constant}
\begin{assum}\label{const}
\text{ } 
\begin{enumerate}
\item[{\em (C1)}]
$a_n := a \in \mathbb{R}_{++}$ and $b_n := b \in \mathbb{R}_{++}$ for all $n\in \mathbb{N}$.
\item[{\em (C2)}]
$\beta^{D} := \sup \{\beta_n^{D} \colon n\in\mathbb{N}\} \in [0,1/2]$, 
$\beta^{G} :=\sup \{ \beta_n^{G} \colon n\in\mathbb{N}\} \in [0,1/2]$.
\item[{\em {(C3)}}] $(\bm{\theta}_n)_{n\in\mathbb{N}}$ and $(\bm{w}_n)_{n\in\mathbb{N}}$ are almost surely bounded.\footnote{The sequence $(\bm{x}_n)_{n\in\mathbb{N}} \subset \mathbb{R}^d$ is said to be almost surely bounded if $\sup\{ \|\bm{x}_n\| \colon n\in\mathbb{N} \} < + \infty$ holds almost surely \cite[(2.1.4)]{Borkar08}.}
\end{enumerate}
\end{assum}
Assumption (C2) is used to prove the boundedness of $(\mathbb{E}[\|\bm{d}_n^D\|])_{n\in\mathbb{N}}$ and $(\mathbb{E}[\|\bm{d}_n^G\|])_{n\in\mathbb{N}}$ (see Lemmas \ref{lem:0} and \ref{lem:0_1}).\footnote{By referring to the proofs of Lemmas \ref{lem:0} and \ref{lem:0_1}, we can check that (*) ensures the existence of $n_0 \in\mathbb{N}$ such that, for all $n \geq n_0$, $\beta_n^{D},\beta_n^{G} \leq 1/2$ implies the boundedness of $(\mathbb{E}[\|\bm{d}_n^D\|])_{n\in\mathbb{N}}$ and $(\mathbb{E}[\|\bm{d}_n^G\|])_{n\in\mathbb{N}}$. This in turn implies that it is sufficient to assume the weaker condition (*) than (C2). In this paper, we use (C2) as a hypothesis for simplicity.} If $(\mathbb{E}[\|\bm{d}_n^D\|])_{n\in\mathbb{N}}$ and $(\mathbb{E}[\|\bm{d}_n^G\|])_{n\in\mathbb{N}}$ are bounded, then (C2) can be omitted. Assumption {(C3)} is the same as (A5) in \cite[(A5)]{Heusel2017}.

The following presents the convergence analysisand convergence rate analysis of Algorithm \ref{algo:1} with constant learning rates. Section \ref{subsub:3.1.1} illustrates some examples of Theorem \ref{thm:1}. 
 
\begin{thm}\label{thm:1}
Suppose that Assumptions \ref{assum:1}(A1)--(A2) and \ref{const}(C1)--{(C3)} hold and let $C_i$ and $\tilde{K}_i$ ($i=1,2$) be positive constants independent of $n$ (see Appendix for the definitions of the constants). Then, the following hold:

{\em (i)} {\em [Convergence]} For all $\bm{\theta} \in \mathbb{R}^\Theta$,
\begin{equation}
\begin{split}\label{liminf:1}
\liminf_{n \to + \infty} \mathbb{E}\left[\left\langle \bm{\theta}_n - \bm{\theta},
\nabla_{\bm{\theta}} \mathcal{L}_G(\bm{\theta}_n, \bm{w}_n) \right\rangle \right]
\\
\leq 2 \tilde{K}_1^2 a + 2 C_1 \tilde{K}_1 \beta^G.
\end{split}
\end{equation}
In particular, there exists a subsequence $((\bm{\theta}_{n_i}, \bm{w}_{n_i}))_{i\in \mathbb{N}}$ of $((\bm{\theta}_{n}, \bm{w}_n))_{n\in\mathbb{N}}$ such that $((\bm{\theta}_{n_{i}}, \bm{w}_{n_{i}}))_{i\in \mathbb{N}}$ converges almost surely to $(\bm{\theta}^\star, \bm{w}^\star)$ satisfying
\begin{align}\label{liminf:1_1}
\mathbb{E}\left[ \left\|\nabla_{\bm{\theta}} \mathcal{L}_G(\bm{\theta}^\star, \bm{w}^\star) \right\|^2 \right] 
\leq 
2 \tilde{K}_1^2 a + 2 C_1 \tilde{K}_1 \beta^G.
\end{align} 
For all $\bm{w} \in \mathbb{R}^W$,
\begin{equation}
\begin{split}\label{liminf:2}
\liminf_{n \to + \infty} \mathbb{E}\left[\left\langle \bm{w}_n - \bm{w},
\nabla_{\bm{w}} \mathcal{L}_D(\bm{\theta}_n, \bm{w}_n) \right\rangle \right]\\
\leq 2 \tilde{K}_2^2 b + 2 C_2 \tilde{K}_2 \beta^D.
\end{split}
\end{equation}
In particular, there exists a subsequence $((\bm{\theta}_{n_j}, \bm{w}_{n_j}))_{j\in \mathbb{N}}$ of $((\bm{\theta}_{n}, \bm{w}_n))_{n\in\mathbb{N}}$ such that $((\bm{\theta}_{n_{j}}, \bm{w}_{n_{j}}))_{j\in \mathbb{N}}$ converges almost surely to $(\bm{\theta}_\star, \bm{w}_\star)$ satisfying
\begin{align}\label{liminf:2_1}
\mathbb{E}\left[ \left\|\nabla_{\bm{w}} \mathcal{L}_D(\bm{\theta}_\star, \bm{w}_\star) \right\|^2 \right] 
\leq 
2 \tilde{K}_2^2 b + 2 C_2 \tilde{K}_2 \beta^D.
\end{align} 
{\em (ii)} {\em [Convergence Rate]} For all $\bm{\theta} \in \mathbb{R}^\Theta$ and all $N \geq 1$,
\begin{equation}
\begin{split}\label{sum_g}
&\frac{1}{N} \sum_{n\in [N]} \mathbb{E}\left[\left\langle \bm{\theta}_n - \bm{\theta}, 
\nabla_{\bm{\theta}} \mathcal{L}_G(\bm{\theta}_n, \bm{w}_n) \right\rangle \right]\\
&\leq
\frac{\mathbb{E} [ \| \bm{\theta}_1 - \bm{\theta}\|^2]}{2a N} 
+ 
2 \tilde{K}_1^2 a
+ 2 C_1 \tilde{K}_1 \beta^{G}.
\end{split}
\end{equation}
For all $\bm{w} \in \mathbb{R}^W$ and all $N \geq 1$,
\begin{equation}
\begin{split}\label{sum_d}
&\frac{1}{N} \sum_{n\in [N]} \mathbb{E}\left[\left\langle \bm{w}_n - \bm{w}, 
\nabla_{\bm{w}} \mathcal{L}_D(\bm{\theta}_n, \bm{w}_n) \right\rangle \right]\\
&\leq
\frac{\mathbb{E} [ \| \bm{w}_1 - \bm{w}\|^2]}{2b N} 
+ 
2\tilde{K}_2^2 b
+ 2 C_2 \tilde{K}_2 \beta^{D}.
\end{split}
\end{equation}
If $((\bm{\theta}_n,\bm{w}_n))_{n\in\mathbb{N}}$ converges almost surely to $(\bm{\theta}^\star, \bm{w}^\star)$,\footnote{For example, the uniqueness of an accumulation point of $((\bm{\theta}_n,\bm{w}_n))_{n\in\mathbb{N}}$ implies that this condition holds.} then the convergent point $(\bm{\theta}^\star, \bm{w}^\star)$ approximates a stationary LNE in the sense that 
\begin{align}\label{ane}
\begin{split}
&\mathbb{E}\left[ \left\|
\nabla_{\bm{\theta}} \mathcal{L}_G(\bm{\theta}^\star, \bm{w}^\star) 
\right\|^2 \right] 
\leq 
2 \tilde{K}_1^2 a + 2 C_1 \tilde{K}_1 \beta^G,\\
&\mathbb{E}\left[ \left\|
\nabla_{\bm{w}} \mathcal{L}_D(\bm{\theta}^\star, \bm{w}^\star) 
\right\|^2 \right] 
\leq 
2 \tilde{K}_2^2 b + 2 C_2 \tilde{K}_2 \beta^D
\end{split}
\end{align} 
with convergence rates \eqref{sum_g} and \eqref{sum_d}.
\end{thm}

\begin{figure*}
 \centering
 \includegraphics[width=0.42\linewidth]{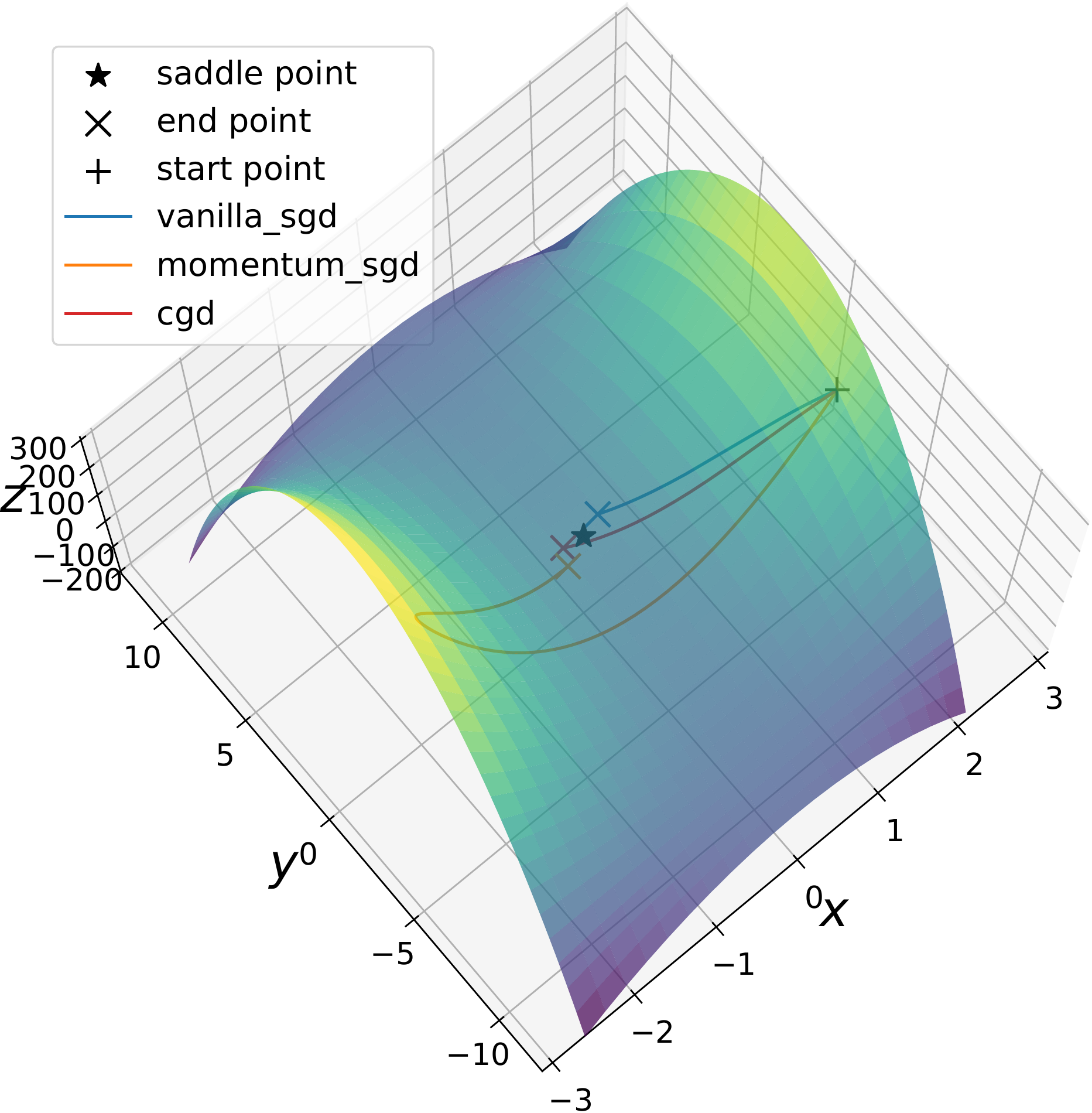}
 \includegraphics[width=0.42\linewidth]{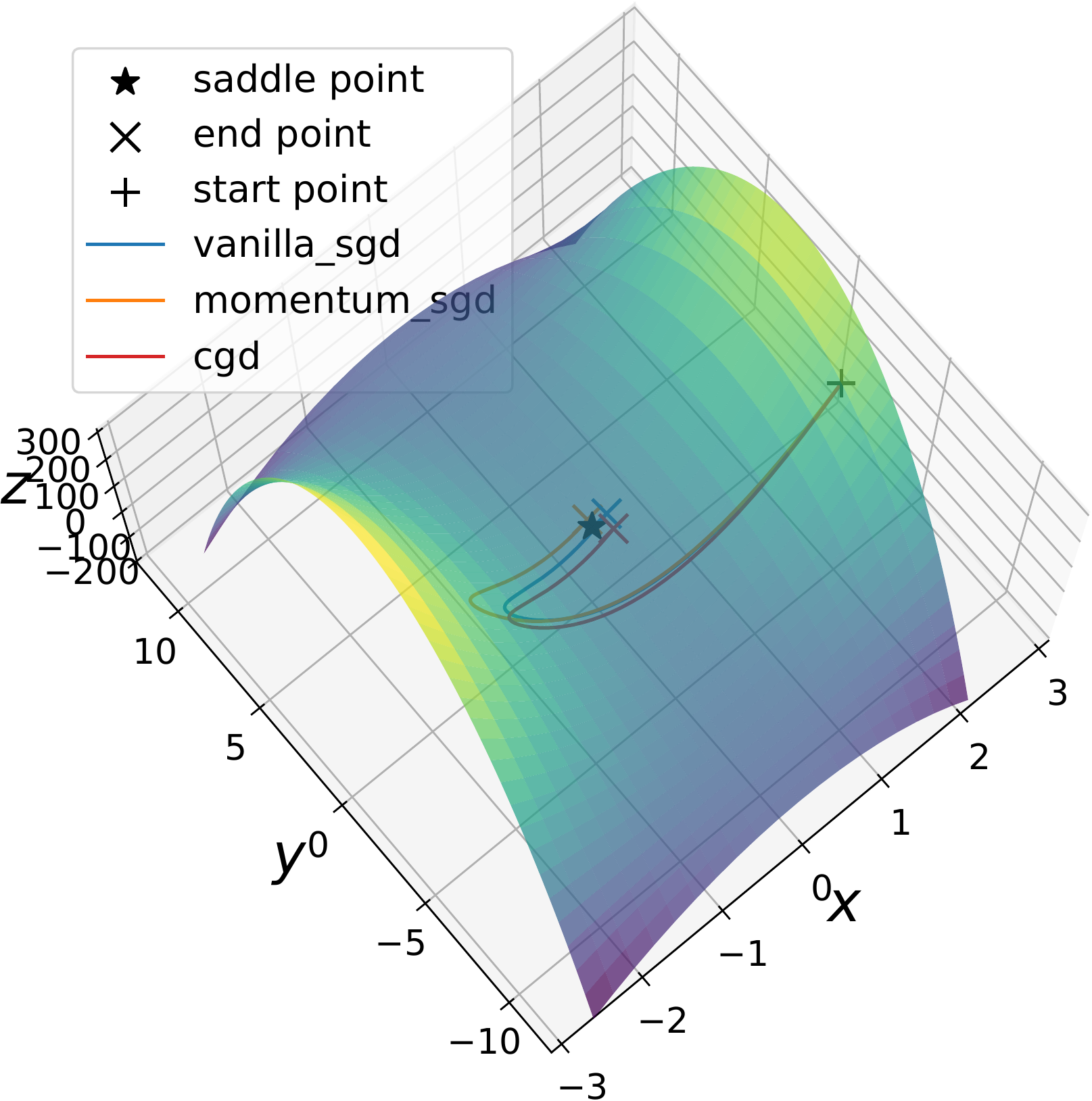}
 \includegraphics[width=0.48\linewidth]{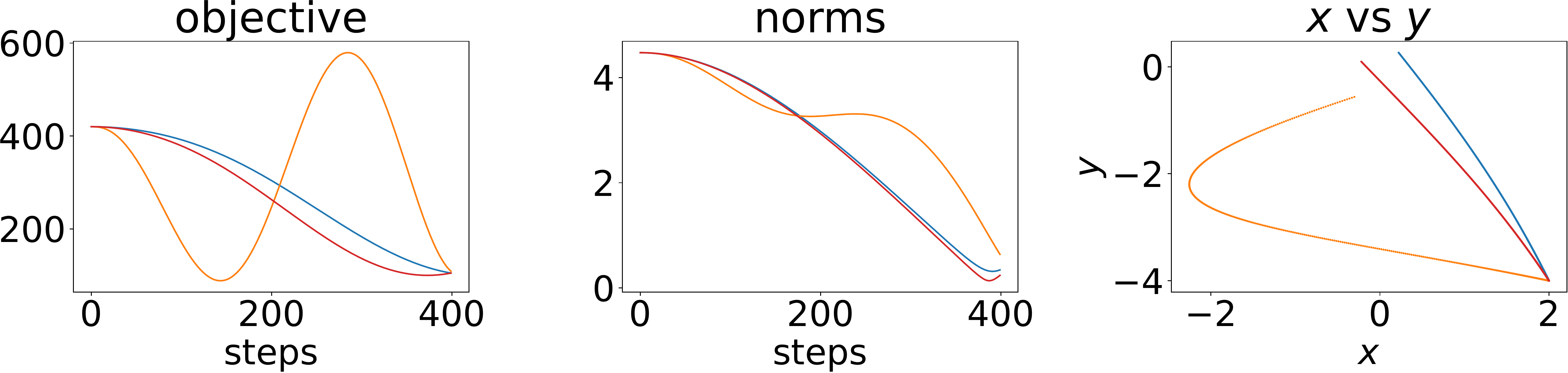}
 \includegraphics[width=0.48\linewidth]{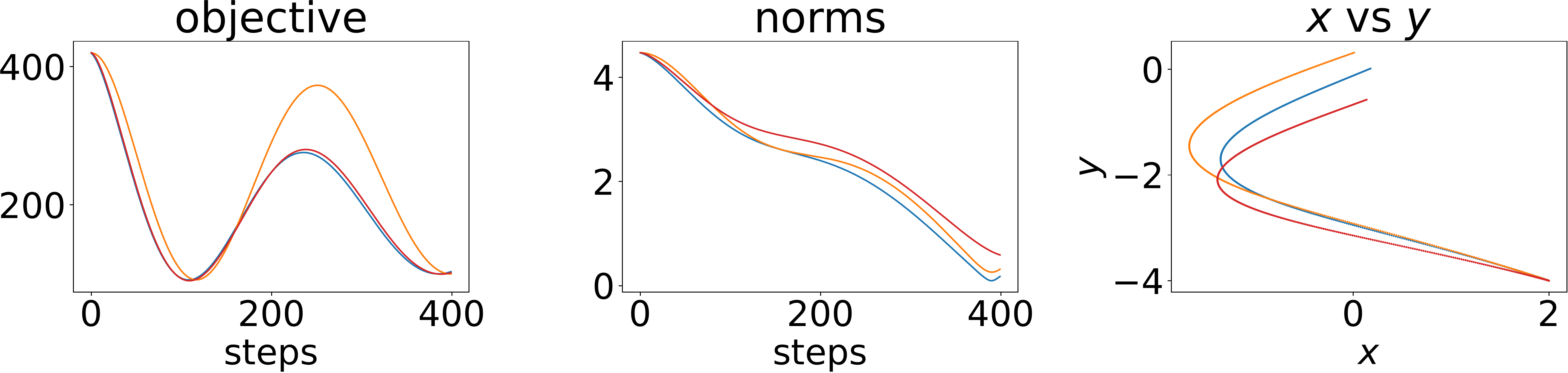}
 \vspace{-2mm}
\caption{ Toy example of minimax optimization. Objective function: $f(x,y) = (1+x^2) \cdot (100-y^2)$. \textbf{Top Left}: trajectory with a {\em constant} learning rate, \textbf{Top Right}: trajectory with a {\em diminishing} learning rate, \textbf{Bottom Left}: objective, norms, and x vs y with a {\em constant} learning rate schedule, \textbf{Bottom Right}: same as the bottom left but with a {\em diminishing} learning rate schedule. We tuned the initial learning rate (See Appendix \ref{appendix:hyperparameters}). }
\label{fig:toy-exp1}
\vspace{-5mm}
\end{figure*}

\subsection{Diminishing learning rate rule}
\label{subsec:diminishing}
\begin{assum}\label{dim}
\text{ } 

\begin{enumerate}
\item[{\em (D1)}] For each $\bm{\theta} \in \mathbb{R}^\Theta$, the ordinary differential equation $\dot{\bm{w}}(t) = \nabla_{\bm{w}} \mathcal{L}_D (\bm{\theta}, \bm{w}(t))$ has a local asymptotically stable attractor $\lambda(\bm{\theta})$ within a domain of attraction such that $\lambda \colon \mathbb{R}^\Theta \to \mathbb{R}^W$ is Lipschitz continuous. The ordinary differential equation $\dot{\bm{\theta}}(t) = \nabla_{\bm{\theta}} \mathcal{L}_G (\bm{\theta}(t), \lambda(\bm{\theta}(t)))$ has a local asymptotically stable attractor $\bm{\theta}^*$ within a domain of attraction.
\item[{\em {(D2)}}] $(a_n)_{n\in\mathbb{N}}$ and $(b_n)_{n\in\mathbb{N}}$ are monotone decreasing sequences satisfying either {\em (i)} or {\em (ii)}: 
\begin{enumerate}
\item[{\em (i)}]
$\displaystyle{\sum_{n=0}^{+\infty} a_n = + \infty}$, 
$\displaystyle{\sum_{n=0}^{+\infty} a_n^2 < + \infty}$,
$\displaystyle{\sum_{n=0}^{+\infty} b_n = + \infty}$, 
$\displaystyle{\sum_{n=0}^{+\infty} b_n^2 < + \infty}$,
and 
$a_n = o (b_n)$;
\item[{\em (ii)}]
$\displaystyle{\lim_{n \to + \infty} (n a_n)^{-1} = \lim_{n \to + \infty} (n b_n)^{-1} = 0}$
and 
$\displaystyle{\lim_{n \to + \infty} n^{-1} \sum_{k=0}^n a_k = \lim_{n \to + \infty} n^{-1} \sum_{k=0}^n b_k = 0}$.
\end{enumerate}
\item[{\em {(D3)}}] $(\beta_n^D)_{n\in\mathbb{N}}$ and $(\beta_n^G)_{n\in\mathbb{N}}$ satisfy that $\displaystyle{\lim_{n \to + \infty} n^{-1} \sum_{k=0}^n \beta_k^D = \lim_{n \to + \infty} n^{-1} \sum_{k=0}^n \beta_k^G = 0}$.
\end{enumerate}

\end{assum}
Assumption (D1) is the same as (A4) in \cite[(A4)]{Heusel2017} (see also \cite[(A1), (A2)]{BORKAR1997291}). {Assumption} (D2)(i) \cite[(A2)]{Heusel2017} {is} needed to guarantee the almost-sure convergence of Algorithm \ref{algo:1}, while Assumptions (D2)(ii) and {(D3)} are used to provide the rate of convergence of Algorithm \ref{algo:1}.

The following presents a convergence analysis as well as a convergence rate analysis of Algorithm \ref{algo:1} with diminishing learning rates. 

\begin{thm}\label{thm:2}
Suppose that Assumptions \ref{assum:1}(A1)--(A2) and \ref{const}(C2)--{(C3)} hold. Then, the following hold:

{\em (i)} {\em [Convergence]} Under Assumption \ref{dim}(D1) {and} (D2)(i), the sequence $((\bm{\theta}_n, \bm{w}_n))_{n\in\mathbb{N}}$ generated by Algorithm \ref{algo:1} converges almost surely to a point $(\bm{\theta}^\star, \bm{w}^\star) \in \mathrm{LNE}(\mathcal{L}_D, \mathcal{L}_G)$.

\vspace{-1mm}
{\em (ii)} {\em [Convergence Rate]} Under Assumption \ref{dim}(D2)(ii) and {(D3)},
\begin{equation}\label{R}
\begin{split}
&\frac{1}{N} \sum_{n\in [N]} \mathbb{E}\left[\left\langle \bm{\theta}_n - \bm{\theta}, 
\nabla_{\bm{\theta}} \mathcal{L}_G(\bm{\theta}_n, \bm{w}_n) \right\rangle \right]\\
&\leq \frac{C_1^2}{2 a_N N}
 + \frac{2 C_1 \tilde{K}_1}{N} \sum_{n\in [N]} \beta_n^{G}
+ 
\frac{2 \tilde{K}_1^2}{N} \sum_{n\in [N]} a_n,\\
&\frac{1}{N} \sum_{n\in [N]} \mathbb{E}\left[\left\langle \bm{w}_n - \bm{w}, 
\nabla_{\bm{w}} \mathcal{L}_D(\bm{\theta}_n, \bm{w}_n) \right\rangle \right] \\
&\leq
\frac{C_2^2}{2 b_N N}
+ \frac{2 C_2 \tilde{K}_2}{N} \sum_{n\in [N]} \beta_n^{D} 
+ 
\frac{2 \tilde{K}_2^2}{N} \sum_{n\in [N]} b_n.
\end{split}
\end{equation}
If we use $a_n = \mathcal{O}(n^{-\eta_a})$, $\beta_n^G = \mathcal{O}(n^{-\eta_a})$, $b_n = \mathcal{O}(n^{-\eta_b})$, and $\beta_n^D = \mathcal{O}(n^{-\eta_b})$, where $\eta_a, \eta_b \in (0,1)$, then 
\begin{equation}\label{R_1}
\begin{split}
&\frac{1}{N} \sum_{n\in [N]} \mathbb{E}\left[\left\langle \bm{\theta}_n - \bm{\theta}, 
\nabla_{\bm{\theta}} \mathcal{L}_G(\bm{\theta}_n, \bm{w}_n) \right\rangle \right]
\leq 
\mathcal{O}\left( \frac{1}{N^{\mu_a}} \right), \\
&
\frac{1}{N} \sum_{n\in [N]} \mathbb{E}\left[\left\langle \bm{w}_n - \bm{w}, 
\nabla_{\bm{w}} \mathcal{L}_D(\bm{\theta}_n, \bm{w}_n) \right\rangle \right]
\leq
\mathcal{O}\left( \frac{1}{N^{\mu_b}} \right),
\end{split}
\end{equation}
where $\mu_a := \min\{ \eta_a, 1 - \eta_a \}$ and $\mu_b := \min\{ \eta_b, 1 - \eta_b \}$. {Let $a, b > 0$ and a rate $\gamma \in (0,1)$ that decays every $T$ iterations. If we use 
\begin{align}\label{decaying_a_b}
\begin{split}
(a_n) 
&= (\underbrace{a,a,\ldots,a}_{T}, 
\underbrace{\gamma a, \gamma a, \ldots, \gamma a}_{T}, \ldots,\\
&\qquad \underbrace{\gamma^{P-1} a, \gamma^{P-1} a, \ldots, \gamma^{P-1} a}_{T}, \ldots),\\
(b_n) 
&= (\underbrace{b,b,\ldots,b}_{T}, 
\underbrace{\gamma b, \gamma b, \ldots, \gamma b}_{T}, \ldots,\\
&\qquad \underbrace{\gamma^{P-1} b, \gamma^{P-1} b, \ldots, \gamma^{P-1} b}_{T}, \ldots),
\end{split}
\end{align}
$\beta_n^G$ with $\sum_{n=0}^{+ \infty} \beta_n^G < + \infty$, and $\beta_n^D$ with $\sum_{n=0}^{+ \infty} \beta_n^D < + \infty$, then
\begin{align*}
&\frac{1}{N} \sum_{n\in [N]} \mathbb{E}\left[\left\langle \bm{\theta}_n - \bm{\theta}, 
\nabla_{\bm{\theta}} \mathcal{L}_G(\bm{\theta}_n, \bm{w}_n) \right\rangle \right]
\leq 
\mathcal{O}\left( \frac{1}{N} \right), \\
&
\frac{1}{N} \sum_{n\in [N]} \mathbb{E}\left[\left\langle \bm{w}_n - \bm{w}, 
\nabla_{\bm{w}} \mathcal{L}_D(\bm{\theta}_n, \bm{w}_n) \right\rangle \right]
\leq
\mathcal{O}\left( \frac{1}{N} \right),
\end{align*}
where $N = TP$.
}
\end{thm}

\begin{figure*}[t]
 \begin{minipage}[b]{0.71\linewidth}
 \centering
 \includegraphics[width=\linewidth]{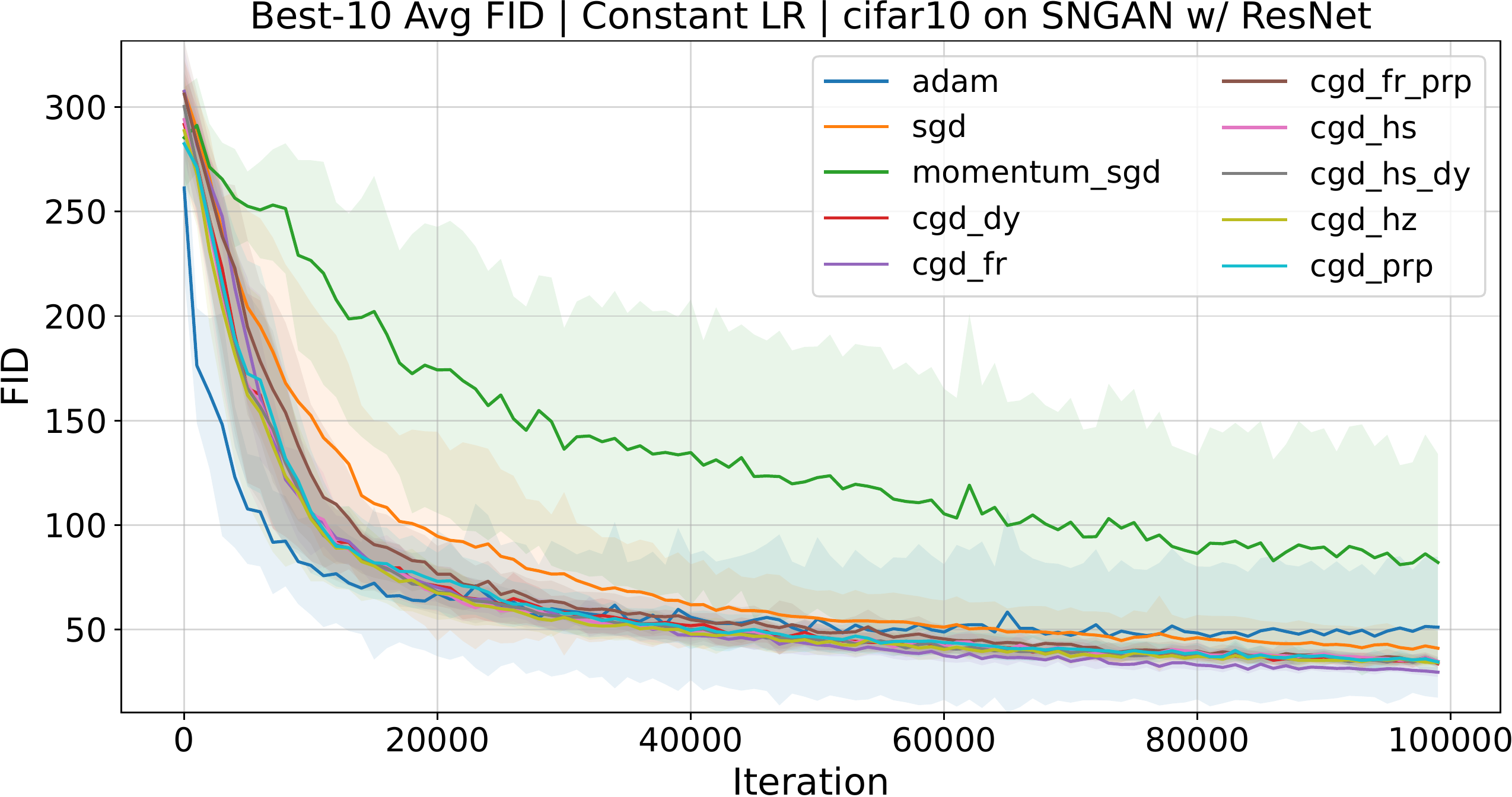}
 \end{minipage}
 \begin{minipage}[b]{0.28\linewidth}
 \centering
 \includegraphics[width=\linewidth]{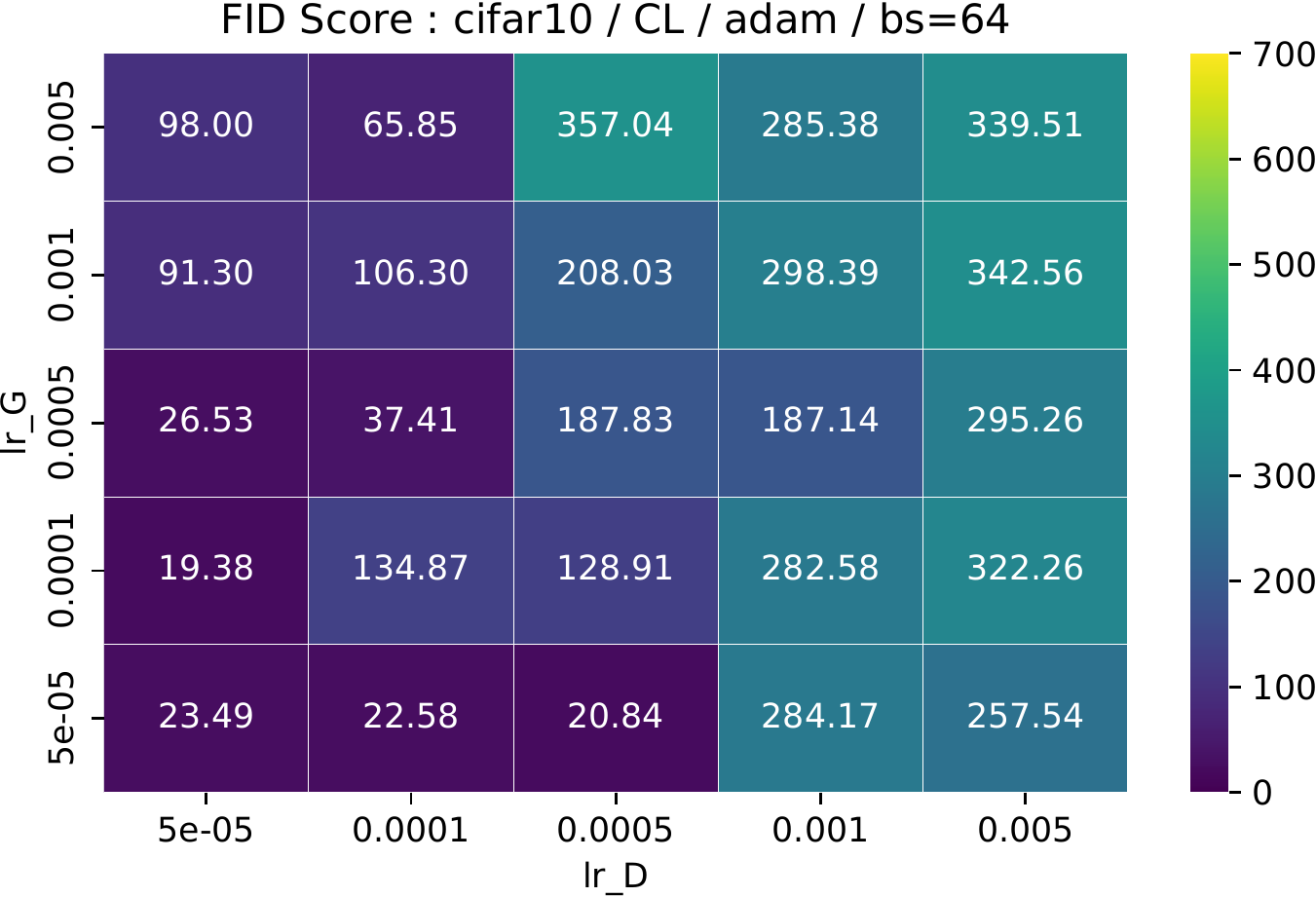}
 \includegraphics[width=\linewidth]{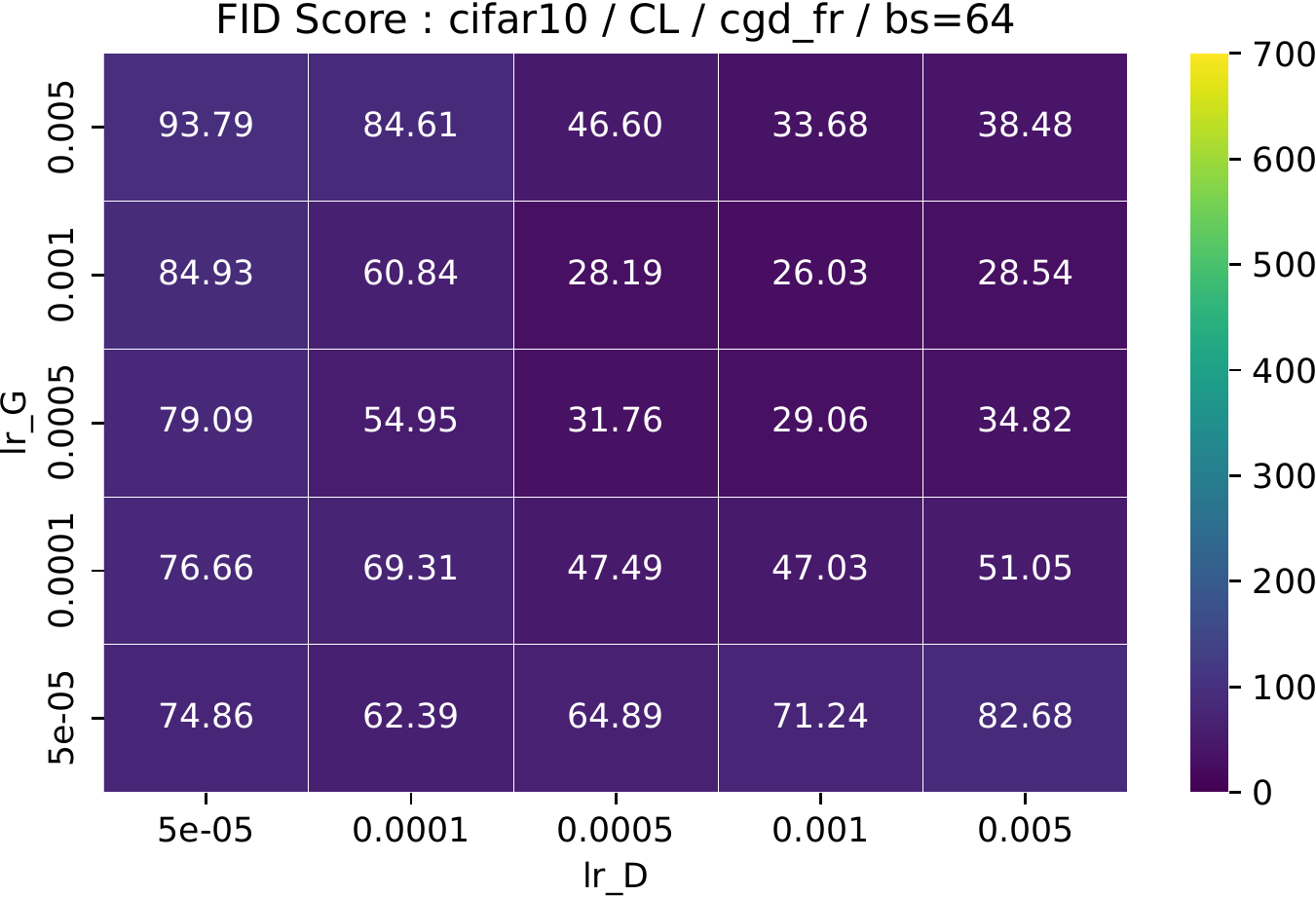}
 \end{minipage}
 \vspace{-3mm}
 \caption{\textbf{Constant LR CIFAR10 Experiments on SNGAN with ResNet Generator} / \textbf{Left}: Mean FID (solid line) bounded by the maximum and minimum over the best ten runs (the shaded areas) in the sense of FID. \textbf{Top Right}: Sensitivity of learning rate to FID scores of Adam. \textbf{Bottom Right}: Sensitivity of learning rate to FID scores of CG method (FR $\beta$ update rule) .}
 \label{fig:fig2}
\end{figure*}

\begin{table*}[tb]
\centering
 \setlength{\tabcolsep}{0.2em} 
\scalebox{0.765}{ 
\begin{tabular}{l||lll|lllllll}

 & adam & sgd & momentum\_sgd & cgd\_dy & cgd\_fr & cgd\_fr\_prp & cgd\_hs & cgd\_hs\_dy & cgd\_hz & cgd\_prp \\ \hline \hline
Miyato+2017 (CL\footnotemark[1]) & 21.7 & - & - & - & - & - & - & - & - & - \\ \hline
Ours (CL) / Best & 19.38 & 30.34 & 34.42 & 30.13 & {\textbf{26.03}} & 30.31 & 29.29 & 29.54 & 29.41 & 28.94 \\ 
Ours (DL\footnotemark[2]) / Best & 51.96 & 40.55 & 73.66 & 35.29 & 32.17 & 31.06 & 30.03 & 29.09 & {\textbf{29.64}} & 30.47 \\ \hline
Ours (CL) / Average & 51.16±33.71 & 41.09±8.13 & 82.15±51.82 & 33.70±2.04 & {\textbf{29.63±2.26}} & 34.78±2.17 & 34.82±1.07 & 34.12±1.31 & 34.28±0.96 & 34.53±2.01 \\ 
Ours (DL) / Average & 135.86±32.65 & 51.80±7.84 & 205.29±59.333 & 42.39±12.89 & 38.20±8.82 & 37.39±2.74 & {\textbf{36.96±2.77}} & 39.90±3.06 & 38.24±2.97 & 37.94±3.62 \\ 

 \end{tabular}
 } 
\caption{Best and Average FID scores for CIFAR10 on SNGAN with ResNet generator for {\em constant} and {\em dimimishing} learning rate scheduling. \ }
 \label{table:exp-fid-const}
\end{table*}

\section{Numerical Experiments}
\label{section:experiments}
\subsection{Overview}
\label{section:experiments-overview}
This section consists of two parts. In the first part, we examine the convergence of SGD, momentum SGD, and CG methods to a LNE when they use {\em constant} and {\em diminishing} learning rates in minimax optimization by using a toy example. In the second part, since this is the first attempt at applying CG methods to GAN training, we compare and evaluate the CG method against SGD, momentum SGD, and Adam. 

\subsection{Toy example}
\label{section:experiments-toy-example}
As an objective function, we chose $f(x,y) = (1+x^2) \cdot (100-y^2)$, which matches the experimental setting of \cite{Heusel2017}. Here, $x$ is minimized with the derivative $f_x = 2x \cdot (100-y^2)$ as the gradient direction and $y$ is maximized with the derivative $f_y = -2y \cdot (1+x^2)$ as the gradient direction. With $f(x,y)$ as the objective function, the LNE is $(x,y) = (0,0)$.

The trajectory of optimization is shown in Figure \ref{fig:toy-exp1} for a constant learning rate and diminishing learning rate with $\eta_a = \eta_b = $ 0.5. This toy example confirms convergence to a LNE for an appropriate learning rate. We can see that each optimizer follows a different trajectory, but eventually reaches the same convergence point. The objective oscillates, but converges to $f(x,y)=100$ (the value of the objective at LNE), and the norm decreases almost monotonically.

\footnotetext[1]{CL: {\em constant} learning rate}
\footnotetext[2]{DL: {\em diminishing} learning rate}

\subsection{Experiments on GANs}
\label{section:experiments-realworld-data-exp}
\begin{figure*}[t]
 \begin{minipage}[b]{0.71\linewidth}
 \centering
 \includegraphics[width=\linewidth]{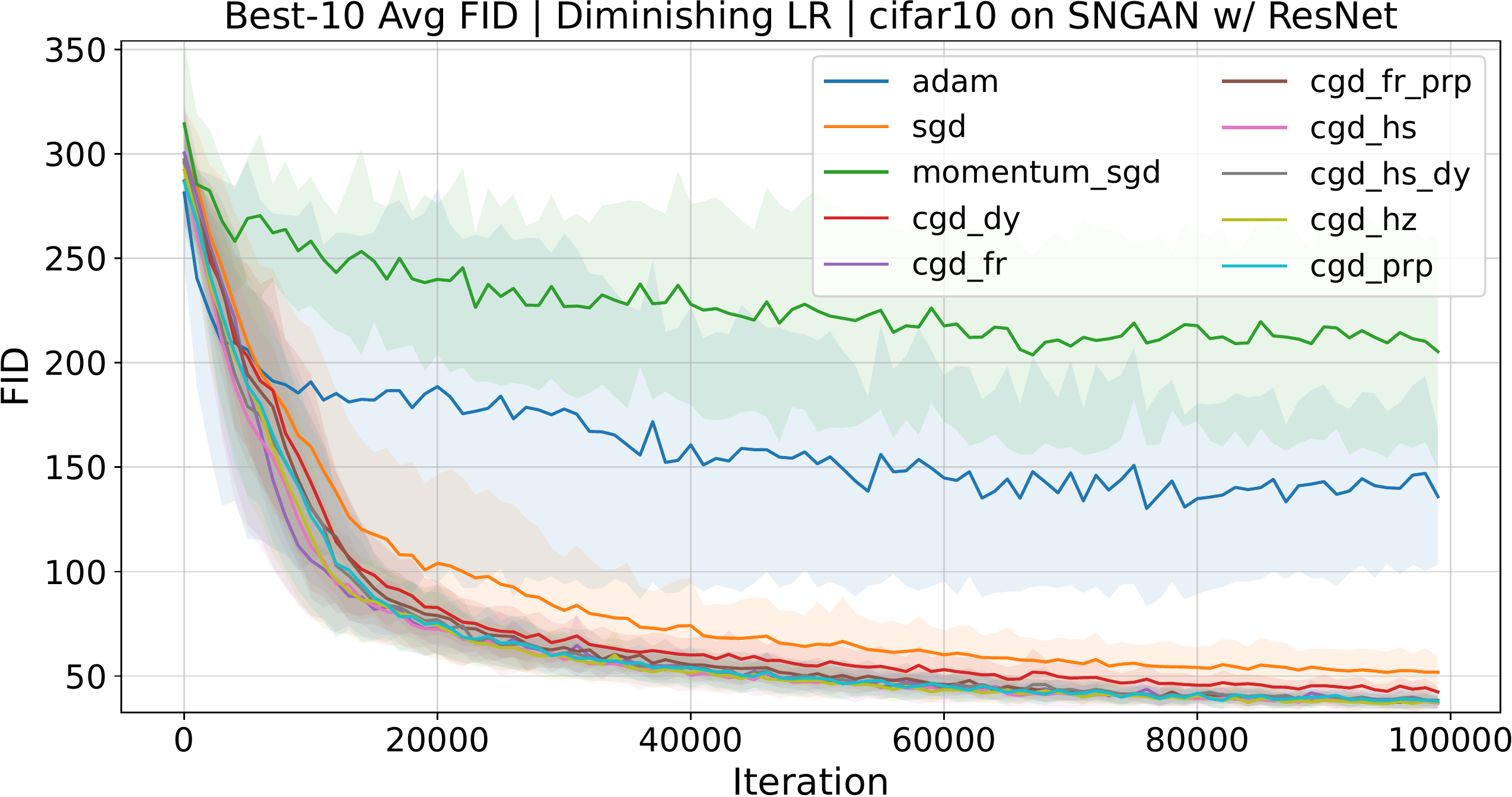}
 \end{minipage}
 \begin{minipage}[b]{0.28\linewidth}
 \centering
 \includegraphics[width=\linewidth]{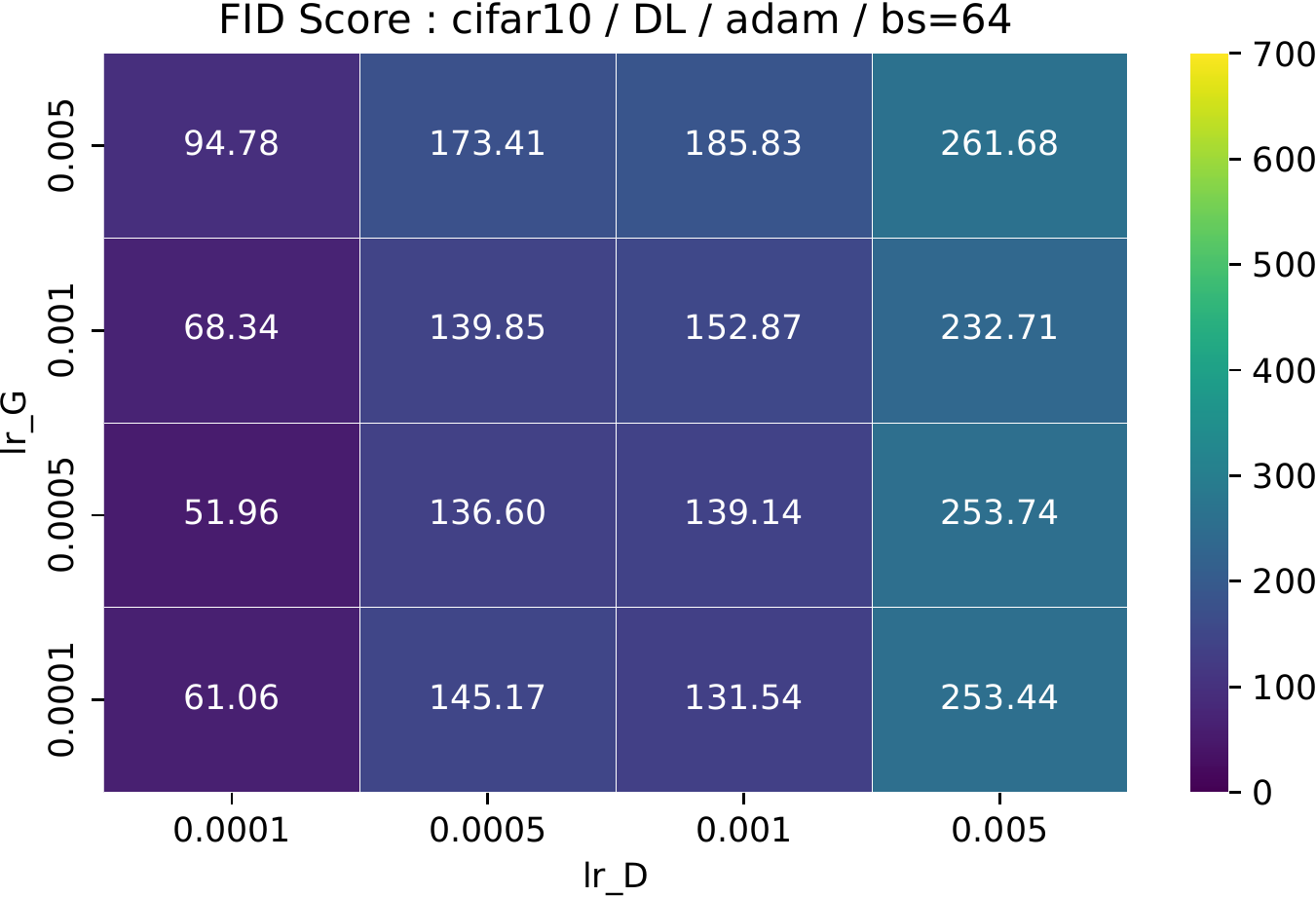}
 \includegraphics[width=\linewidth]{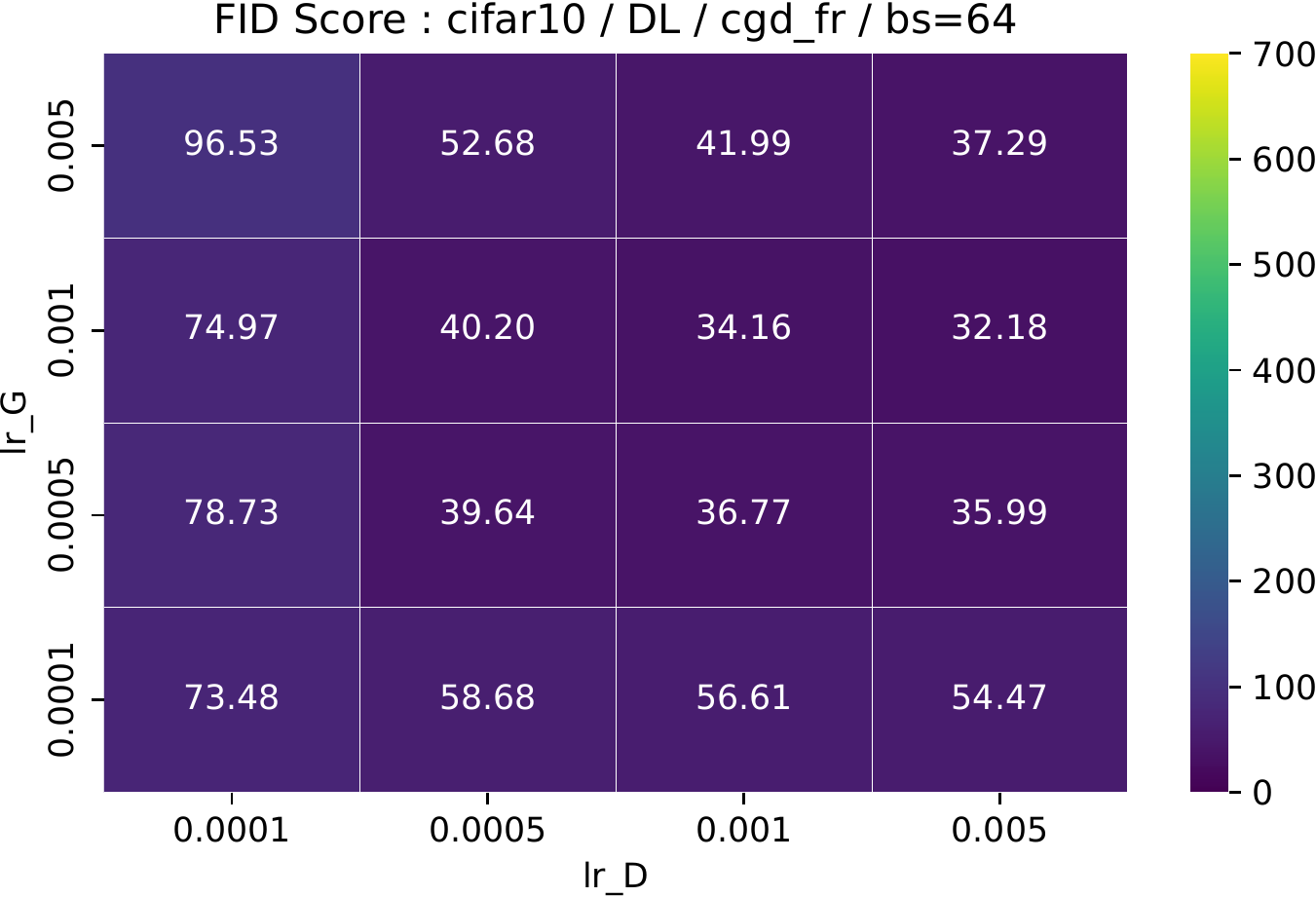}
 \end{minipage}
 \vspace{-3mm}
 \caption{\textbf{Diminishing LR CIFAR10 Experiments on SNGAN with ResNet Generator} / \textbf{Left}: Mean FID (solid line) bounded by the maximum and minimum over the best ten runs (the shaded areas) in the sense of FID. \textbf{Top Right}: Sensitivity of learning rate to FID scores of Adam. \textbf{Bottom Right}: Sensitivity of learning rate to FID scores of CG method (FR $\beta$ update rule) .}
 \label{fig:fig3}
\end{figure*}

{
We conducted a series of experiments using SNGAN with ResNet generator \cite{DBLP:conf/nips/ZhuangTDTDPD20}. To meet Assumption \ref{assum:1}(A2), we replaced the batch normalization layer with a spectral normalization (SN) layer \cite{DBLP:journals/corr/abs-1802-05957}.
}

{Here, we show that assumptions in Theorems \ref{thm:1} and \ref{thm:2} hold in the experiments. We used an SN layer to guarantee that (A2) holds (see also the above paragraph). We set $a_n$, $b_n$, $\beta_n^D$, and $\beta_n^G$ to satisfy (C1), (C2), (D2), and (D3) before implementing Algorithm \ref{algo:1}. The experiments indicated the boundedness (C3) of $(\bm{\theta}_n)_{n\in\mathbb{N}}$ and $(\bm{w}_n)_{n\in\mathbb{N}}$. To guarantee that Algorithm \ref{algo:1} satisfies (C3) before implementing the algorithm, we replaced $\bm{w}_{n+1}$ and $\bm{\theta}_{n+1}$ in Algorithm \ref{algo:1} with $\bm{w}_{n+1} = P^D (\bm{w}_n + b_n \bm{d}_n^D)$ and $\bm{\theta}_{n+1} = P^G (\bm{\theta}_n + a_n \bm{d}_n^G)$, where $P^D$ and $P^G$ are projections onto closed balls with sufficiently large radii. A discussion similar to the one proving Theorems \ref{thm:1} and \ref{thm:2}, together with the nonexpansivity condition of the projection ($\|P(\bm{x})-P(\bm{y})\| \leq \|\bm{x}-\bm{y}\|$), leads to versions of Theorems \ref{thm:1} and \ref{thm:2} for all $\bm{w}$ and $\bm{\theta}$ belonging to the closed balls. Meanwhile, it would be difficult to check whether (D1) holds in practical GANs. However, even if (D1) does not hold, Theorems \ref{thm:1} and \ref{thm:2}(ii) are guaranteed to hold.
}

In this experimental setting, the loss function is high-dimensional and non-convex, so a simple gradient norm does not necessarily correspond to the quality of learning of the GANs. Here, the Fréchet inception distance (FID) \cite{Heusel2017} is commonly used to evaluate the quality of GAN training, so we chose it as a metric to evaluate the training on real-world data. The smaller the value of this metric is, the better the training of the generator of the GANs becomes. For the CG methods, we tested the seven beta update rules described in Section \ref{cgm}. 

For the hyperparameter search, we conducted a grid search for all optimizers and datasets. The details of the hyperparameters are in Appendix \ref{appendix:hyperparameters}, and the remaining experimental settings are detailed in Appendix \ref{appendix:setting}.

\subsubsection{Constant learning rate rule}
{
A {\em constant} learning rate is practically used for training GANs, including those in \cite{Heusel2017}, \cite{DBLP:conf/nips/ZhuangTDTDPD20}. Before conducting the {\em diminishing} learning rate experiment, we compared the training behaviors of the four optimizers when using a {\em constant} learning rate. We focused on the optimization methods shown in Table \ref{table:1}. Still, for real-world problems, it is desirable to consider cases in which a popular optimizer such as Adam is used. Therefore, we compared SGD, Momentum SGD, Adam, and seven CG algorithms when SNGAN with ResNet generator is used as the model on the CIFAR10 dataset.
}

{
Comparing the FID scores of the trained model by the CG method and the other optimizers in Table \ref{table:exp-fid-const} reveals that the CG methods outperformed the conventional optimizers (SGD and momentum SGD) in the case of the constant learning rate schedule.
}

The experimental results showed that not only did the CG method outperform the other optimizers (including Adam) on average in terms of FID score in the best ten trials (Figure \ref{fig:fig2} Left), but it was also more robust than any other optimizer in terms of hyperparameter sensitivities (Figure \ref{fig:fig2} Right) (Details are in Appendix \ref{appendix:resnet-lr-sensitivity}).

\subsubsection{Diminishing learning rate rule}
\label{section:experiments-dim}
{
Here, we present experimental results in the case of a {\em diminishing} learning rate. To avoid learning stagnation due to an excessively decaying learning rate, we implemented the methods using LR $a_n = (\underbrace{a,a,\ldots,a}_{T}, \underbrace{\gamma a, \gamma a,\ldots,\gamma a}_{T}, \ldots, \underbrace{\gamma^{P-1} a, \gamma^{P-1} a,\ldots,\gamma^{P-1} a}_{T} )$ with an initial LR $a$ and a rate $\gamma > 0$ that decays every $T$ iterations, where $a \in \{5\cdot 10^{-5}, 10^{-4}, 5 \cdot 10^{-4}, 10^{-3}, 5\cdot 10^{-3} \}$, $\gamma = 0.9$, $T = 10$K, and $P= N/T = 10$.
}

{
As can be seen from Table \ref{table:exp-fid-const}, the CG method achieved a competitive FID score close to that with a {\em constant} learning rate. In contrast, methods such as SGD, Momentum SGD, and Adam showed degraded performance. It should be noted that Adam can also achieve a reasonably good FID score by carefully tuning the hyperparameters, as shown in Figure \ref{fig:fig3} (Right). In contrast, the CG method is less sensitive to hyperparameters in the range of our experiments. The details of the other optimizers and results for other beta update rules for the CG method are shown in Appendix \ref{appendix:resnet-lr-sensitivity}.
}

{Table \ref{table:exp-fid-const} also indicates that FR, HS, and HZ, which is a modification of HS, minimize the FID scores. This would be because the FR and HZ for training GANs inherit their descent properties in minimizing general non-convex functions (see \cite{Hager2006} for the sufficient descent properties of CG methods).}

\section{Conclusion}
For training GANs, we proposed CG-type algorithms to solve LNE problems and proved their convergence to LNEs under {\em constant} and {\em diminishing} learning rates under mild assumptions. Furthermore, we analyzed the convergence rates of the SGD, momentum SGD, and CG methods. We compared our method with SGD and momentum SGD on a toy problem, with experimental results consistent with theory. Finally, we evaluated the optimizer by training GANs using the FID score as a metric. We demonstrated that under {\em constant} and {\em diminishing} learning rates, the CG method outperformed SGD and momentum SGD and outperformed Adam on average in real-world problem settings. 

{
One limitation of this study is that it involved experiments on only CIFAR10 datasets. Although the CG method minimized the FID scores in both cases, validations on other real-world datasets will be needed to support our claims. Another issue is that we did not use the state-of-the-art GAN model. We used a simple model to validate the theoretical contribution in Table \ref{table:1} under corresponding conditions. Thus, our method still needs to be compared with, e.g., StyleGAN \cite{karras2019style} and other models that have shown high performance in recent years.
}

\section{Acknowledgements}
{We are sincerely grateful to the meta-reviewer and the five anonymous reviewers for helping us improve the original manuscript.} We would like to thank Charles Guille-Escuret (Mila, UdeM) and Reyhane Askari (Mila, UdeM) for their helpful feedback and also thank Tetsuya Motokawa (Tsukuba University) and Izumu Hosoi (University of Tokyo) for their help on the PyTorch implementation. This work was supported by the Masason Foundation Fellowship awarded to Hiroki Naganuma and by the Japan Society for the Promotion of Science (JSPS) KAKENHI Grant Number 21K11773 awarded to Hideaki Iiduka. This research used the computational resources of the TSUBAME3.0 supercomputer\footnote{\url{https://www.t3.gsic.titech.ac.jp/en}} provided by Tokyo Institute of Technology through the Exploratory Joint Research Project Support Program from JHPCN (EX22401) and TSUBAME Encouragement Program for Young / Female Users.

\printbibliography[title={References}]

\onecolumn{

\begin{appendices}

\section{Lemmas}
\label{appendix:lemma}
We define $\bm{d}^{D}(\bm{\theta}_n, \bm{w}_n) = \bm{d}_n^{D}$ and $\bm{d}^{G}(\bm{\theta}_n, \bm{w}_n) = \bm{d}_n^{G}$. We prove the following lemma:
 
\begin{lem}\label{lem:0}
Under (A1), (A2), and {(C3)}, for all $n\in\mathbb{N}$,
\begin{align*}
\mathbb{E} \left[ \left\| \mathcal{G}(\bm{\theta}_n, \bm{w}_n) \right\|^2 \right]
\leq
K_1^2 :=
\frac{(C_1 L_1 + \tilde{C}_1)^2}{m^2},
\end{align*}
where $C_1$ and $\tilde{C}_1$ denote upper bounds of $\|\bm{\theta}_n - \bm{\theta}\|$ and $\|\mathcal{G}_{\bm{w}_n}(\bm{\theta}) \|$ for all $\bm{\theta} \in \mathbb{R}^\Theta$ and all $n\in\mathbb{N}$, respectively. Additionally, under (C2), we have that, for all $n\in\mathbb{N}$,
\begin{align*}
\mathbb{E} \left[ \left\| \bm{d}^G (\bm{\theta}_n, \bm{w}_n) \right\|^2 \right] \leq 4 \tilde{K}_1^2,
\end{align*}
where $\tilde{K}_1^2 := \max \{ K_1^2, \| \bm{d}^G (\bm{\theta}_{-1}, \bm{w}_{-1}) \|^2 \}$.
\end{lem}

{\em Proof of Lemma \ref{lem:0}:} Let $\bm{\theta} \in \mathbb{R}^\Theta$ be fixed arbitrarily. Assumption {(C3)} implies that there exists $C_1 \in \mathbb{R}_{++}$ such that $\sup \{ \| \bm{\theta}_n - \bm{\theta} \| \colon n\in\mathbb{N} \} \leq C_1$. Assumption (A2) thus implies that, for all $n\in\mathbb{N}$,
\begin{align*}
\left\| \mathcal{G}_{\bm{w}_n}(\bm{\theta}_n) - \mathcal{G}_{\bm{w}_n}(\bm{\theta}) \right\|
\leq L_1 \left\| \bm{\theta}_n - \bm{\theta} \right\|
\leq C_1 L_1.
\end{align*}
Assumption (A1) ensures that $\|\mathcal{G}_{(\cdot)}(\bm{\theta}) \| \colon \mathbb{R}^W \to \mathbb{R}$ is continuous. Hence, {(C3)} guarantees that there exists $\tilde{C}_1 \in \mathbb{R}_{++}$ such that $\sup \{ \|\mathcal{G}_{\bm{w}_n}(\bm{\theta}) \| \colon n\in\mathbb{N} \} \leq \tilde{C}_1$. The triangle inequality guarantees that, for all $n\in\mathbb{N}$,
\begin{align*}
\left\| \mathcal{G}_{\bm{w}_n}(\bm{\theta}_n) \right\|
\leq
\left\| \mathcal{G}_{\bm{w}_n}(\bm{\theta}_n) - \mathcal{G}_{\bm{w}_n}(\bm{\theta}) \right\|
+ 
\left\| \mathcal{G}_{\bm{w}_n}(\bm{\theta}) \right\|
\leq
C_1 L_1 + \tilde{C}_1,
\end{align*}
which implies that
\begin{align*}
\left\| \mathcal{G}(\bm{\theta}_n, \bm{w}_n) \right\|^2
\leq
\frac{1}{m^2} \left\| \mathcal{G}_{\bm{w}_n}(\bm{\theta}_n) \right\|^2
\leq
\frac{(C_1 L_1 + \tilde{C}_1)^2}{m^2}.
\end{align*}
Accordingly, 
\begin{align*}
\mathbb{E} \left[ \left\| \mathcal{G}(\bm{\theta}_n, \bm{w}_n) \right\|^2 \right]
\leq
\frac{(C_1 L_1 + \tilde{C}_1)^2}{m^2} 
=: K_1^2.
\end{align*}

Let us define $\tilde{K}_1^2 := \max \{ K_1^2, \| \bm{d}^G (\bm{\theta}_{-1}, \bm{w}_{-1}) \|^2 \}$. We will use mathematical induction to show that, for all $n\in\mathbb{N}$, $\mathbb{E}[ \| \bm{d}^G (\bm{\theta}_n, \bm{w}_n) \|^2 ] \leq 4 \tilde{K}_1^2$. When $n=1$, we have that
\begin{align*}
\mathbb{E}\left[ \left\| \bm{d}^G (\bm{\theta}_1, \bm{w}_1) \right\|^2 \right]
&\leq
2 \mathbb{E}\left[\left\| \mathcal{G}(\bm{\theta}_1, \bm{w}_1) \right\|^2 \right]
+ 
2{\beta_1^G}^2 \mathbb{E}\left[\left\| \bm{d}^G (\bm{\theta}_{-1}, \bm{w}_{-1}) \right\|^2 \right]\\
&\leq
2 K_1^2 + 2 \left(\frac{1}{2} \right)^2 4 \tilde{K}_1^2
= 4 \tilde{K}_1^2,
\end{align*}
where the second inequality comes from the condition $\beta_n^G \in [0,1/2]$. Assume that $\mathbb{E}[ \| \bm{d}^G (\bm{\theta}_n, \bm{w}_n) \|^2 ] \leq 4 \tilde{K}_1^2$ for some $n$. We have that
\begin{align*}
\mathbb{E}\left[ \left\| \bm{d}^G (\bm{\theta}_{n+1}, \bm{w}_{n+1}) \right\|^2 \right]
&\leq
2 \mathbb{E}\left[\left\| \mathcal{G}(\bm{\theta}_{n+1}, \bm{w}_{n+1}) \right\|^2 \right]
+ 
2{\beta_n^G}^2 \mathbb{E}\left[\left\| \bm{d}^G (\bm{\theta}_{n}, \bm{w}_{n}) \right\|^2 \right]\\
&\leq
2 K_1^2 + 2 \left(\frac{1}{2} \right)^2 4 \tilde{K}_1^2
= 4 \tilde{K}_1^2.
\end{align*}
Hence, we have that, for all $n\in\mathbb{N}$, $\mathbb{E}[ \| \bm{d}^G (\bm{\theta}_n, \bm{w}_n) \|^2 ] \leq 4 \tilde{K}_1^2$. $\square$

A discussion similar to the one proving Lemma \ref{lem:0} leads to the following.

\begin{lem}\label{lem:0_1}
Under (A1), (A2), and {(C3)}, for all $n\in\mathbb{N}$,
\begin{align*}
\mathbb{E} \left[ \left\| \mathcal{D}(\bm{\theta}_n, \bm{w}_n) \right\|^2 \right]
\leq
K_2^2 :=
\frac{(C_2 L_2 + \tilde{C}_2)^2}{m^2},
\end{align*}
where $C_2$ and $\tilde{C}_2$ denote upper bounds of $\|\bm{w}_n - \bm{w}\|$ and $\|\mathcal{D}_{\bm{\theta}_n}(\bm{w}) \|$ for all $\bm{w} \in \mathbb{R}^W$ and all $n\in\mathbb{N}$, respectively. Additionally, under (C2), we have that, for all $n\in\mathbb{N}$,
\begin{align*}
\mathbb{E} \left[ \left\| \bm{d}^D (\bm{\theta}_n, \bm{w}_n) \right\|^2 \right] \leq 4 \tilde{K}_2^2,
\end{align*}
where $\tilde{K}_2^2 := \max \{ K_2^2, \| \bm{d}^D (\bm{\theta}_{-1}, \bm{w}_{-1}) \|^2 \}$.
\end{lem}

\begin{lem}\label{lem:1}
Under the assumptions in Lemma \ref{lem:0}, for all $\bm{\theta} \in \mathbb{R}^\Theta$, all $\bm{w} \in \mathbb{R}^W$, and all $n\in\mathbb{N}$,
\begin{align*}
&\mathbb{E} \left[ \left\|\bm{\theta}_{n+1} - \bm{\theta} \right\|^2 \right]
\leq \mathbb{E} \left[ \left\| \bm{\theta}_n - \bm{\theta} \right\|^2 \right] 
 + 2 a_n 
 \left( 
 \mathbb{E}\left[\left\langle \bm{\theta} - \bm{\theta}_n, 
\nabla_{\bm{\theta}} \mathcal{L}_G(\bm{\theta}_n, \bm{w}_n) \right\rangle \right] + 2 C_1 \tilde{K}_1 \beta_n^{G}
 \right)
 + 4 \tilde{K}_1^2 a_n^2,\\
 &\mathbb{E} \left[ \left\|\bm{w}_{n+1} - \bm{w} \right\|^2 \right]
\leq \mathbb{E} \left[ \left\| \bm{w}_n - \bm{w} \right\|^2 \right] 
 + 2 b_n 
 \left( 
 \mathbb{E}\left[\left\langle \bm{w} - \bm{w}_n, 
\nabla_{\bm{w}} \mathcal{L}_D(\bm{\theta}_n, \bm{w}_n) \right\rangle \right] + 2 C_2 \tilde{K}_2 \beta_n^{D}
 \right)
 + 4 \tilde{K}_2^2 b_n^2.
\end{align*}
\end{lem}

{\em Proof of Lemma \ref{lem:1}:} Let $\bm{\theta}\in \mathbb{R}^\Theta$ and $n \in \mathbb{N}$ be fixed arbitrarily. The definition of $\bm{d}^{G}(\bm{\theta}_n, \bm{w}_n)$ ensures that 
\begin{align*}
\left\langle \bm{\theta}_n - \bm{\theta}, \bm{d}^{G}(\bm{\theta}_n, \bm{w}_n) \right\rangle
= 
\left\langle \bm{\theta} - \bm{\theta}_n, \mathcal{G}(\bm{\theta}_n, \bm{w}_n) \right\rangle
+ \beta_n^{G} \left\langle \bm{\theta}_n - \bm{\theta}, \bm{d}^{G}(\bm{\theta}_{n-1}, \bm{w}_{n-1}) \right\rangle,
\end{align*}
which, together with the Cauchy--Schwarz inequality and Lemma \ref{lem:0}, implies that 
\begin{align*}
\left\langle \bm{\theta}_n - \bm{\theta}, \bm{d}^{G}(\bm{\theta}_n, \bm{w}_n) \right\rangle
\leq 
\left\langle \bm{\theta} - \bm{\theta}_n, \mathcal{G}(\bm{\theta}_n, \bm{w}_n) \right\rangle
+ \beta_n^{G} C_1 \left\| \bm{d}^{G}(\bm{\theta}_{n-1}, \bm{w}_{n-1}) \right\|.
\end{align*}
Accordingly, Lemma \ref{lem:0} and Jensen's inequality give 
\begin{align*}
\mathbb{E}\left[ \left\langle \bm{\theta}_n - \bm{\theta}, \bm{d}^{G}(\bm{\theta}_n, \bm{w}_n) \right\rangle \right]
\leq 
\mathbb{E}\left[\left\langle \bm{\theta} - \bm{\theta}_n, \mathcal{G}(\bm{\theta}_n, \bm{w}_n) \right\rangle \right]
+ 2 C_1 \tilde{K}_1 \beta_n^{G}.
\end{align*}
From the definition of $\mathcal{G}(\bm{\theta}_n, \bm{w}_n)$, we also have that
\begin{align*}
\mathbb{E}\left[\left\langle \bm{\theta} - \bm{\theta}_n, \mathcal{G}(\bm{\theta}_n, \bm{w}_n) \right\rangle \right]
&=
\frac{1}{m} \mathbb{E}\left[\left\langle \bm{\theta} - \bm{\theta}_n, \mathcal{G}_{\bm{w}_n}(\bm{\theta}_n) \right\rangle \right].
\end{align*}
Let us denote the independent and identically distributed samples considered here by $\xi_0, \xi_1, \ldots$ and denote the history of process $\xi_0, \xi_1, \ldots$ to time step $n$ by $\xi_{[n]} := (\xi_0, \xi_1, \ldots, \xi_n)$. Accordingly, we have
\begin{align*}
\mathbb{E}\left[\left\langle \bm{\theta} - \bm{\theta}_n, \frac{1}{m}\mathcal{G}_{\bm{w}_n}(\bm{\theta}_n) \right\rangle \right]
&=
\mathbb{E}\left[ \mathbb{E} \left[ 
\left\langle \bm{\theta} - \bm{\theta}_n, \frac{1}{m} \mathcal{G}_{\bm{w}_n}(\bm{\theta}_n) \right\rangle \bigg| \xi_{[n]} \right] \right]
=
\mathbb{E}\left[ 
\left\langle \bm{\theta} - \bm{\theta}_n, \mathbb{E} \left[ \frac{1}{m}\mathcal{G}_{\bm{w}_n}(\bm{\theta}_n) \bigg| \xi_{[n]} \right]\right\rangle \right]\\
&
= 
\mathbb{E}\left[\left\langle \bm{\theta} - \bm{\theta}_n, 
\nabla_{\bm{\theta}} \mathcal{L}_G(\bm{\theta}_n, \bm{w}_n) \right\rangle \right].
\end{align*}
Therefore, we have
\begin{align*}
\mathbb{E}\left[ \left\langle \bm{\theta}_n - \bm{\theta}, \bm{d}^{G}(\bm{\theta}_n, \bm{w}_n) \right\rangle \right]
\leq 
\mathbb{E}\left[\left\langle \bm{\theta} - \bm{\theta}_n, 
\nabla_{\bm{\theta}} \mathcal{L}_G(\bm{\theta}_n, \bm{w}_n) \right\rangle \right] + 2 C_1 \tilde{K}_1 \beta_n^{G}.
\end{align*}
The definition of $\bm{\theta}_{n+1}$ guarantees that
\begin{align*}
\left\|\bm{\theta}_{n+1} - \bm{\theta} \right\|^2 
&= \left\| (\bm{\theta}_n - \bm{\theta}) + a_n \bm{d}^{G}(\bm{\theta}_n, \bm{w}_n) \right\|^2\\
&= \left\| \bm{\theta}_n - \bm{\theta} \right\|^2 
 + 2 a_n \left\langle \bm{\theta}_n - \bm{\theta}, \bm{d}^{G}(\bm{\theta}_n, \bm{w}_n) \right\rangle 
 + a_n^2 \left\| \bm{d}^{G}(\bm{\theta}_n, \bm{w}_n) \right\|^2.
\end{align*}
Taking the expectation of the above inequality gives
\begin{align*}
\mathbb{E} \left[ \left\|\bm{\theta}_{n+1} - \bm{\theta} \right\|^2 \right]
\leq \mathbb{E} \left[ \left\| \bm{\theta}_n - \bm{\theta} \right\|^2 \right] 
 + 2 a_n 
 \left( 
 \mathbb{E}\left[\left\langle \bm{\theta} - \bm{\theta}_n, 
\nabla_{\bm{\theta}} \mathcal{L}_G(\bm{\theta}_n, \bm{w}_n) \right\rangle \right] + 2 C_1 \tilde{K}_1 \beta_n^{G}
 \right)
 + 4 \tilde{K}_1^2 a_n^2.
\end{align*}
A discussion similar to the one showing the above inequality leads to the following inequality:
\begin{align*}
\mathbb{E} \left[ \left\|\bm{w}_{n+1} - \bm{w} \right\|^2 \right]
\leq \mathbb{E} \left[ \left\| \bm{w}_n - \bm{w} \right\|^2 \right] 
 + 2 b_n 
 \left( 
 \mathbb{E}\left[\left\langle \bm{w} - \bm{w}_n, 
\nabla_{\bm{w}} \mathcal{L}_D(\bm{\theta}_n, \bm{w}_n) \right\rangle \right] + 2 C_2 \tilde{K}_2 \beta_n^{D}
 \right)
 + 4 \tilde{K}_2^2 b_n^2.
\end{align*}
This completes the proof. $\square$

\section{Proofs of Theorems \ref{thm:1} and \ref{thm:2}}
\label{appendix:proof}
\subsection{Outline of Theorem \ref{thm:1}} 
First, we provide a brief outline of the proof. Assumptions \ref{assum:1}(A1)--(A2) and \ref{const}(C2)--{(C3)} imply that $(\mathbb{E}[\|\bm{d}^D(\bm{\theta}_n, \bm{w}_n)\|])_{n\in\mathbb{N}}$ and $(\mathbb{E}[\|\bm{d}^G(\bm{\theta}_n, \bm{w}_n)\|])_{n\in\mathbb{N}}$ are bounded (Lemmas \ref{lem:0} and \ref{lem:0_1}). Next, we evaluate the squared norm $\|\bm{\theta}_{n+1} - \bm{\theta}\|^2$ ($\bm{\theta}\in \mathbb{R}^\Theta$) to find the relationship between $\|\bm{\theta}_{n+1} - \bm{\theta}\|$ and $\|\bm{\theta}_{n} - \bm{\theta}\|$. Using the expansion of the squared norm leads to
\begin{align*}
 \left \|\bm{\theta}_{n+1} - \bm{\theta}\right\|^2
 \leq \left \|\bm{\theta}_{n} - \bm{\theta}\right\|^2
 + 2 a_n \left\langle \bm{\theta}_{n} - \bm{\theta}, \bm{d}^G (\bm{\theta}_n, \bm{w}_n) \right\rangle
 + a_n^2 \left\|\bm{d}^G (\bm{\theta}_n, \bm{w}_n) \right\|^2.
\end{align*}
The definition of $\bm{d}^G (\bm{\theta}_n, \bm{w}_n)$ and the boundedness of $(\mathbb{E}[\|\bm{d}^G(\bm{\theta}_n, \bm{w}_n)\|])_{n\in\mathbb{N}}$ imply that there exist positive constants $C_1$ and $\tilde{K}_1$, which depend on (A2) and {(C3)}, such that, for all $n\in\mathbb{N}$,
\begin{align}\label{key:1}
\mathbb{E} \left[ \left\|\bm{\theta}_{n+1} - \bm{\theta} \right\|^2 \right]
\leq \mathbb{E} \left[ \left\| \bm{\theta}_n - \bm{\theta} \right\|^2 \right] 
 + 2 a_n 
 \left( 
 \mathbb{E}\left[\left\langle \bm{\theta} - \bm{\theta}_n, 
\nabla_{\bm{\theta}} \mathcal{L}_G(\bm{\theta}_n, \bm{w}_n) \right\rangle \right] + 2 C_1 \tilde{K}_1 \beta_n^{G}
 \right)
 + 4 \tilde{K}_1^2 a_n^2.
\end{align}
Inequality \eqref{key:1} is a key inequality to prove Theorem \ref{thm:1}. We can show \eqref{liminf:1} by contradiction and \eqref{key:1} with (C1). The limit inferior of $\mathbb{E}[\langle \bm{\theta}_n - \bm{\theta}, \nabla_{\bm{\theta}} \mathcal{L}_G(\bm{\theta}_n, \bm{w}_n) \rangle]$ in \eqref{liminf:1} ensures that there exists a subsequence $((\bm{\theta}_{n_i}, \bm{w}_{n_i}))_{i\in \mathbb{N}}$ of $((\bm{\theta}_{n}, \bm{w}_n))_{n\in\mathbb{N}}$ such that $((\bm{\theta}_{n_{i}}, \bm{w}_{n_{i}}))_{i\in \mathbb{N}}$ converges almost surely to $(\bm{\theta}^\star, \bm{w}^\star)$ satisfying that, for all $\bm{\theta} \in \mathbb{R}^\Theta$,
\begin{align*}
\mathbb{E}\left[\left\langle \bm{\theta}^\star - \bm{\theta},
\nabla_{\bm{\theta}} \mathcal{L}_G(\bm{\theta}^\star, \bm{w}^\star) \right\rangle \right]
=
\liminf_{n \to + \infty} \mathbb{E}\left[\left\langle \bm{\theta}_n - \bm{\theta},
\nabla_{\bm{\theta}} \mathcal{L}_G(\bm{\theta}_n, \bm{w}_n) \right\rangle \right],
\end{align*}
which, together with $\bm{\theta} := \bm{\theta}^\star - \nabla_{\bm{\theta}} \mathcal{L}_G(\bm{\theta}^\star, \bm{w}^\star)$, implies that
\begin{align*}
 \mathbb{E}\left[\left\| 
\nabla_{\bm{\theta}} \mathcal{L}_G(\bm{\theta}^\star, \bm{w}^\star) \right\|^2 \right]
=
\liminf_{n \to + \infty} \mathbb{E}\left[\left\langle \bm{\theta}_n - (\bm{\theta}^\star - \nabla_{\bm{\theta}} \mathcal{L}_G(\bm{\theta}^\star, \bm{w}^\star)),
\nabla_{\bm{\theta}} \mathcal{L}_G(\bm{\theta}_n, \bm{w}_n) \right\rangle \right].
\end{align*}
Accordingly, we have \eqref{liminf:1_1}. A discussion similar to the one showing \eqref{key:1} leads to the following key inequality: there exist positive constants $C_2$ and $\tilde{K}_2$, which depend on (A2) and {(C3)}, such that, for all $n\in\mathbb{N}$,
\begin{align}\label{key:2}
\mathbb{E} \left[ \left\|\bm{w}_{n+1} - \bm{w} \right\|^2 \right]
\leq \mathbb{E} \left[ \left\| \bm{w}_n - \bm{w} \right\|^2 \right] 
 + 2 b_n 
 \left( 
 \mathbb{E}\left[\left\langle \bm{w} - \bm{w}_n, 
\nabla_{\bm{w}} \mathcal{L}_D(\bm{\theta}_n, \bm{w}_n) \right\rangle \right] + 2 C_2 \tilde{K}_2 \beta_n^{D}
 \right)
 + 4 \tilde{K}_2^2 b_n^2.
\end{align}
A discussion similar to the one showing \eqref{liminf:1} and \eqref{liminf:1_1}, together with \eqref{key:2}, leads to \eqref{liminf:2} and \eqref{liminf:2_1}. Summing \eqref{key:1} with (C1) from $n=1$ to $n = N$ gives \eqref{sum_g}. Similarly, summing \eqref{key:2} from $n=1$ to $n=N$ gives \eqref{sum_d}. Let us consider the case where $((\bm{\theta}_n, \bm{w}_n))_{n\in\mathbb{N}}$ converges almost surely to $(\bm{\theta}^\star, \bm{w}^\star)$. Inequalities \eqref{liminf:1_1} and \eqref{liminf:2_1} thus guarantee \eqref{ane}.

The following is the detailed proof of Theorem \ref{thm:1}.

{\em Proof of Theorem \ref{thm:1}:} Let $\bm{\theta} \in \mathbb{R}^\Theta$ and $\bm{w} \in \mathbb{R}^W$ be fixed arbitrarily. First, we show that, for all $\epsilon > 0$, 
\begin{align}\label{inf}
\liminf_{n \to + \infty} \mathbb{E}\left[\left\langle \bm{\theta}_n - \bm{\theta}, 
\nabla_{\bm{\theta}} \mathcal{L}_G(\bm{\theta}_n, \bm{w}_n) \right\rangle \right]
\leq 2 \tilde{K}_1^2 a + 2 C_1 \tilde{K}_1 \beta^G + \epsilon.
\end{align}
If \eqref{inf} does not hold, then there exists $\epsilon_0 > 0$ such that 
\begin{align*}
\liminf_{n \to + \infty} \mathbb{E}\left[\left\langle \bm{\theta}_n - \bm{\theta}, 
\nabla_{\bm{\theta}} \mathcal{L}_G(\bm{\theta}_n, \bm{w}_n) \right\rangle \right]
> 2 \tilde{K}_1^2 a + 2 C_1 \tilde{K}_1 \beta^G 
+ \epsilon_0.
\end{align*}
Since there exists $n_0 \in \mathbb{N}$ such that, for all $n\geq n_0$,
\begin{align*}
\liminf_{n \to + \infty} \mathbb{E}\left[\left\langle \bm{\theta}_n - \bm{\theta}, 
\nabla_{\bm{\theta}} \mathcal{L}_G(\bm{\theta}_n, \bm{w}_n) \right\rangle \right] 
- \frac{\epsilon_0}{2}
\leq 
\mathbb{E}\left[\left\langle \bm{\theta}_n - \bm{\theta}, 
\nabla_{\bm{\theta}} \mathcal{L}_G(\bm{\theta}_n, \bm{w}_n) \right\rangle \right],
\end{align*}
we have that, for all $n \geq n_0$,
\begin{align*}
\mathbb{E}\left[\left\langle \bm{\theta}_n - \bm{\theta}, 
\nabla_{\bm{\theta}} \mathcal{L}_G(\bm{\theta}_n, \bm{w}_n) \right\rangle \right]
> 
2 \tilde{K}_1^2 a + 2 C_1 \tilde{K}_1 \beta^G + \frac{\epsilon_0}{2}.
\end{align*}
Lemma \ref{lem:1}, together with (C1) and (C2), ensures that, for all $n\in\mathbb{N}$,
\begin{align}\label{ineq:1}
\mathbb{E} \left[ \left\|\bm{\theta}_{n+1} - \bm{\theta} \right\|^2 \right]
\leq \mathbb{E} \left[ \left\| \bm{\theta}_n - \bm{\theta} \right\|^2 \right] 
 + 2 a 
 \left( 
 \mathbb{E}\left[\left\langle \bm{\theta} - \bm{\theta}_n, 
\nabla_{\bm{\theta}} \mathcal{L}_G(\bm{\theta}_n, \bm{w}_n) \right\rangle \right] + 2 C_1 \tilde{K}_1 \beta^{G}
 \right)
 + 4 \tilde{K}_1^2 a^2.
\end{align}
Accordingly, for all $n \geq n_0$,
\begin{align*}
\mathbb{E} \left[ \left\|\bm{\theta}_{n+1} - \bm{\theta} \right\|^2 \right]
&< \mathbb{E} \left[ \left\| \bm{\theta}_n - \bm{\theta} \right\|^2 \right] 
 - 2a
 \left(
 2 \tilde{K}_1^2 a + 2 C_1 \tilde{K}_1 \beta^G 
 + \frac{\epsilon_0}{2}
 \right) 
 + 4 a C_1 \tilde{K}_1 \beta^{G}
 + 4 \tilde{K}_1^2 a^2\\
&\quad 
= 
\mathbb{E} \left[ \left\| \bm{\theta}_n - \bm{\theta} \right\|^2 \right]
- 
a \epsilon_0\\
&\quad 
<
\mathbb{E} \left[ \left\| \bm{\theta}_{n_0} - \bm{\theta} \right\|^2 \right]
- 
a \epsilon_0 (n+1-n_0).
\end{align*}
$\mathbb{E} [ \| \bm{\theta}_{n_0} - \bm{\theta}\|^2] - a \epsilon_0(n+1-n_0)$ approaches minus infinity as $n$ approaches positive infinity. This is a contradiction. Hence, \eqref{inf} holds. Since $\epsilon > 0$ is arbitrary, we have that 
\begin{align}\label{ineq:inf}
\liminf_{n \to + \infty} \mathbb{E}\left[\left\langle \bm{\theta}_n - \bm{\theta}, 
\nabla_{\bm{\theta}} \mathcal{L}_G(\bm{\theta}_n, \bm{w}_n) \right\rangle \right]
\leq 2 \tilde{K}_1^2 a + 2 C_1 \tilde{K}_1 \beta^G.
\end{align}
A discussion similar to the one showing the above inequality leads to the following: 
\begin{align}\label{ineq:inf_2}
\liminf_{n \to + \infty} \mathbb{E}\left[\left\langle \bm{w}_n - \bm{w}, 
\nabla_{\bm{w}} \mathcal{L}_D(\bm{\theta}_n, \bm{w}_n) \right\rangle \right]
\leq 2 \tilde{K}_2^2 b + 2 C_2 \tilde{K}_2 \beta^D.
\end{align}
Inequality \eqref{ineq:inf} ensures that there exists a subsequence $((\bm{\theta}_{n_i}, \bm{w}_{n_i}))_{i\in \mathbb{N}}$ of $((\bm{\theta}_{n}, \bm{w}_n))_{n\in\mathbb{N}}$ such that, for all $\bm{\theta} \in \mathbb{R}^\Theta$, 
\begin{align*}
\lim_{i \to + \infty} \mathbb{E}\left[\left\langle \bm{\theta}_{n_i} - \bm{\theta},
\nabla_{\bm{\theta}} \mathcal{L}_G(\bm{\theta}_{n_i}, \bm{w}_{n_i}) \right\rangle \right]
=
\liminf_{n \to + \infty} \mathbb{E}\left[\left\langle \bm{\theta}_n - \bm{\theta},
\nabla_{\bm{\theta}} \mathcal{L}_G(\bm{\theta}_n, \bm{w}_n) \right\rangle \right]
\leq 2 \tilde{K}_1^2 a + 2 C_1 \tilde{K}_1 \beta^G.
\end{align*}
Assumption {(C3)} implies that there exists $((\bm{\theta}_{n_{i_j}}, \bm{w}_{n_{i_j}}))_{j\in \mathbb{N}}$ of $((\bm{\theta}_{n_i}, \bm{w}_{n_i}))_{i\in\mathbb{N}}$ such that $((\bm{\theta}_{n_{i_j}}, \bm{w}_{n_{i_j}}))_{j\in \mathbb{N}}$ converges almost surely to $(\bm{\theta}^\star, \bm{w}^\star)$. Accordingly, for all $\bm{\theta} \in \mathbb{R}^\Theta$, 
\begin{align*}
\mathbb{E}\left[\left\langle \bm{\theta}^\star - \bm{\theta},
\nabla_{\bm{\theta}} \mathcal{L}_G(\bm{\theta}^\star, \bm{w}^\star) \right\rangle \right]
\leq 2 \tilde{K}_1^2 a + 2 C_1 \tilde{K}_1 \beta^G,
\end{align*}
which implies that 
\begin{align}\label{1}
\mathbb{E}\left[\left\| 
\nabla_{\bm{\theta}} \mathcal{L}_G(\bm{\theta}^\star, \bm{w}^\star) \right\|^2 \right]
\leq 2 \tilde{K}_1^2 a + 2 C_1 \tilde{K}_1 \beta^G.
\end{align}
A discussion similar to the one showing the above inequality, together with \eqref{ineq:inf_2}, means that there exists a subsequence $((\bm{\theta}_{n_k}, \bm{w}_{n_k}))_{k\in \mathbb{N}}$ of $((\bm{\theta}_{n}, \bm{w}_n))_{n\in\mathbb{N}}$ such that $((\bm{\theta}_{n_k}, \bm{w}_{n_k}))_{k\in \mathbb{N}}$ converges almost surely to $(\bm{\theta}_\star, \bm{w}_\star)$ satisfying 
\begin{align}\label{2}
\mathbb{E}\left[\left\| 
\nabla_{\bm{w}} \mathcal{L}_D(\bm{\theta}_\star, \bm{w}_\star) \right\|^2 \right]
\leq 2 \tilde{K}_2^2 b + 2 C_2 \tilde{K}_2 \beta^D.
\end{align}

Inequality \eqref{ineq:1} implies that
\begin{align*}
\mathbb{E}\left[\left\langle \bm{\theta}_n - \bm{\theta}, 
\nabla_{\bm{\theta}} \mathcal{L}_G(\bm{\theta}_n, \bm{w}_n) \right\rangle \right]
\leq
\frac{1}{2a} \left\{ 
\mathbb{E} \left[ \left\| \bm{\theta}_n - \bm{\theta} \right\|^2 \right] 
-
\mathbb{E} \left[ \left\| \bm{\theta}_{n+1} - \bm{\theta} \right\|^2 \right] 
\right\}
+ 
2\tilde{K}_1^2 a
+ 2 C_1 \tilde{K}_1 \beta_n^{G}.
\end{align*}
Summing the above inequality from $n=1$ to $n = N$ guarantees that
\begin{align*}
&\sum_{n\in [N]} \mathbb{E}\left[\left\langle \bm{\theta}_n - \bm{\theta}, 
\nabla_{\bm{\theta}} \mathcal{L}_G(\bm{\theta}_n, \bm{w}_n) \right\rangle \right]\\
&\quad\leq
\frac{1}{2a} \left\{ 
\mathbb{E} \left[ \left\| \bm{\theta}_1 - \bm{\theta} \right\|^2 \right] 
-
\mathbb{E} \left[ \left\| \bm{\theta}_{N+1} - \bm{\theta} \right\|^2 \right] 
\right\}
+ 
2 \tilde{K}_1^2 N a
+ 2 C_1 \tilde{K}_1 \sum_{n\in [N]}\beta_n^{G}\\
&\quad\leq
\frac{1}{2a} 
\mathbb{E} \left[ \left\| \bm{\theta}_1 - \bm{\theta} \right\|^2 \right] 
+ 
2\tilde{K}_1^2 N a
+ 2 C_1 \tilde{K}_1 \sum_{n\in [N]}\beta_n^{G},
\end{align*}
which implies that
\begin{align}\label{ineq:e_d}
\frac{1}{N} \sum_{n\in [N]} \mathbb{E}\left[\left\langle \bm{\theta}_n - \bm{\theta}, 
\nabla_{\bm{\theta}} \mathcal{L}_G(\bm{\theta}_n, \bm{w}_n) \right\rangle \right]
\leq
\frac{1}{2a N} 
\mathbb{E} \left[ \left\| \bm{\theta}_1 - \bm{\theta} \right\|^2 \right] 
+ 
2 \tilde{K}_1^2 a
+ \frac{2 C_1 \tilde{K}_1}{N} \sum_{n\in [N]}\beta_n^{G}.
\end{align}
A discussion similar to the one showing \eqref{ineq:e_d} leads to the following:
\begin{align*}
&\frac{1}{N} \sum_{n\in [N]} \mathbb{E}\left[\left\langle \bm{w}_n - \bm{w}, 
\nabla_{\bm{w}} \mathcal{L}_D(\bm{\theta}_n, \bm{w}_n) \right\rangle \right]
\leq
\frac{1}{2b N} 
\mathbb{E} \left[ \left\| \bm{w}_1 - \bm{w} \right\|^2 \right] 
+ 
2\tilde{K}_2^2 b
+ \frac{2 C_2 \tilde{K}_2}{N} \sum_{n\in [N]}\beta_n^{D}.
\end{align*}
Let us consider the case where $((\bm{\theta}_n, \bm{w}_n))_{n\in\mathbb{N}}$ converges almost surely to $(\bm{\theta}^\star, \bm{w}^\star)$. Inequalities \eqref{1} and \eqref{2} indicate that
\begin{align*}
&\mathbb{E}\left[\left\| 
\nabla_{\bm{\theta}} \mathcal{L}_G(\bm{\theta}^\star, \bm{w}^\star) \right\|^2 \right]
\leq 2 \tilde{K}_1^2 a + 2 C_1 \tilde{K}_1 \beta^G,\\
&\mathbb{E}\left[\left\| 
\nabla_{\bm{w}} \mathcal{L}_D(\bm{\theta}^\star, \bm{w}^\star) \right\|^2 \right]
\leq 2 \tilde{K}_2^2 b + 2 C_2 \tilde{K}_2 \beta^D,
\end{align*}
which completes the proof. $\square$

\subsection{Proof outline of Theorem \ref{thm:2}} 
First, we provide a brief outline of the proof. The flow is the same as in \cite{BORKAR1997291}. The definitions of $\bm{\theta}_{n+1}$ and $\bm{w}_{n+1}$ imply that, for all $n\in \mathbb{N}$,
\begin{align}\label{error}
 \begin{split}
 &\bm{\theta}_{n+1} = \bm{\theta}_{n} - b_n \frac{a_n}{b_n} \mathcal{G}(\bm{\theta}_{n}, \bm{w}_n) + a_n \beta_n^G \bm{d}^G (\bm{\theta}_{n-1}, \bm{w}_{n-1}),\\
 &\bm{w}_{n+1} = \bm{w}_n - b_n \mathcal{D}(\bm{\theta}_{n}, \bm{w}_n) + b_n \beta_n^D \bm{d}^D (\bm{\theta}_{n-1}, \bm{w}_{n-1}).
 \end{split}
\end{align}
We can easily check that \eqref{error} is as a discretized version of the ordinary differential equations $\dot{\bm{x}}(t) = \bm{0}$ and $\dot{\bm{y}}(t) = \mathcal{D}(\bm{x}(t), \bm{y}(t))$ with a step size $b_n$ and errors at the $n$th iteration defined by 
\begin{align}\label{error_1}
 -\frac{a_n}{b_n} \mathcal{G}(\bm{\theta}_{n}, \bm{w}_n) + a_n \beta_n^G \bm{d}^G (\bm{\theta}_{n-1},\bm{w}_{n-1}) \text{ and }
 b_n \beta_n^D \bm{d}^D (\bm{\theta}_{n-1}, \bm{w}_{n-1}).
\end{align}
Assumptions \ref{const}(C2) {and} \ref{dim}(D2)(i) thus guarantee that two sequences defined by \eqref{error_1} converge almost surely to zero. Hence, an argument similar to the one for obtaining Theorem 1.1 in \cite{BORKAR1997291} (see \cite[Section 2]{BORKAR1997291} for the detailed proof) leads to Theorem \ref{thm:2}(i). Theorem \ref{thm:2}(ii) depends on the key inequalities \eqref{key:1} and \eqref{key:2}, which come from Assumptions \ref{assum:1}(A1)--(A2) and \ref{const}(C2)--{(C3)}. Summing \eqref{key:1} and \eqref{key:2} from $n=1$ to $n=N$ gives \eqref{R}. In particular, let us define $a_n = \mathcal{O}(n^{-\eta_a})$, $\beta_n^G = \mathcal{O}(n^{-\eta_a})$, $b_n = \mathcal{O}(n^{-\eta_b})$, and $\beta_n^D = \mathcal{O}(n^{-\eta_b})$, where $\eta_a, \eta_b \in (0,1)$. These learning rates satisfy Assumption \ref{dim}(D2)(ii) and {(D3)}. Hence, \eqref{R} implies \eqref{R_1}. {Moreover, we use $a_n$ and $b_n$ defined by \eqref{decaying_a_b}. These learning rates satisfy Assumption \ref{dim}(D2)(ii) and {(D3)}. Hence, \eqref{R} implies \eqref{R_1}.}

The following is the detailed proof of Theorem \ref{thm:2}.

{\em Proof of Theorem \ref{thm:2}:} (i) The definitions of $\bm{\theta}_{n+1}$ and $\bm{w}_{n+1}$ imply that, for all $n\in \mathbb{N}$,
\begin{align}\label{Error}
\begin{split}
 &\bm{\theta}_{n+1} = \bm{\theta}_{n} - b_n \frac{a_n}{b_n} \mathcal{G}(\bm{\theta}_{n}, \bm{w}_n) + a_n \beta_n^G \bm{d}^G (\bm{\theta}_{n-1}, \bm{w}_{n-1}),\\
 &\bm{w}_{n+1} = \bm{w}_n - b_n \mathcal{D}(\bm{\theta}_{n}, \bm{w}_n) + b_n \beta_n^D \bm{d}^D (\bm{\theta}_{n-1}, \bm{w}_{n-1}).
 \end{split}
\end{align}
Lemma \ref{lem:0} ensures that $(\|\mathcal{G}(\bm{\theta}_{n}, \bm{w}_n)\|)_{n\in \mathbb{N}}$, $(\|\mathcal{D}(\bm{\theta}_{n}, \bm{w}_n)\|)_{n\in \mathbb{N}}$, $(\bm{d}^G (\bm{\theta}_{n}, \bm{w}_{n}))_{n\in\mathbb{N}}$, and $(\bm{d}^D (\bm{\theta}_{n}, \bm{w}_{n}))_{n\in\mathbb{N}}$ are almost surely bounded. Assumption \ref{dim}(D1) ensures that \eqref{Error} can be regarded as a discretized version of the ordinary differential equations $\dot{\bm{x}}(t) = \bm{0}$ and $\dot{\bm{y}}(t) = \mathcal{D}(\bm{x}(t), \bm{y}(t))$ with a step size $b_n$ and errors at the $n$-th iteration defined by 
\begin{align}\label{Error_1}
 -\frac{a_n}{b_n} \mathcal{G}(\bm{\theta}_{n}, \bm{w}_n) + a_n \beta_n^G \bm{d}^G (\bm{\theta}_{n-1},\bm{w}_{n-1}) \text{ and }
 b_n \beta_n^D \bm{d}^D (\bm{\theta}_{n-1}, \bm{w}_{n-1}).
\end{align}
Assumptions \ref{const}(C2) and \ref{dim}(D2)(i) thus guarantee that two sequences defined by \eqref{Error_1} converge almost surely to zero. Hence, an argument similar to the one proving Theorem 1.1 in \cite{BORKAR1997291} (see \cite[Section 2]{BORKAR1997291} for the details of the proof) leads to Theorem \ref{thm:2}(i).

(ii) Lemma \ref{lem:1} guarantees that, for all $\bm{\theta} \in \mathbb{R}^\Theta$ and all $n\in\mathbb{N}$,
\begin{align*}
\mathbb{E}\left[\left\langle \bm{\theta}_n - \bm{\theta}, 
\nabla_{\bm{\theta}} \mathcal{L}_G(\bm{\theta}_n, \bm{w}_n) \right\rangle \right]
\leq 
\frac{1}{2 a_n}
\left\{
\mathbb{E} \left[ \left\| \bm{\theta}_n - \bm{\theta} \right\|^2 \right] 
-
\mathbb{E} \left[ \left\| \bm{\theta}_{n+1} - \bm{\theta} \right\|^2 \right]
\right\}
+ 2 C_1 \tilde{K}_1 \beta_n^{G}
+ 
2 \tilde{K}_1^2 a_n.
\end{align*}
Summing the above inequality from $n=1$ to $n = N$ gives, for all $\bm{\theta} \in \mathbb{R}^\Theta$,
\begin{align*}
&\sum_{n\in [N]} \mathbb{E}\left[\left\langle \bm{\theta}_n - \bm{\theta}, 
\nabla_{\bm{\theta}} \mathcal{L}_G(\bm{\theta}_n, \bm{w}_n) \right\rangle \right]\\
&\quad \leq 
\sum_{n\in [N]}\frac{1}{2 a_n}
\left\{
\mathbb{E} \left[ \left\| \bm{\theta}_n - \bm{\theta} \right\|^2 \right] 
-
\mathbb{E} \left[ \left\| \bm{\theta}_{n+1} - \bm{\theta} \right\|^2 \right]
\right\}
+ 2 C_1 \tilde{K}_1 \sum_{n\in [N]} \beta_n^{G} 
+ 
2 \tilde{K}_1^2 \sum_{n\in [N]} a_n.
\end{align*}
We have that
\begin{align*}
&\sum_{n\in [N]}\frac{1}{a_n}
\left\{
\mathbb{E} \left[ \left\| \bm{\theta}_n - \bm{\theta} \right\|^2 \right] 
-
\mathbb{E} \left[ \left\| \bm{\theta}_{n+1} - \bm{\theta} \right\|^2 \right]
\right\}\\
&\quad = 
\frac{\mathbb{E} \left[ \left\| \bm{\theta}_1 - \bm{\theta} \right\|^2 \right]}{a_1}
+ 
\sum_{k=2}^N 
\left\{
\frac{\mathbb{E} \left[ \left\| \bm{\theta}_k - \bm{\theta} \right\|^2 \right]}{a_k} 
-
\frac{\mathbb{E} \left[ \left\| \bm{\theta}_{k} - \bm{\theta} \right\|^2 \right]}{a_{k-1}}
\right\} 
- \frac{\mathbb{E} \left[ \left\| \bm{\theta}_{N+1} - \bm{\theta} \right\|^2 \right]}{a_{N}}\\
&\quad \leq
\frac{\mathbb{E} \left[ \left\| \bm{\theta}_1 - \bm{\theta} \right\|^2 \right]}{a_1}
+ 
\mathbb{E}\left[
\sum_{k=2}^N \sum_{i\in [\Theta]}
\left(
\frac{1}{a_k} - \frac{1}{a_{k-1}}
\right)
(\theta_{k,i} - \theta_i)^2
\right].
\end{align*}
Since $(a_n)_{n\in\mathbb{N}}$ is monotone decreasing, we have that, for all $k \geq 1$, $a_k^{-1} - a_{k-1}^{-1} \geq 0$. Moreover, for all $k \in \mathbb{N}$, $\sum_{i\in [\Theta]}(\theta_{k,i} - \theta_i)^2 \leq \|\bm{\theta}_k - \bm{\theta}\|^2 \leq C_1^2$. Hence, 
\begin{align*}
\sum_{n\in [N]}\frac{1}{a_n}
\left\{
\mathbb{E} \left[ \left\| \bm{\theta}_n - \bm{\theta} \right\|^2 \right] 
-
\mathbb{E} \left[ \left\| \bm{\theta}_{n+1} - \bm{\theta} \right\|^2 \right]
\right\}
\leq
\frac{C_1^2}{a_1}
+ 
C_1^2
\left(
\frac{1}{a_N} - \frac{1}{a_{1}}
\right)
= 
\frac{C_1^2}{a_N}.
\end{align*}
Therefore, for all $\bm{\theta} \in \mathbb{R}^\Theta$ and all $N \geq 1$,
\begin{align*}
\frac{1}{N} \sum_{n\in [N]} \mathbb{E}\left[\left\langle \bm{\theta}_n - \bm{\theta}, 
\nabla_{\bm{\theta}} \mathcal{L}_G(\bm{\theta}_n, \bm{w}_n) \right\rangle \right]
\leq 
\frac{C_1^2}{2 a_N N}
+ \frac{2 C_1 \tilde{K}_1}{N} \sum_{n\in [N]} \beta_n^{G} 
+ 
\frac{2 \tilde{K}_1^2}{N} \sum_{n\in [N]} a_n.
\end{align*}
A discussion similar to the one showing the above inequality leads to the following: for all $\bm{w} \in \mathbb{R}^W$ and all $N \geq 1$,
\begin{align*}
&\frac{1}{N} \sum_{n\in [N]} \mathbb{E}\left[\left\langle \bm{w}_n - \bm{w}, 
\nabla_{\bm{w}} \mathcal{L}_D(\bm{\theta}_n, \bm{w}_n) \right\rangle \right]
\leq 
\frac{C_2^2}{2 b_N N}
+ \frac{2 C_2 \tilde{K}_2}{N} \sum_{n\in [N]} \beta_n^{D} 
+ 
\frac{2 \tilde{K}_2^2}{N} \sum_{n\in [N]} b_n.
\end{align*}
Let us define $\alpha_n := n^{-\eta}$, where $\eta \in (0,1)$. Then, we have
\begin{align*}
\frac{1}{\alpha_N N} = \frac{1}{N^{1-\eta}}
\end{align*}
and 
\begin{align*}
\frac{1}{N} \sum_{n\in [N]} \alpha_n
\leq 
\frac{1}{N} \left\{
1 + \int_1^N \frac{\mathrm{d}t}{t^\eta}
\right\} 
\leq
\frac{1}{N}\frac{N^{1-\eta}}{1-\eta}
= 
\frac{1}{(1-\eta)N^\eta}.
\end{align*}
Theorem \ref{thm:2}(ii) thus leads to 
\begin{align*}
&\frac{1}{N} \sum_{n\in [N]} \mathbb{E}\left[\left\langle \bm{\theta}_n - \bm{\theta}, 
\nabla_{\bm{\theta}} \mathcal{L}_G(\bm{\theta}_n, \bm{w}_n) \right\rangle \right]
\leq 
\mathcal{O}\left( \frac{1}{N^{\mu_a}} \right), \\
&
\frac{1}{N} \sum_{n\in [N]} \mathbb{E}\left[\left\langle \bm{w}_n - \bm{w}, 
\nabla_{\bm{w}} \mathcal{L}_D(\bm{\theta}_n, \bm{w}_n) \right\rangle \right] 
\leq
\mathcal{O}\left( \frac{1}{N^{\mu_b}} \right),
\end{align*}
where $\mu_a := \min \{ \eta_a, 1 - \eta_a \}$ and $\mu_b := \min \{ \eta_b, 1 - \eta_b \}$.

{Let us define 
\begin{align*}
(a_n) 
= (\underbrace{a,a,\ldots,a}_{T}, 
\underbrace{\gamma a, \gamma a, \ldots, \gamma a}_{T}, \ldots, \underbrace{\gamma^{P-1} a, \gamma^{P-1} a, \ldots, \gamma^{P-1} a}_{T}, \ldots),
\end{align*}
where $a, T >0$, $P \geq 1$, and $\gamma \in (0,1)$. Let $N = TP$ and $\underline{a} > 0$ such that $\underline{a} \leq a_N$. Then, we have 
\begin{align*}
 \frac{1}{\alpha_N N} \leq 
 \frac{1}{\underline{a} N}
 \text{ and }
 \sum_{n \in [N]} a_n 
 \leq \sum_{n=0}^{+\infty} a T \gamma^n = \frac{aT}{1 - \gamma}.
\end{align*}
When $\beta_n^G$ satisfies $\sum_{n=0}^{+\infty} \beta_n^G < + \infty$, Theorem \ref{thm:2}(ii) leads to 
\begin{align*}
&\frac{1}{N} \sum_{n\in [N]} \mathbb{E}\left[\left\langle \bm{\theta}_n - \bm{\theta}, 
\nabla_{\bm{\theta}} \mathcal{L}_G(\bm{\theta}_n, \bm{w}_n) \right\rangle \right]
\leq 
\mathcal{O}\left( \frac{1}{N} \right), \\
&
\frac{1}{N} \sum_{n\in [N]} \mathbb{E}\left[\left\langle \bm{w}_n - \bm{w}, 
\nabla_{\bm{w}} \mathcal{L}_D(\bm{\theta}_n, \bm{w}_n) \right\rangle \right] 
\leq
\mathcal{O}\left( \frac{1}{N} \right).
\end{align*}
}
This completes the proof. $\square$

\section{Examples of Algorithm \ref{algo:1}}
\subsection{Examples of Algorithm \ref{algo:1} with constant learning rates}
\label{subsub:3.1.1}
Properties \eqref{sum_g}, \eqref{sum_d}, and \eqref{ane} in Theorem \ref{thm:1} indicate that, for sufficiently small learning rates $a$ and $b$ and CG parameters $\beta^G$ and $\beta^D$, 
\begin{align*}
\mathbb{E}\left[ \left\|
\nabla_{\bm{\theta}} \mathcal{L}_G(\bm{\theta}^\star, \bm{w}^\star)
\right\|^2 \right] 
\approx 0, \text{ } 
\mathbb{E}\left[ \left\|
\nabla_{\bm{w}} \mathcal{L}_D(\bm{\theta}^\star, \bm{w}^\star)
\right\|^2 \right] 
\approx 0
\end{align*} 
with 
\begin{align}\label{n}
\begin{split}
\frac{1}{N} \sum_{n\in [N]} \mathbb{E}\left[\left\langle \bm{\theta}_n - \bm{\theta}, 
\nabla_{\bm{\theta}} \mathcal{L}_G(\bm{\theta}_n, \bm{w}_n) \right\rangle \right] 
\approx
\mathcal{O}\left( \frac{1}{N} \right),\\
\frac{1}{N} \sum_{n\in [N]} \mathbb{E}\left[\left\langle \bm{w}_n - \bm{w}, 
\nabla_{\bm{w}} \mathcal{L}_D(\bm{\theta}_n, \bm{w}_n) \right\rangle \right] 
\approx
\mathcal{O}\left( \frac{1}{N} \right). 
\end{split}
\end{align}

Let us discuss the results in Theorem \ref{thm:1} under certain conditions. For simplicity, we assume that $((\bm{\theta}_n,\bm{w}_n))_{n\in\mathbb{N}}$ generated by Algorithm \ref{algo:1} converges almost surely to $(\bm{\theta}^\star, \bm{w}^\star)$.

(I) [SGD] First, we consider the case where $\beta^G = \beta^D = 0$. In this case, Algorithm \ref{algo:1} for the generator (resp. the discriminator) is based on the SGD method with a learning rate $a$ (resp. $b$). Theorem \ref{thm:1} indicates that the convergent point of the SGD-type algorithm satisfies 
\begin{align*}
\mathbb{E}\left[ \left\|
\nabla_{\bm{\theta}} \mathcal{L}_G(\bm{\theta}^\star, \bm{w}^\star) 
\right\|^2 \right] 
\leq 
2\tilde{K}_1^2 a, \text{ } 
\mathbb{E}\left[ \left\|
\nabla_{\bm{w}} \mathcal{L}_D(\bm{\theta}^\star, \bm{w}^\star) 
\right\|^2 \right] 
\leq 
2\tilde{K}_2^2 b.
\end{align*} 

(II) [Momentum SGD] We consider the case where $\beta_n^G := \beta^G \in [0,1/2]$ and $\beta_n^D := \beta^D \in [0,1/2]$. In this case, Algorithm \ref{algo:1} for the generator (resp. the discriminator) is based on the momentum method with a learning rate $a$ (resp. $b$) and momentum coefficient $a \beta^G$ (resp. $b \beta^D$). Theorem \ref{thm:1} indicates that the convergent point of the momentum-type algorithm satisfies 
\begin{align}\label{moment_1}
\mathbb{E}\left[ \left\|
\nabla_{\bm{\theta}} \mathcal{L}_G(\bm{\theta}^\star, \bm{w}^\star) 
\right\|^2 \right] 
\leq 
2 \tilde{K}_1^2 a + 2 C_1 \tilde{K}_1 \beta^G, \text{ } 
\mathbb{E}\left[ \left\|
\nabla_{\bm{w}} \mathcal{L}_D(\bm{\theta}^\star, \bm{w}^\star) 
\right\|^2 \right] 
\leq 
2 \tilde{K}_2^2 b + 2 C_2 \tilde{K}_2 \beta^D.
\end{align} 

(III) [CG] We consider the case where $\beta_n^G$ and $\beta_n^D$ are based on the CG formulas defined in Section \ref{cgm}. The parameters $\beta_n^G$ and $\beta_n^D$ satisfying (C2) are, for example, as follows:
\begin{align}\label{CG_G}
\beta_n^G 
= 
\begin{cases}
\beta_n^{\mathrm{FR},G}
=
\begin{cases}
\min \left\{ \frac{\|\mathcal{G}_n \|^2}{\|\mathcal{G}_{n-1} \|^2}, \frac{1}{2} \right\} 
&\text{ if } \|\mathcal{G}_{n-1} \| \neq 0,\\
0
&\text{ otherwise}
\end{cases}\\
\beta_n^{\mathrm{PRP},G}
=
\begin{cases}
\min \left\{ \frac{\langle \mathcal{G}_{n-1}, \mathcal{G}_{n} - \mathcal{G}_{n-1} \rangle}{\|\mathcal{G}_{n-1} \|^2}, \frac{1}{2} \right\} 
&\text{ if } \|\mathcal{G}_{n-1} \| \neq 0,\\
0
&\text{ otherwise}
\end{cases}\\
\beta_n^{\mathrm{HS},G}
=
\begin{cases}
\min \left\{ \frac{\langle \mathcal{G}_{n-1}, \mathcal{G}_{n} - \mathcal{G}_{n-1} \rangle}{\langle \bm{d}_{n-1}^G, \mathcal{G}_{n} - \mathcal{G}_{n-1} \rangle}, \frac{1}{2} \right\} 
&\text{ if } \langle \bm{d}_{n-1}^G, \mathcal{G}_{n} - \mathcal{G}_{n-1} \rangle \neq 0,\\
0
&\text{ otherwise}
\end{cases}\\
\beta_n^{\mathrm{DY},G}
=
\begin{cases}
\min \left\{ \frac{\| \mathcal{G}_{n} \|^2}{\langle \bm{d}_{n-1}^G, \mathcal{G}_{n} - \mathcal{G}_{n-1} \rangle}, \frac{1}{2} \right\} 
&\text{ if } \langle \bm{d}_{n-1}^G, \mathcal{G}_{n} - \mathcal{G}_{n-1} \rangle \neq 0,\\
0
&\text{ otherwise}
\end{cases}\\
\beta_n^{\mathrm{HZ},G}
=
\begin{cases}
\min \left\{ \beta_n^{\mathrm{HS},G} - \mu \frac{\| \mathcal{G}_{n} - \mathcal{G}_{n-1} \|^2 
\langle \mathcal{G}_{n}, \bm{d}_{n-1}^G \rangle}{\langle \bm{d}_{n-1}^G, \mathcal{G}_{n} - \mathcal{G}_{n-1} \rangle^2}, \frac{1}{2} \right\} 
&\text{ if } \langle \bm{d}_{n-1}^G, \mathcal{G}_{n} - \mathcal{G}_{n-1} \rangle \neq 0,\\
0
&\text{ otherwise}
\end{cases}\\
\beta_n^{\mathrm{Hyb1},G} 
= 
\max \left\{ 0, \min \left\{ \beta_n^{\mathrm{HS},G}, \beta_n^{\mathrm{DY},G} \right\} \right\}\\
\beta_n^{\mathrm{Hyb2},G}
= 
\max \left\{ 0, \min \left\{ \beta_n^{\mathrm{FR},G}, \beta_n^{\mathrm{PRP},G} \right\} \right\}
\end{cases}
\end{align}
where $\mathcal{G}_n := \mathcal{G}(\bm{\theta}_{n}, \bm{w}_{n})$, $\bm{d}_n^G := \bm{d}^G (\bm{\theta}_{n}, \bm{w}_{n})$, and $\mu > 1/4$.
\begin{align}\label{CG_D}
\beta_n^D
= 
\begin{cases}
\beta_n^{\mathrm{FR},D}
=
\begin{cases}
\min \left\{ \frac{\|\mathcal{D}_{n} \|^2}{\|\mathcal{D}_{n-1} \|^2}, \frac{1}{2} \right\} 
&\text{ if } \|\mathcal{D}_{n-1} \| \neq 0,\\
0
&\text{ otherwise}
\end{cases}\\
\beta_n^{\mathrm{PRP},D}
=
\begin{cases}
\min \left\{ \frac{\langle \mathcal{D}_{n-1}, \mathcal{D}_{n} - \mathcal{D}_{n-1} \rangle}{\|\mathcal{D}_{n-1} \|^2}, \frac{1}{2} \right\} 
&\text{ if } \|\mathcal{D}_{n-1} \| \neq 0,\\
0
&\text{ otherwise}
\end{cases}\\
\beta_n^{\mathrm{HS},D}
=
\begin{cases}
\min \left\{ \frac{\langle \mathcal{D}_{n-1}, \mathcal{D}_{n} - \mathcal{D}_{n-1} \rangle}{\langle \bm{d}^D_{n-1}, \mathcal{D}_{n} - \mathcal{D}_{n-1} \rangle}, \frac{1}{2} \right\} 
&\text{ if } \langle \bm{d}^D_{n-1}, \mathcal{D}_{n} - \mathcal{D}_{n-1} \rangle \neq 0,\\
0
&\text{ otherwise}
\end{cases}\\
\beta_n^{\mathrm{DY},D}
=
\begin{cases}
\min \left\{ \frac{\| \mathcal{D}_{n} \|^2}{\langle \bm{d}_{n-1}^D, \mathcal{D}_{n} - \mathcal{D}_{n-1} \rangle}, \frac{1}{2} \right\} 
&\text{ if } \langle \bm{d}_{n-1}^D, \mathcal{D}_{n} - \mathcal{D}_{n-1} \rangle \neq 0,\\
0
&\text{ otherwise}
\end{cases}\\
\beta_n^{\mathrm{HZ},D}
=
\begin{cases}
\min \left\{ \beta_n^{\mathrm{HS},D} - \mu \frac{\| \mathcal{D}_{n} - \mathcal{D}_{n-1} \|^2 
\langle \mathcal{D}_{n}, \bm{d}_{n-1}^D \rangle}{\langle \bm{d}_{n-1}^D, \mathcal{D}_{n} - \mathcal{D}_{n-1} \rangle^2}, \frac{1}{2} \right\} 
&\text{ if } \langle \bm{d}_{n-1}^D, \mathcal{D}_{n} - \mathcal{D}_{n-1} \rangle \neq 0,\\
0
&\text{ otherwise}
\end{cases}\\
\beta_n^{\mathrm{Hyb1},D} 
= 
\max \left\{ 0, \min \left\{ \beta_n^{\mathrm{HS},D}, \beta_n^{\mathrm{DY},D} \right\} \right\}\\
\beta_n^{\mathrm{Hyb2},D}
= 
\max \left\{ 0, \min \left\{ \beta_n^{\mathrm{FR},D}, \beta_n^{\mathrm{PRP},D} \right\} \right\}
\end{cases}
\end{align}
where $\mathcal{D}_{n} := \mathcal{D}(\bm{\theta}_{n}, \bm{w}_{n})$, $\bm{d}_n^D := \bm{d}^D (\bm{\theta}_{n}, \bm{w}_{n})$, and $\mu > 1/4$. Theorem \ref{thm:1} indicates that the convergent point of the CG-type algorithm with $\beta_n^G$ and $\beta_n^D$ defined by \eqref{CG_G} and \eqref{CG_D} satisfies \eqref{moment_1}.

\subsection{Examples of Algorithm \ref{algo:1} with diminishing learning rates}
\label{subsub:3.2.1}
Property \eqref{R_1} together with\footnote{The maximum value of $\min\{ \eta, 1 - \eta \}$ for $\eta \in (0,1)$ is $1/2$ when $\eta = 1/2$.} 
\begin{align}\label{ab}
\eta_a = \eta_b = \frac{1}{2}
\end{align}
indicates that Algorithm \ref{algo:1} satisfies
\begin{align}\label{improve}
&\frac{1}{N} \sum_{n\in [N]} \mathbb{E}\left[\left\langle \bm{\theta}_n - \bm{\theta}, 
\nabla_{\bm{\theta}} \mathcal{L}_G(\bm{\theta}_n, \bm{w}_n) \right\rangle \right]
=
\mathcal{O}\left( \frac{1}{\sqrt{N}} \right), \text{ } 
\frac{1}{N} \sum_{n\in [N]} \mathbb{E}\left[\left\langle \bm{w}_n - \bm{w}, 
\nabla_{\bm{w}} \mathcal{L}_D(\bm{\theta}_n, \bm{w}_n) \right\rangle \right] 
=
\mathcal{O}\left( \frac{1}{\sqrt{N}} \right),
\end{align}
where we assume that the stochastic gradient errors $M^{(\bm{\theta})}$ and $M^{(\bm{w})}$ are zero. To guarantee that Algorithm \ref{algo:1} converges almost surely to a stationary Nash equilibrium, we need to set diminishing learning rates $a_n$ and $b_n$ satisfying Assumption \ref{dim}(D2)(i), i.e., 
\begin{align}\label{ab_1}
\frac{1}{2} < \eta_b < \eta_a < 1
\end{align}
(see Theorem \ref{thm:2}(i)). Then, property \eqref{R_1} ensures that Algorithm \ref{algo:1} has the following convergence rate:
\begin{align}\label{rate}
&\frac{1}{N} \sum_{n\in [N]} \mathbb{E}\left[\left\langle \bm{\theta}_n - \bm{\theta}, 
\nabla_{\bm{\theta}} \mathcal{L}_G(\bm{\theta}_n, \bm{w}_n) \right\rangle \right]
= 
\mathcal{O}\left( \frac{1}{N^{\mu_a}} \right), \text{ } \frac{1}{N} \sum_{n\in [N]} \mathbb{E}\left[\left\langle \bm{w}_n - \bm{w}, 
\nabla_{\bm{w}} \mathcal{L}_D(\bm{\theta}_n, \bm{w}_n) \right\rangle \right]
=
\mathcal{O}\left( \frac{1}{N^{\mu_b}} \right),
\end{align}
where $\mu_a := \min\{ \eta_a, 1 - \eta_a \}$ and $\mu_b := \min\{ \eta_b, 1 - \eta_b \}$.

Let us discuss the results in Theorem \ref{thm:2} under certain conditions. 

(I) [SGD] First, we consider the case where $\beta^G = \beta^D = 0$. In this case, Algorithm \ref{algo:1} for the generator (resp. the discriminator) is based on the SGD method with a learning rate $a_n$ (resp. $b_n$). This algorithm was presented in \cite[(1)]{Heusel2017}. Theorem \ref{thm:2}(ii) indicates that the SGD-type algorithm satisfies \eqref{improve} and \eqref{rate}; i.e., 
\begin{align*}
&\frac{1}{N} \sum_{n\in [N]} \mathbb{E}\left[\left\langle \bm{\theta}_n - \bm{\theta}, 
\nabla_{\bm{\theta}} \mathcal{L}_G(\bm{\theta}_n, \bm{w}_n) \right\rangle \right]
=
\begin{cases}
\mathcal{O}\left( \frac{1}{\sqrt{N}} \right) \text{ if \eqref{ab} holds}, \\
\mathcal{O}\left( \frac{1}{N^{\mu_a}} \right) \text{ if \eqref{ab_1} holds}, 
\end{cases}\\
&\frac{1}{N} \sum_{n\in [N]} \mathbb{E}\left[\left\langle \bm{w}_n - \bm{w}, 
\nabla_{\bm{w}} \mathcal{L}_D(\bm{\theta}_n, \bm{w}_n) \right\rangle \right]
=
\begin{cases}
\mathcal{O}\left( \frac{1}{\sqrt{N}} \right) \text{ if \eqref{ab} holds}, \\
\mathcal{O}\left( \frac{1}{N^{\mu_b}} \right) \text{ if \eqref{ab_1} holds}.
\end{cases}
\end{align*}

(II) [Momentum SGD] Let us consider Algorithm \ref{algo:1} for the generator (resp. the discriminator) based on the momentum method with a learning rate $a_n$ (resp. $b_n$) and momentum coefficient $a_n \beta_n^G$ (resp. $b_n \beta_n^D$). Let us consider $\beta_n^G$ and $\beta_n^D$ with \eqref{ab_1}. Then, there exists $n_1 \in \mathbb{N}$ such that, for all $n \geq n_1$, $\beta_n^G, \beta_n^D \leq 1/2$, which implies that $\beta_n^G$ and $\beta_n^D$ satisfy (C2). Theorem \ref{thm:2}(ii) thus ensures that the momentum-type algorithm has the same convergence rate as in (I).

(III) [CG] Let us consider the CG-type algorithm, i.e., Algorithm \ref{algo:1} with $a_n$ and $b_n$ with \eqref{ab} and $\beta_n^G$ and $\beta_n^D$ defined by \eqref{CG_G} and \eqref{CG_D}. Theorem \ref{thm:2}(i) guarantees that the CG-type algorithm converges almost surely to a point in $\mathrm{LNE}(\mathcal{L}_D, \mathcal{L}_G)$. Next, let us consider $\beta_n^G$ and $\beta_n^D$ defined as follows:
\begin{align}\label{CG_G_dim}
\beta_n^G 
= 
\begin{cases}
\beta_n^{\mathrm{FR},G}
=
\begin{cases}
\min \left\{ \frac{\|\mathcal{G}_n \|^2}{\|\mathcal{G}_{n-1} \|^2}, \frac{1}{2} \right\} \frac{1}{n^{\eta_a}}
&\text{ if } \|\mathcal{G}_{n-1} \| \neq 0,\\
0
&\text{ otherwise}
\end{cases}\\
\beta_n^{\mathrm{PRP},G}
=
\begin{cases}
\min \left\{ \frac{\langle \mathcal{G}_{n-1}, \mathcal{G}_{n} - \mathcal{G}_{n-1} \rangle}{\|\mathcal{G}_{n-1} \|^2}, \frac{1}{2} \right\} \frac{1}{n^{\eta_a}}
&\text{ if } \|\mathcal{G}_{n-1} \| \neq 0,\\
0
&\text{ otherwise}
\end{cases}\\
\beta_n^{\mathrm{HS},G}
=
\begin{cases}
\min \left\{ \frac{\langle \mathcal{G}_{n-1}, \mathcal{G}_{n} - \mathcal{G}_{n-1} \rangle}{\langle \bm{d}_{n-1}^G, \mathcal{G}_{n} - \mathcal{G}_{n-1} \rangle}, \frac{1}{2} \right\} \frac{1}{n^{\eta_a}}
&\text{ if } \langle \bm{d}_{n-1}^G, \mathcal{G}_{n} - \mathcal{G}_{n-1} \rangle \neq 0,\\
0
&\text{ otherwise}
\end{cases}\\
\beta_n^{\mathrm{DY},G}
=
\begin{cases}
\min \left\{ \frac{\| \mathcal{G}_{n} \|^2}{\langle \bm{d}_{n-1}^G, \mathcal{G}_{n} - \mathcal{G}_{n-1} \rangle}, \frac{1}{2} \right\} \frac{1}{n^{\eta_a}}
&\text{ if } \langle \bm{d}_{n-1}^G, \mathcal{G}_{n} - \mathcal{G}_{n-1} \rangle \neq 0,\\
0
&\text{ otherwise}
\end{cases}\\
\beta_n^{\mathrm{HZ},G}
=
\begin{cases}
\min \left\{ \beta_n^{\mathrm{HS},G} - \mu \frac{\| \mathcal{G}_{n} - \mathcal{G}_{n-1} \|^2 
\langle \mathcal{G}_{n}, \bm{d}_{n-1}^G \rangle}{\langle \bm{d}_{n-1}^G, \mathcal{G}_{n} - \mathcal{G}_{n-1} \rangle^2}, \frac{1}{2} \right\} \frac{1}{n^{\eta_a}}
&\text{ if } \langle \bm{d}_{n-1}^G, \mathcal{G}_{n} - \mathcal{G}_{n-1} \rangle \neq 0,\\
0
&\text{ otherwise}
\end{cases}\\
\beta_n^{\mathrm{Hyb1},G} 
= 
\max \left\{ 0, \min \left\{ \beta_n^{\mathrm{HS},G}, \beta_n^{\mathrm{DY},G} \right\} \right\}\\
\beta_n^{\mathrm{Hyb2},G}
= 
\max \left\{ 0, \min \left\{ \beta_n^{\mathrm{FR},G}, \beta_n^{\mathrm{PRP},G} \right\} \right\}
\end{cases}
\end{align}
where $\mathcal{G}_n := \mathcal{G}(\bm{\theta}_{n}, \bm{w}_{n})$, $\bm{d}_n^G := \bm{d}^G (\bm{\theta}_{n}, \bm{w}_{n})$, and $\mu > 1/4$.
\begin{align}\label{CG_D_dim}
\beta_n^D
= 
\begin{cases}
\beta_n^{\mathrm{FR},D}
=
\begin{cases}
\min \left\{ \frac{\|\mathcal{D}_{n} \|^2}{\|\mathcal{D}_{n-1} \|^2}, \frac{1}{2} \right\} \frac{1}{n^{\eta_b}} 
&\text{ if } \|\mathcal{D}_{n-1} \| \neq 0,\\
0
&\text{ otherwise}
\end{cases}\\
\beta_n^{\mathrm{PRP},D}
=
\begin{cases}
\min \left\{ \frac{\langle \mathcal{D}_{n-1}, \mathcal{D}_{n} - \mathcal{D}_{n-1} \rangle}{\|\mathcal{D}_{n-1} \|^2}, \frac{1}{2} \right\} \frac{1}{n^{\eta_b}}
&\text{ if } \|\mathcal{D}_{n-1} \| \neq 0,\\
0
&\text{ otherwise}
\end{cases}\\
\beta_n^{\mathrm{HS},D}
=
\begin{cases}
\min \left\{ \frac{\langle \mathcal{D}_{n-1}, \mathcal{D}_{n} - \mathcal{D}_{n-1} \rangle}{\langle \bm{d}^D_{n-1}, \mathcal{D}_{n} - \mathcal{D}_{n-1} \rangle}, \frac{1}{2} \right\} \frac{1}{n^{\eta_b}}
&\text{ if } \langle \bm{d}^D_{n-1}, \mathcal{D}_{n} - \mathcal{D}_{n-1} \rangle \neq 0,\\
0
&\text{ otherwise}
\end{cases}\\
\beta_n^{\mathrm{DY},D}
=
\begin{cases}
\min \left\{ \frac{\| \mathcal{D}_{n} \|^2}{\langle \bm{d}_{n-1}^D, \mathcal{D}_{n} - \mathcal{D}_{n-1} \rangle}, \frac{1}{2} \right\} \frac{1}{n^{\eta_b}} 
&\text{ if } \langle \bm{d}_{n-1}^D, \mathcal{D}_{n} - \mathcal{D}_{n-1} \rangle \neq 0,\\
0
&\text{ otherwise}
\end{cases}\\
\beta_n^{\mathrm{HZ},D}
=
\begin{cases}
\min \left\{ \beta_n^{\mathrm{HS},D} - \mu \frac{\| \mathcal{D}_{n} - \mathcal{D}_{n-1} \|^2 
\langle \mathcal{D}_{n}, \bm{d}_{n-1}^D \rangle}{\langle \bm{d}_{n-1}^D, \mathcal{D}_{n} - \mathcal{D}_{n-1} \rangle^2}, \frac{1}{2} \right\} \frac{1}{n^{\eta_b}}
&\text{ if } \langle \bm{d}_{n-1}^D, \mathcal{D}_{n} - \mathcal{D}_{n-1} \rangle \neq 0,\\
0
&\text{ otherwise}
\end{cases}\\
\beta_n^{\mathrm{Hyb1},D} 
= 
\max \left\{ 0, \min \left\{ \beta_n^{\mathrm{HS},D}, \beta_n^{\mathrm{DY},D} \right\} \right\}\\
\beta_n^{\mathrm{Hyb2},D}
= 
\max \left\{ 0, \min \left\{ \beta_n^{\mathrm{FR},D}, \beta_n^{\mathrm{PRP},D} \right\} \right\}
\end{cases}
\end{align}
where $\mathcal{D}_{n} := \mathcal{D}(\bm{\theta}_{n}, \bm{w}_{n})$, $\bm{d}_n^D := \bm{d}^D (\bm{\theta}_{n}, \bm{w}_{n})$, and $\mu > 1/4$. The sequences $(\beta_n^G)_{n\in\mathbb{N}}$ and $(\beta_n^D)_{n\in\mathbb{N}}$ defined by \eqref{CG_G_dim} and \eqref{CG_D_dim} satisfy (C2) and (D4). Property \eqref{R_1} ensures that the CG-type algorithm has the the same convergence rate as in (I).

\newpage
\section{Experimental Settings}
\label{appendix:setting}
\subsection{Implementation and environment}
We performed our experiment on TSUBAME, a supercomputer owned by the Tokyo Institute of Technology. The total computation amounted to 65,900 GPU hours.

{
The dataset of CIFAR10\footnote{https://www.cs.toronto.edu/~kriz/cifar.html} does not indicate a license. The pre-trained model is under Apache License 2.0 for the FID inception weights\footnote{https://github.com/mseitzer/pytorch-fid/releases}.
}

Our code can be found at the link below.\\
\url{https://github.com/Hiroki11x/ConjugateGradient_GAN}

\subsection{Model and datasets}
{
We studied and evaluated SNGAN with ResNet generator \cite{DBLP:conf/nips/ZhuangTDTDPD20} for GAN experiments. We used CIFAR10 as the dataset. The workload for each experimental setup is shown in Table \ref{table:workloads}.
}

\begin{table}[H]
\centering
\caption{Workload}
\label{table:workloads}
\setlength{\tabcolsep}{0.5em} 
{\renewcommand{\arraystretch}{1.2}
\begin{tabular}[t]{lcccc}
\hline
Model & Dataset & Dataset Size & Step Budget\\
\hline \hline
SNGAN with ResNet Generator \cite{DBLP:conf/nips/ZhuangTDTDPD20} & CIFAR10 & 50000 & 100K\\ \hline

\end{tabular}\\ 
\hspace{10cm}
}
\end{table}

\subsection{Hyperparameters and detailed configuration}
\label{appendix:hyperparameters}
\subsubsection{Toy-example experiments}
The hyperparameters for the results shown in Figure \ref{fig:toy-exp1} of Section \ref{section:experiments-toy-example} are explained here. A momentum coefficient of 0.9 was used for momentum SGD, and FR was set as a beta update for the CG method. Each optimizer had a learning rate to update x and a learning rate to update y. For the constant learning rate settings, we set $3.75e-08$ as the learning rate of x and $1e-06$ as the learning rate of y in SGD. Moreover, we set $3.75e-08$ and $1e-07$ in the momentum SGD and $2.5e-08$ and $5e-07$ in the CG method. The number of steps was 400. After updating x, the gradient of y was updated for the objective function, which used the updated x.

\subsubsection{GAN experiments}
{
Here, we report the hyperparameter's search space. For SGD, we only searched the learning rate $\eta$ and batch size $B$. For momentum SGD, was added a fixed value of 0.9 as a parameter to control the momentum $\gamma$. As for the diminishing learning rates, we performed experiments as described in Section \ref{section:experiments-dim}. 
}

\begin{landscape}
\begin{table}[H]
\caption{Hyperparameter Search Range: Constant Learning Rate, CIFAR10 on SNGAN with ResNet Generator}
\label{table:cifar10-hyperparams2}
\setlength{\tabcolsep}{0.2em} 
{\renewcommand{\arraystretch}{1.2}
\begin{tabular}{llllll}
\hline
Optimizer & $\beta$ Update Rule & $B$ & $a$ \footnotemark[1] (Learning rate for Generator) & $b$ \footnotemark[2] (Learning rate for Discriminator) & Learning rate Scheduler Type \\ 
\hline \hline
SGD & - & \{64\} & \begin{tabular}[c]{@{}l@{}}\{5e-5, 1e-4, 5e-4. 1e-3, 5e-3\}\end{tabular} & \begin{tabular}[c]{@{}l@{}}\{5e-5, 1e-4, 5e-4. 1e-3, 5e-3\}\end{tabular} & \begin{tabular}[c]{@{}l@{}}\{Constant\}\end{tabular} \\
Momentum SGD & - & \{64\} & \begin{tabular}[c]{@{}l@{}}\{5e-5, 1e-4, 5e-4. 1e-3, 5e-3\}\end{tabular} & \begin{tabular}[c]{@{}l@{}}\{5e-5, 1e-4, 5e-4. 1e-3, 5e-3\}\end{tabular} & \begin{tabular}[c]{@{}l@{}}\{Constant\}\end{tabular} \\
Adam & - & \{64\} & \begin{tabular}[c]{@{}l@{}}\{5e-5, 1e-4, 5e-4. 1e-3, 5e-3\}\end{tabular} & \begin{tabular}[c]{@{}l@{}}\{5e-5, 1e-4, 5e-4. 1e-3, 5e-3\}\end{tabular} & \begin{tabular}[c]{@{}l@{}}\{Constant\}\end{tabular} \\
\hline
CG & FR & \{64\} & \begin{tabular}[c]{@{}l@{}}\{5e-5, 1e-4, 5e-4. 1e-3, 5e-3\}\end{tabular} & \begin{tabular}[c]{@{}l@{}}\{5e-5, 1e-4, 5e-4. 1e-3, 5e-3\}\end{tabular} & \begin{tabular}[c]{@{}l@{}}\{Constant\}\end{tabular} \\
CG & PRP & \{64\} & \begin{tabular}[c]{@{}l@{}}\{5e-5, 1e-4, 5e-4. 1e-3, 5e-3\}\end{tabular} & \begin{tabular}[c]{@{}l@{}}\{5e-5, 1e-4, 5e-4. 1e-3, 5e-3\}\end{tabular} & \begin{tabular}[c]{@{}l@{}}\{Constant\}\end{tabular} \\
CG & HS & \{64\} & \begin{tabular}[c]{@{}l@{}}\{5e-5, 1e-4, 5e-4. 1e-3, 5e-3\}\end{tabular} & \begin{tabular}[c]{@{}l@{}}\{5e-5, 1e-4, 5e-4. 1e-3, 5e-3\}\end{tabular} & \begin{tabular}[c]{@{}l@{}}\{Constant\}\end{tabular} \\
CG & DY & \{64\} & \begin{tabular}[c]{@{}l@{}}\{5e-5, 1e-4, 5e-4. 1e-3, 5e-3\}\end{tabular} & \begin{tabular}[c]{@{}l@{}}\{5e-5, 1e-4, 5e-4. 1e-3, 5e-3\}\end{tabular} & \begin{tabular}[c]{@{}l@{}}\{Constant\}\end{tabular} \\
CG & HZ & \{64\} & \begin{tabular}[c]{@{}l@{}}\{5e-5, 1e-4, 5e-4. 1e-3, 5e-3\}\end{tabular} & \begin{tabular}[c]{@{}l@{}}\{5e-5, 1e-4, 5e-4. 1e-3, 5e-3\}\end{tabular} & \begin{tabular}[c]{@{}l@{}}\{Constant\}\end{tabular} \\
CG & Hyb1 (HS, DY) & \{64\} & \begin{tabular}[c]{@{}l@{}}\{5e-5, 1e-4, 5e-4. 1e-3, 5e-3\}\end{tabular} & \begin{tabular}[c]{@{}l@{}}\{5e-5, 1e-4, 5e-4. 1e-3, 5e-3\}\end{tabular} & \begin{tabular}[c]{@{}l@{}}\{Constant\}\end{tabular} \\
CG & Hyb2 (FR, PRP) & \{64\} & \begin{tabular}[c]{@{}l@{}}\{5e-5, 1e-4, 5e-4. 1e-3, 5e-3\}\end{tabular} & \begin{tabular}[c]{@{}l@{}}\{5e-5, 1e-4, 5e-4. 1e-3, 5e-3\}\end{tabular} & \begin{tabular}[c]{@{}l@{}}\{Constant\}\end{tabular}
\\\hline
\end{tabular}
}
\end{table}

\begin{table}[H]
\caption{Hyperparameter Search Range: Diminishing Learning Rate, CIFAR10 on SNGAN with ResNet Generator}
\label{table:cifar10-hyperparams3}
\setlength{\tabcolsep}{0.2em} 
{\renewcommand{\arraystretch}{1.2}
\begin{tabular}{llllll}
\hline
Optimizer & $\beta$ Update Rule & $B$ & $a$ \footnotemark[1] (Learning rate for Generator) & $b$ \footnotemark[2] (Learning rate for Discriminator) & Learning rate Scheduler Type \\ 
\hline \hline
SGD & - & \{64\} & \begin{tabular}[c]{@{}l@{}}\{1e-4, 5e-4. 1e-3, 5e-3\}\end{tabular} & \begin{tabular}[c]{@{}l@{}}\{1e-4, 5e-4. 1e-3, 5e-3\}\end{tabular} & \begin{tabular}[c]{@{}l@{}}\{Diminishing\}\end{tabular} \\
Momentum SGD & - & \{64\} & \begin{tabular}[c]{@{}l@{}}\{1e-4, 5e-4. 1e-3, 5e-3\}\end{tabular} & \begin{tabular}[c]{@{}l@{}}\{1e-4, 5e-4. 1e-3, 5e-3\}\end{tabular} & \begin{tabular}[c]{@{}l@{}}\{Diminishing\}\end{tabular} \\
Adam & - & \{64\} & \begin{tabular}[c]{@{}l@{}}\{1e-4, 5e-4. 1e-3, 5e-3\}\end{tabular} & \begin{tabular}[c]{@{}l@{}}\{1e-4, 5e-4. 1e-3, 5e-3\}\end{tabular} & \begin{tabular}[c]{@{}l@{}}\{Diminishing\}\end{tabular} \\
\hline
CG & FR & \{64\} & \begin{tabular}[c]{@{}l@{}}\{1e-4, 5e-4. 1e-3, 5e-3\}\end{tabular} & \begin{tabular}[c]{@{}l@{}}\{1e-4, 5e-4. 1e-3, 5e-3\}\end{tabular} & \begin{tabular}[c]{@{}l@{}}\{Diminishing\}\end{tabular} \\
CG & PRP & \{64\} & \begin{tabular}[c]{@{}l@{}}\{1e-4, 5e-4. 1e-3, 5e-3\}\end{tabular} & \begin{tabular}[c]{@{}l@{}}\{1e-4, 5e-4. 1e-3, 5e-3\}\end{tabular} & \begin{tabular}[c]{@{}l@{}}\{Diminishing\}\end{tabular} \\
CG & HS & \{64\} & \begin{tabular}[c]{@{}l@{}}\{1e-4, 5e-4. 1e-3, 5e-3\}\end{tabular} & \begin{tabular}[c]{@{}l@{}}\{1e-4, 5e-4. 1e-3, 5e-3\}\end{tabular} & \begin{tabular}[c]{@{}l@{}}\{Diminishing\}\end{tabular} \\
CG & DY & \{64\} & \begin{tabular}[c]{@{}l@{}}\{1e-4, 5e-4. 1e-3, 5e-3\}\end{tabular} & \begin{tabular}[c]{@{}l@{}}\{1e-4, 5e-4. 1e-3, 5e-3\}\end{tabular} & \begin{tabular}[c]{@{}l@{}}\{Diminishing\}\end{tabular} \\
CG & HZ & \{64\} & \begin{tabular}[c]{@{}l@{}}\{1e-4, 5e-4. 1e-3, 5e-3\}\end{tabular} & \begin{tabular}[c]{@{}l@{}}\{1e-4, 5e-4. 1e-3, 5e-3\}\end{tabular} & \begin{tabular}[c]{@{}l@{}}\{Diminishing\}\end{tabular} \\
CG & Hyb1 (HS, DY) & \{64\} & \begin{tabular}[c]{@{}l@{}}\{1e-4, 5e-4. 1e-3, 5e-3\}\end{tabular} & \begin{tabular}[c]{@{}l@{}}\{1e-4, 5e-4. 1e-3, 5e-3\}\end{tabular} & \begin{tabular}[c]{@{}l@{}}\{Diminishing\}\end{tabular} \\
CG & Hyb2 (FR, PRP) & \{64\} & \begin{tabular}[c]{@{}l@{}}\{1e-4, 5e-4. 1e-3, 5e-3\}\end{tabular} & \begin{tabular}[c]{@{}l@{}}\{1e-4, 5e-4. 1e-3, 5e-3\}\end{tabular} & \begin{tabular}[c]{@{}l@{}}\{Diminishing\}\end{tabular}
\\\hline
\end{tabular}
}
\end{table}

\footnotetext[1]{$a$: Initial learning rate for generator}
\footnotetext[2]{$b$: Initial learning rate for discriminator}

\end{landscape}

\newpage
\clearpage
\newpage
\section{Supplemental Experimental Results}
\label{appendix:additional-experiments}
This section provides information on experimental results omitted from the text.

{
Adam, a popular optimizer in the training of GANs, was compared. Here, when Adam is used to train GANs, it has been reported that setting $\beta_1$, the hyperparameter of Adam's first order moment, to 0.5 leads to stable learning \cite{radford2015unsupervised}, and it was used in \cite{Heusel2017}'s work. Therefore, we chose this setting.
}

\subsubsection{Constant learning rate rule}
\label{appendix:resnet-constant-experiments}
{
In the case of a constant learning rate, the discriminator's loss decreased steadily, increasing the generator's loss (Figure \ref{appendix_fig:res-exp-const-loss}). The CG method using FR as a $\beta$ update rule had the best FID score: the discriminator's loss decreased steadily, and the generator's loss increased, suggesting convergence to the saddle point. The norm transition for the gradient is shown in Figure \ref{appendix_fig:res-exp-const-norm}. The FID transition is shown in Figure \ref{appendix_fig:resnet-exp-const-fid}, and the final scores are shown in Table \ref{table:exp-fid-const}.
}

{
Additionally, the details of the FID scores for each learning-rate combination are shown in Figure \ref{additional_fig:res-lr_analysis_constlr_cifar10_bs64} in Appendix \ref{appendix:lr-sensitivity}. 
 }

\begin{figure}[H]
 \centering
 \includegraphics[width=\linewidth]{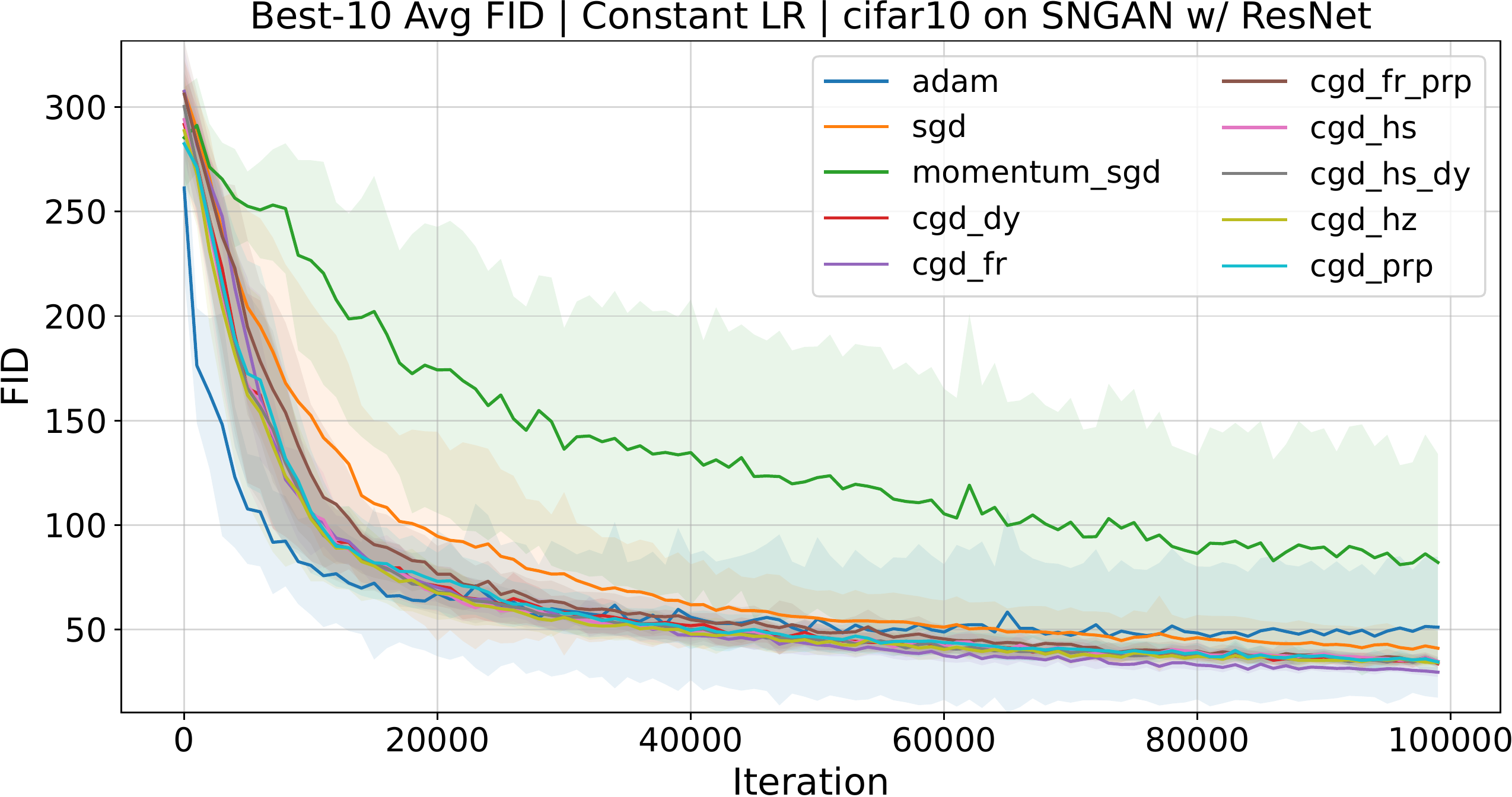}
\caption{Mean FID (solid line) bounded by the maximum and minimum (the shaded areas) over the best ten runs in the sense of FID when using a constant learning rate. Results for CIFAR-10 on SNGAN with ResNet generator.}
\label{appendix_fig:resnet-exp-const-fid}
\vspace{-3mm}
\end{figure}

\begin{figure}[H]
 \centering
 \includegraphics[width=0.48\linewidth]{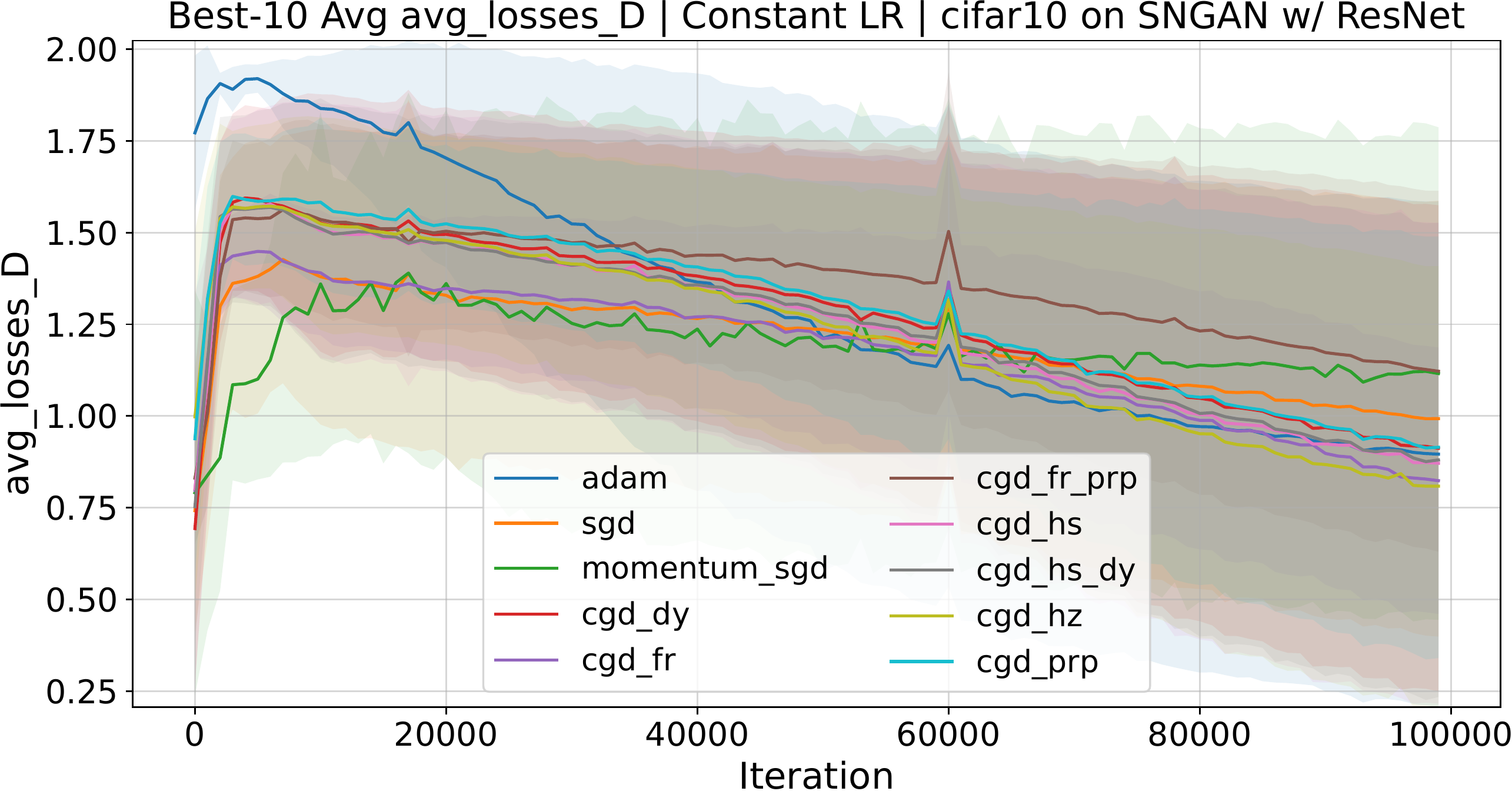}
 \includegraphics[width=0.48\linewidth]{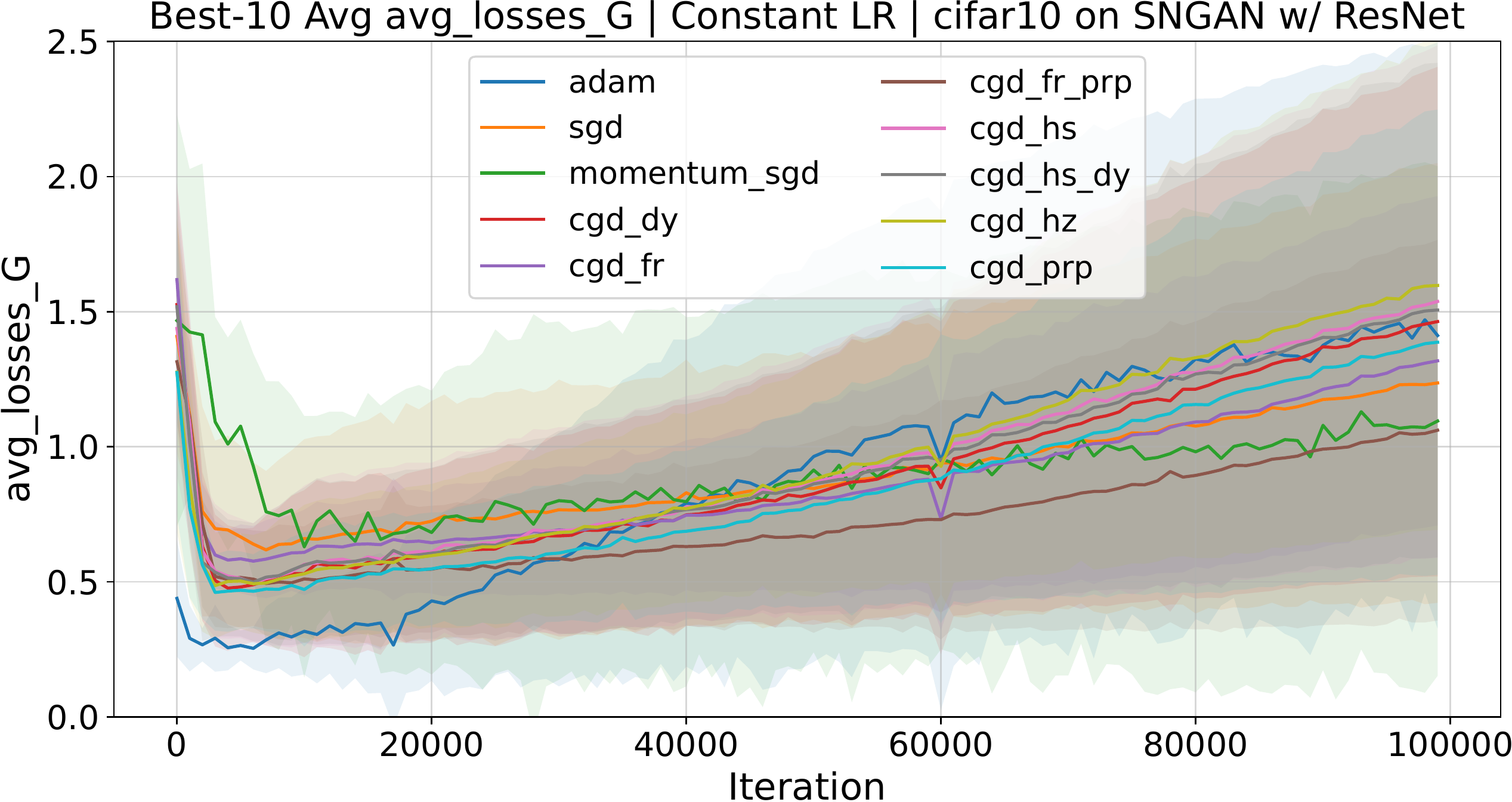}
\caption{Mean loss (solid line surrounded by the shaded areas) bounded by the maximum and minimum over the best ten runs in the sense of FID. \textbf{Left}: CIFAR-10 discriminator loss, \textbf{Right}: CIFAR-10 generator loss}
\label{appendix_fig:res-exp-const-loss}
\vspace{-3mm}
\end{figure}

\begin{figure}[H]

 \centering
 \includegraphics[width=0.48\linewidth]{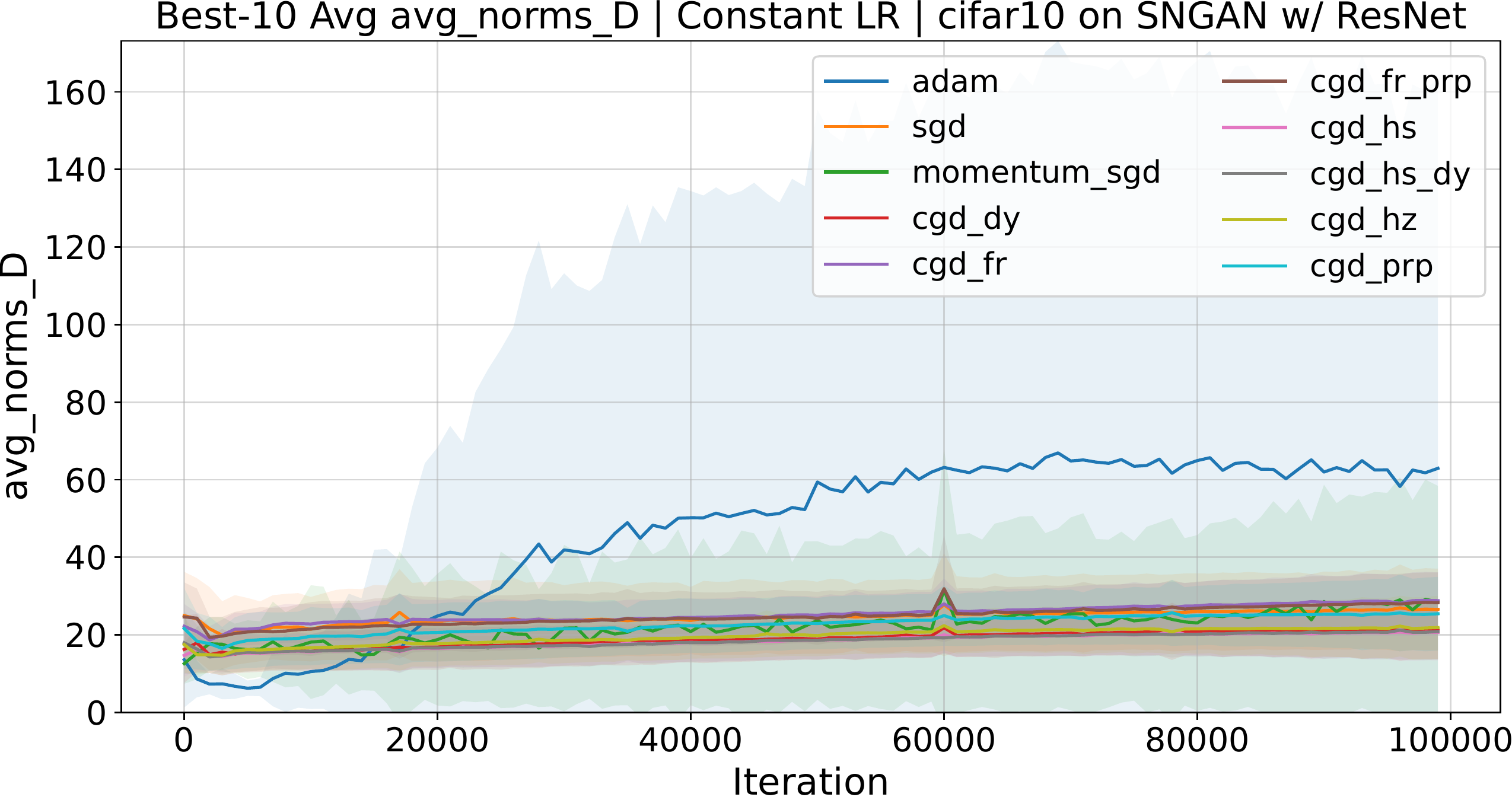}
 \includegraphics[width=0.48\linewidth]{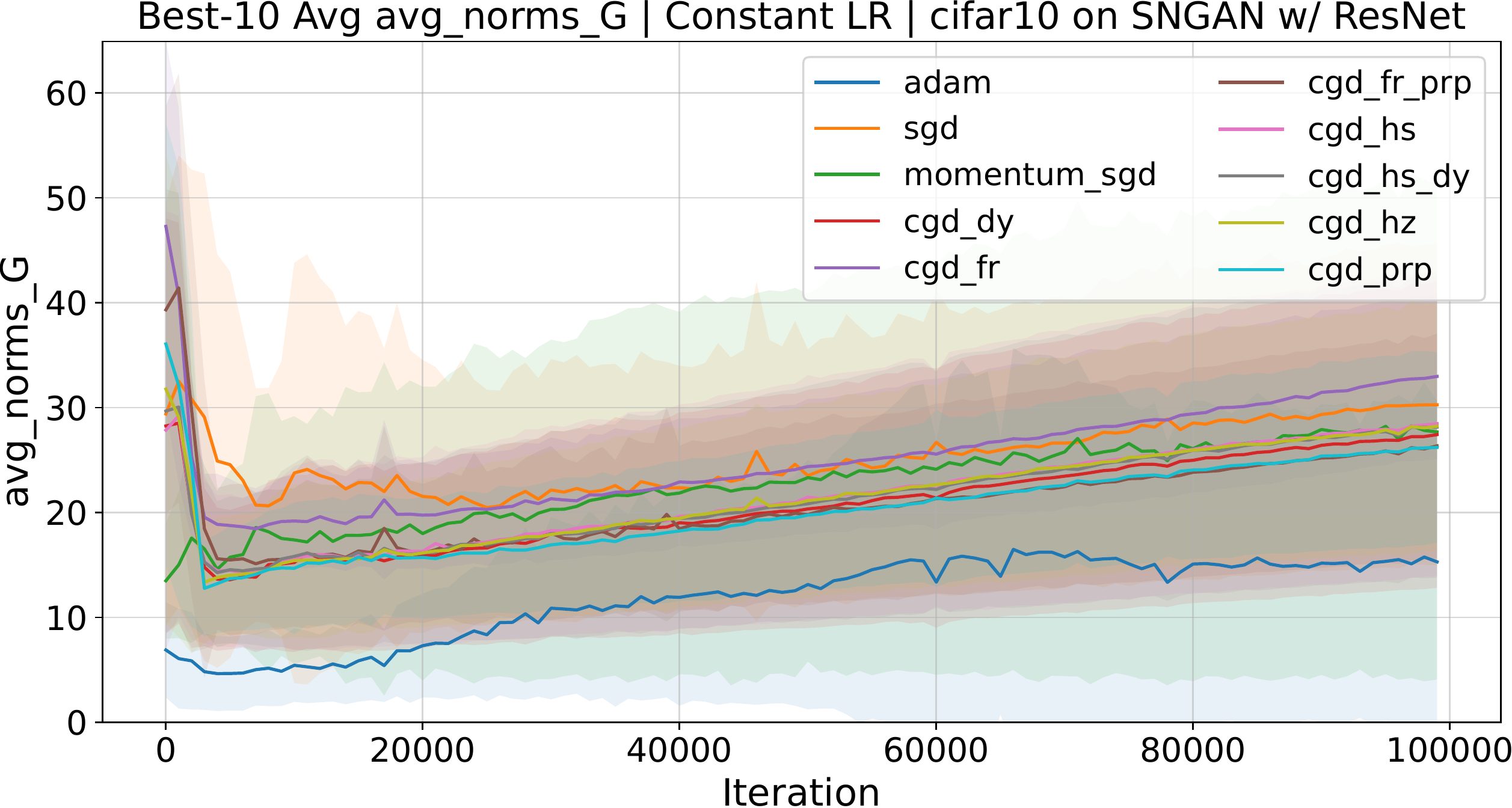}
\caption{Mean norm of the gradient (solid line surrounded by the shaded areas) bounded by the maximum and minimum over the best ten runs in the sense of FID. \textbf{Left}: CIFAR-10 discriminator, \textbf{Right}: CIFAR-10 generator}
\label{appendix_fig:res-exp-const-norm}
\vspace{-3mm}
\end{figure}

\subsubsection{Diminishing learning rate rule}
\label{appendix:resnet-diminishing-experiments}
{
The inverse square learning rate was not used in the previous studies \cite{Heusel2017, DBLP:journals/corr/Goodfellow17, radford2015unsupervised} because of the rapid decay of the learning rate from the beginning to the middle of the learning process, which is challenging to balance in minimax optimization. Therefore, we employed a scheduling scheme that decays the learning rate every $T$ steps.
}

{
The details of the FID scores for each learning-rate combination are shown in Figures \ref{additional_fig:res-lr_analysis_diminishing_cifar10_bs64} in Appendix \ref{appendix:lr-sensitivity}. The FID scores for training with the diminishing learning rate are shown in Figure \ref{additional_fig:resnet-exp-diminishing-fid}. The losses and norms of each generator and discriminator for training on the MNIST and CIFAR10 are shown in Figure \ref{additional_fig:res-exp-diminishing-loss} and Figure \ref{additional_fig:res-exp-diminishing-norm}, respectively. 
}

\begin{figure}[H]
 \centering
 \includegraphics[width=\linewidth]{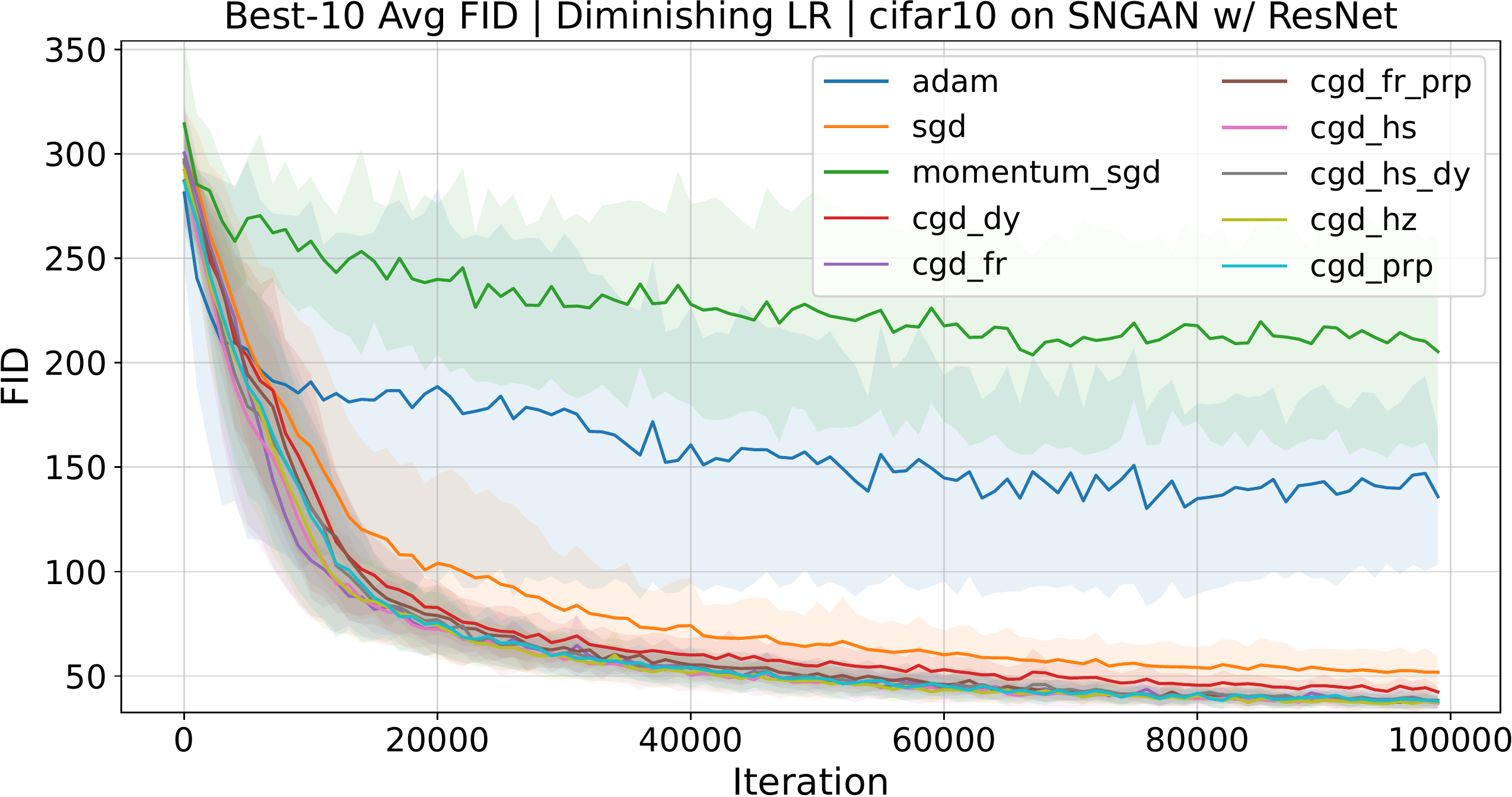}
\caption{Mean FID (solid line) bounded by the maximum and minimum (the shaded areas) over the best ten runs in the sense of FID when using a diminishing learning rate. Results for CIFAR-10 on SNGAN with ResNet generator.}
\label{additional_fig:resnet-exp-diminishing-fid}
\vspace{-3mm}
\end{figure}

\begin{figure}[H]
 \centering
 \includegraphics[width=0.48\linewidth]{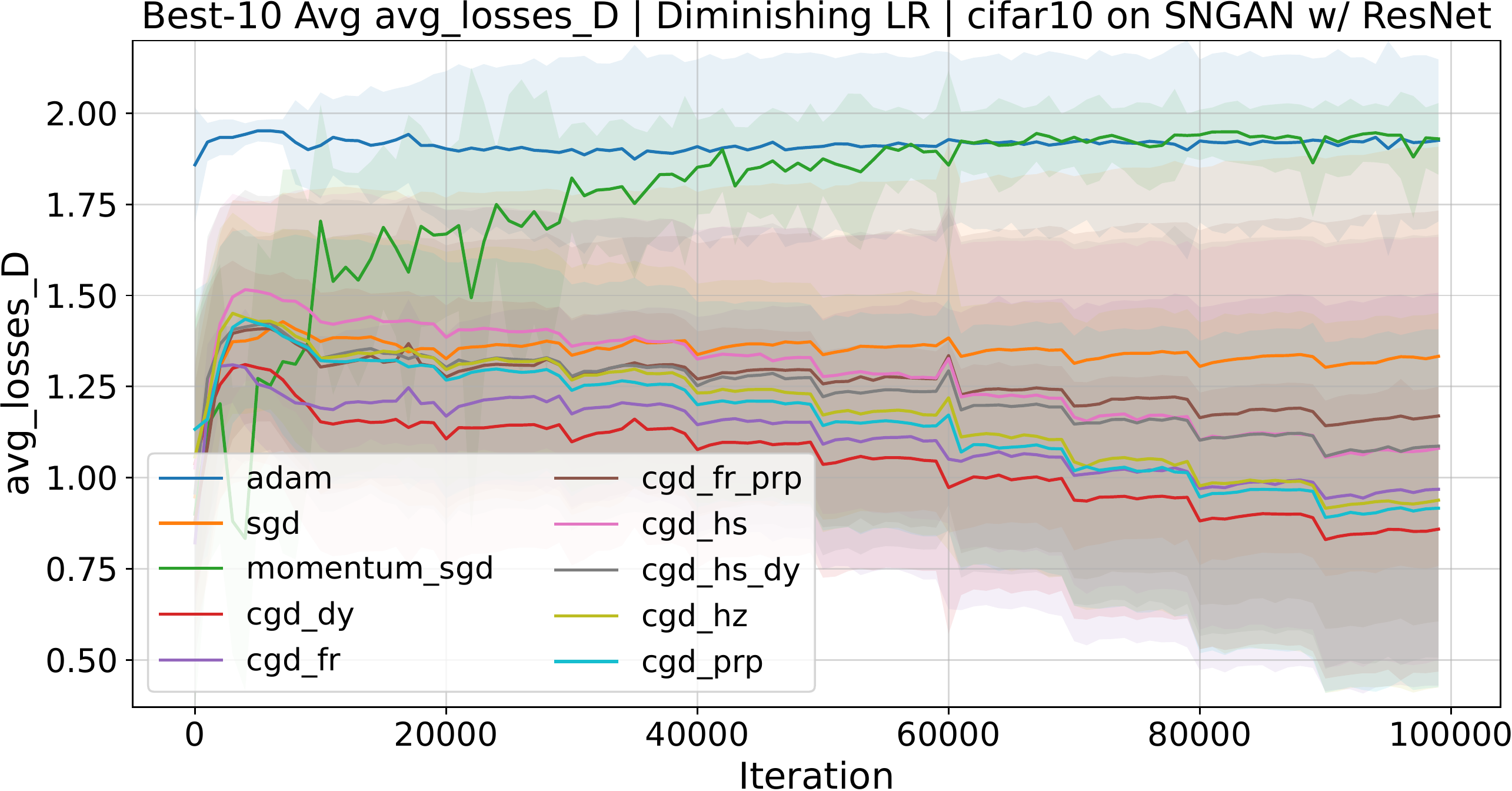}
 \includegraphics[width=0.48\linewidth]{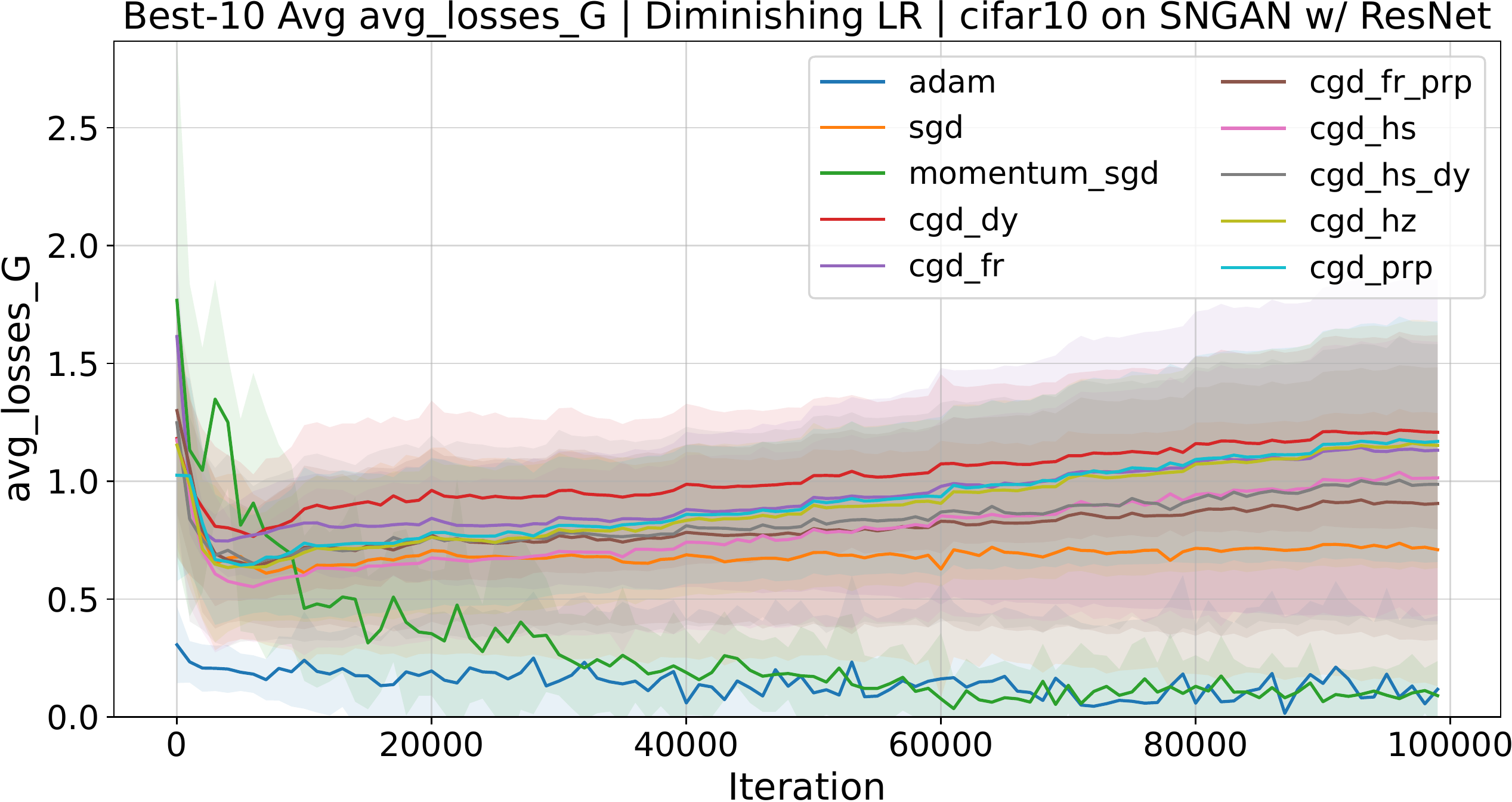}
\caption{Mean loss (solid line surrounded by the shaded areas) bounded by the maximum and minimum over the best ten runs in the sense of FID. \textbf{Left}: CIFAR-10 discriminator loss, \textbf{Right}: CIFAR-10 generator loss}
\label{additional_fig:res-exp-diminishing-loss}
\vspace{-3mm}
\end{figure}

\begin{figure}[H]

 \centering
 \includegraphics[width=0.48\linewidth]{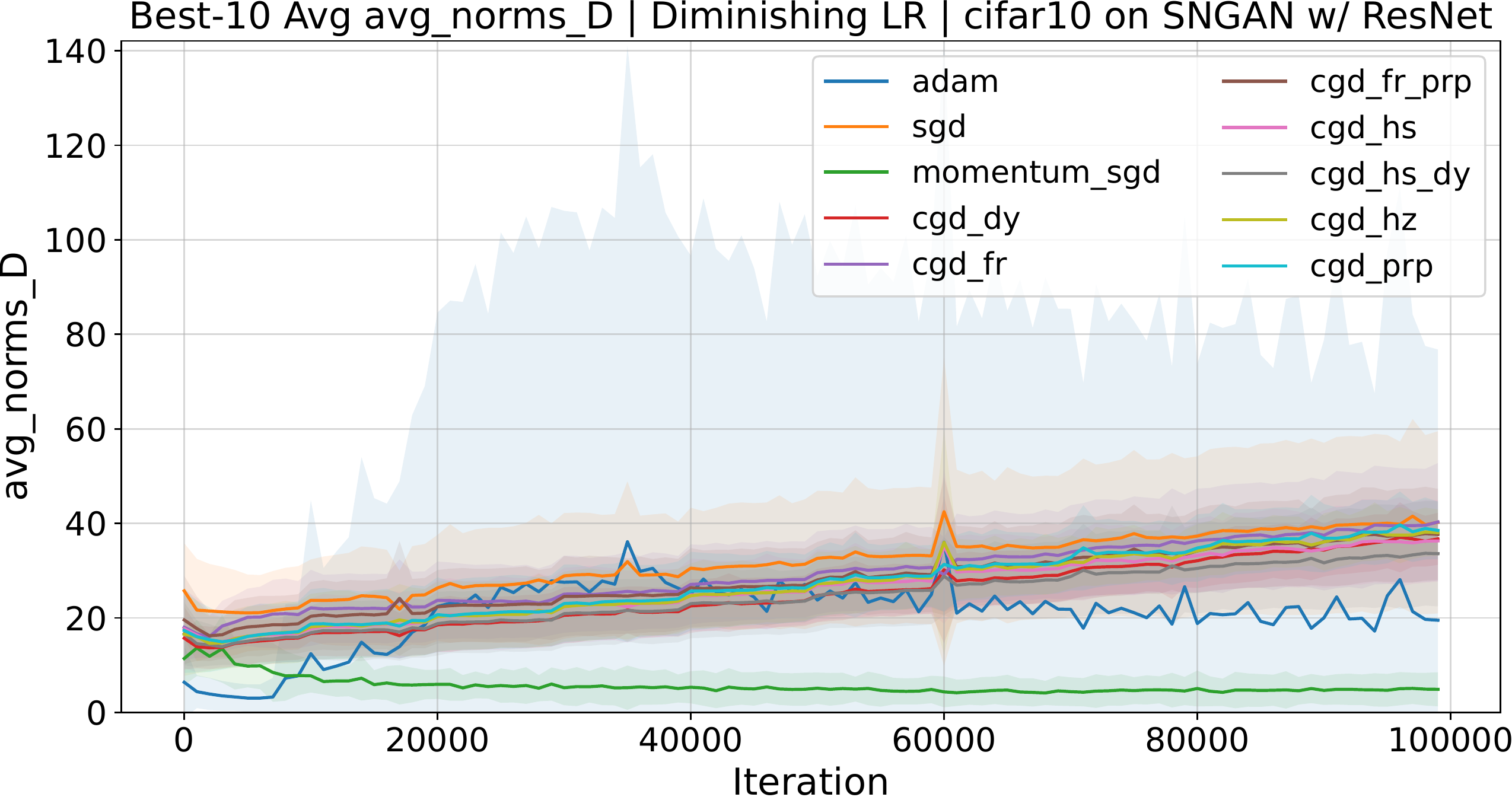}
 \includegraphics[width=0.48\linewidth]{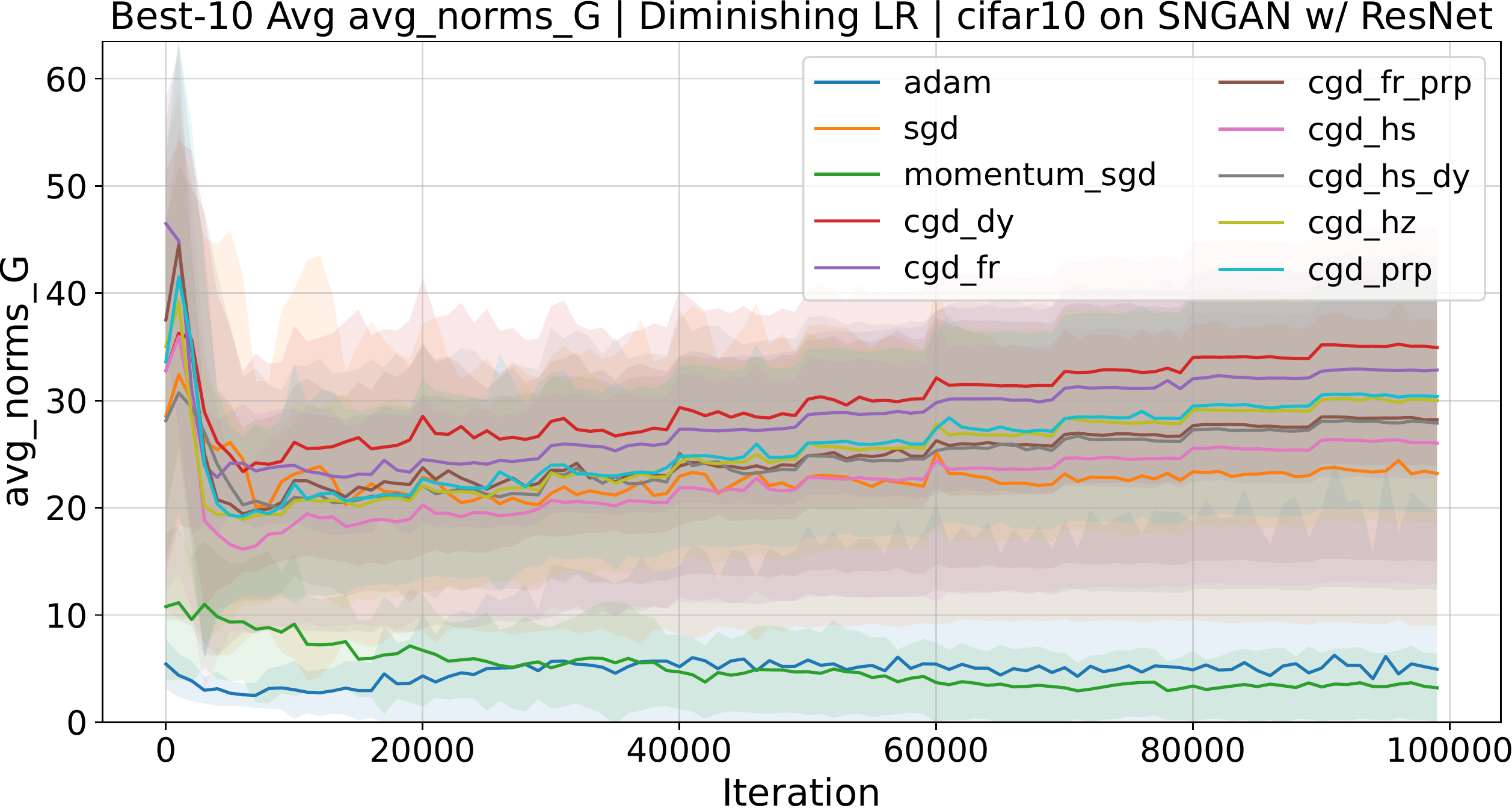}
\caption{Mean norm of the gradient (solid line surrounded by the shaded areas) bounded by the maximum and minimum over the best ten runs in the sense of FID. \textbf{Left}: CIFAR-10 discriminator, \textbf{Right}: CIFAR-10 generator}
\label{additional_fig:res-exp-diminishing-norm}
\vspace{-3mm}
\end{figure}

\clearpage
\newpage
\section{Sensitivity of learning rate}
\label{appendix:lr-sensitivity}
\label{appendix:resnet-lr-sensitivity}
\vspace{-3mm}

\subsection{Constant learning rate rule}
\vspace{-3mm}
{
For the dataset $\times$ optimizer combinations, we performed a grid search of 25 hyperparameter combinations, including five combinations each for LR\_G (initial learning rate for the generator) and LR\_D (initial learning rate for the discriminator). To evaluate the optimizer's performance on LR\_G and LR\_D, we created a heat map of FID scores.
}

\begin{figure}[ht]
 \centering
\includegraphics[width=0.32\linewidth]{figs/v3/grid/cifar10/64-adam.pdf}
 \includegraphics[width=0.32\linewidth]{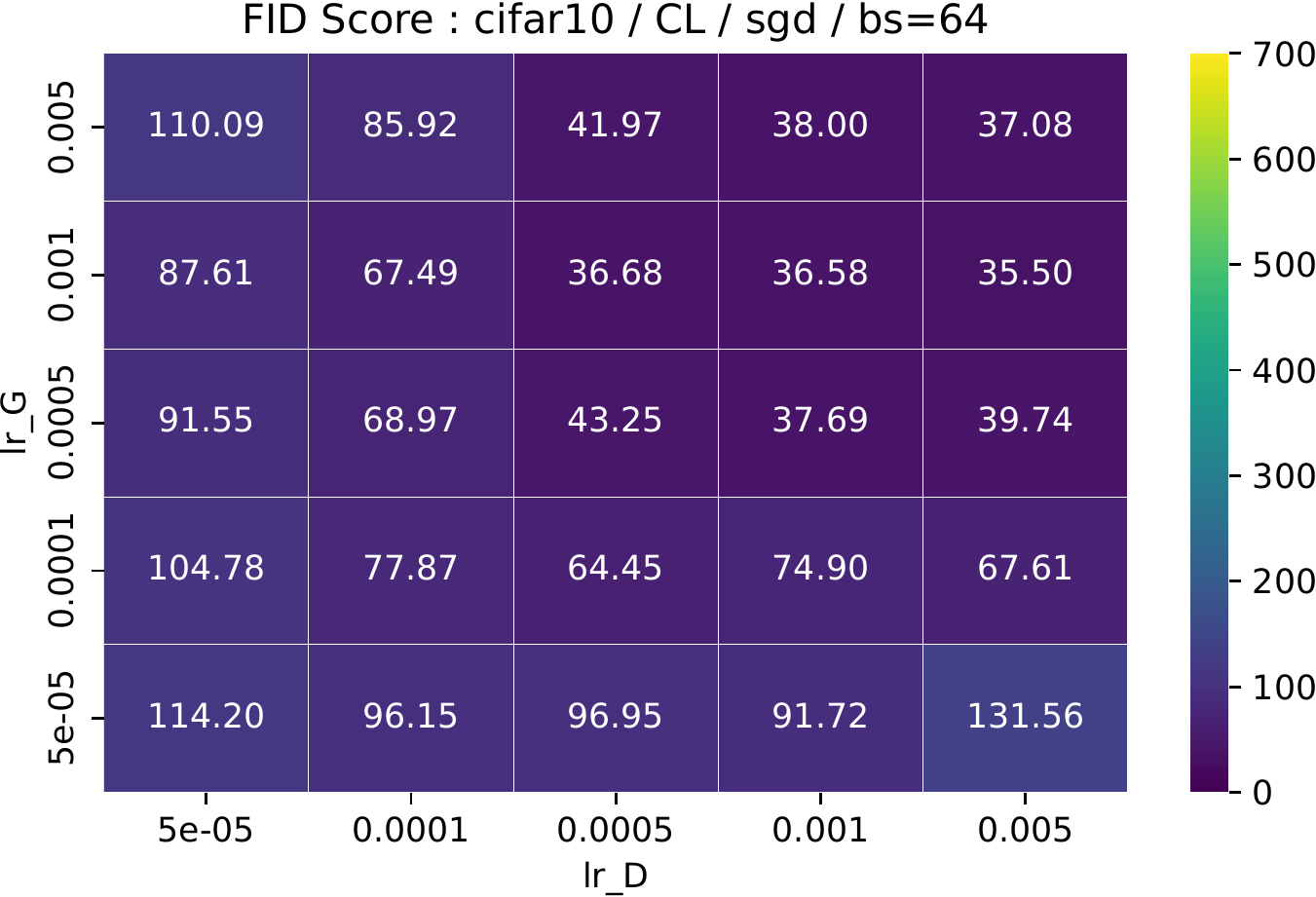}
 \includegraphics[width=0.32\linewidth]{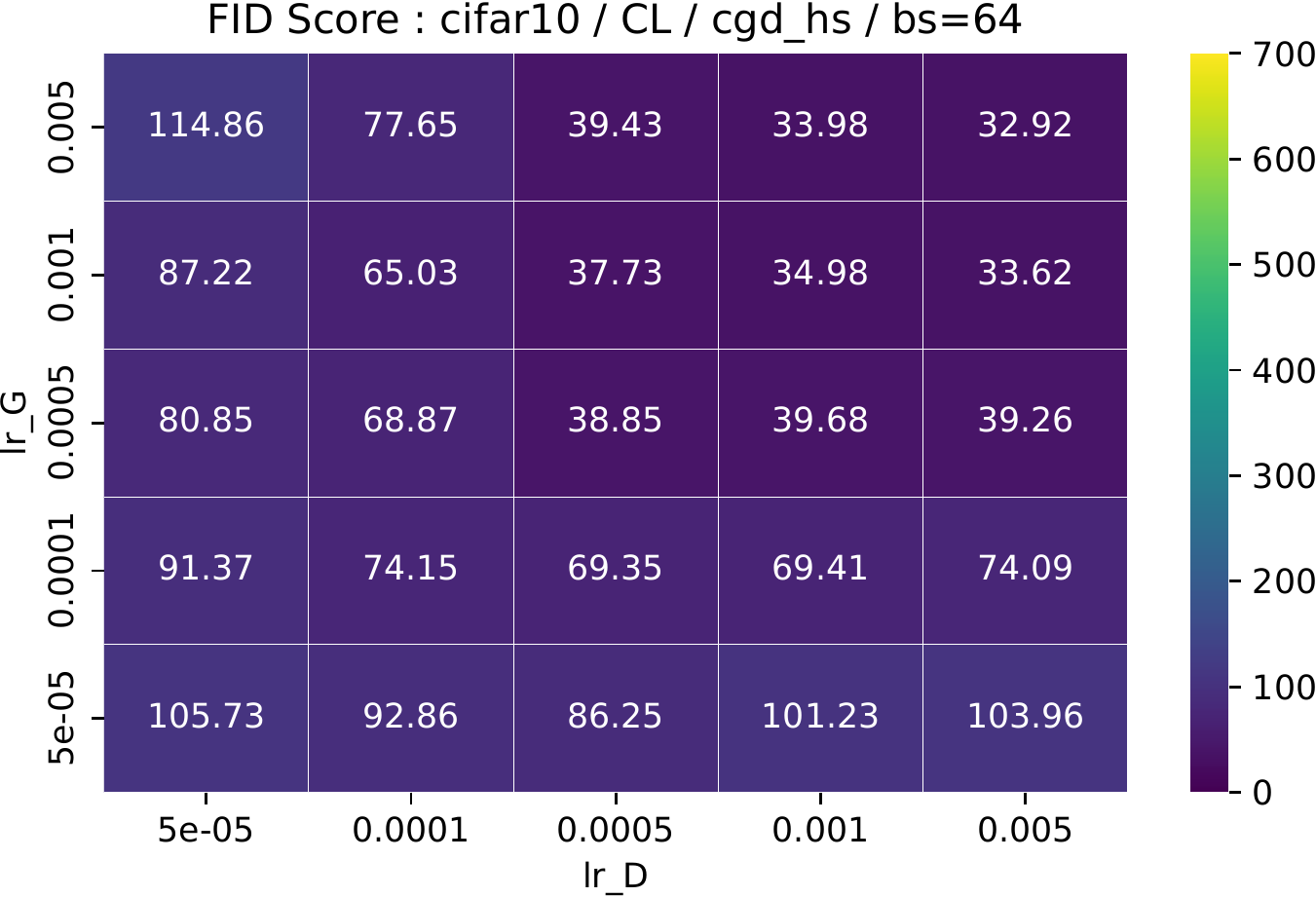}
 \includegraphics[width=0.32\linewidth]{figs/v3/grid/cifar10/64-cgd_fr.pdf}
 \includegraphics[width=0.32\linewidth]{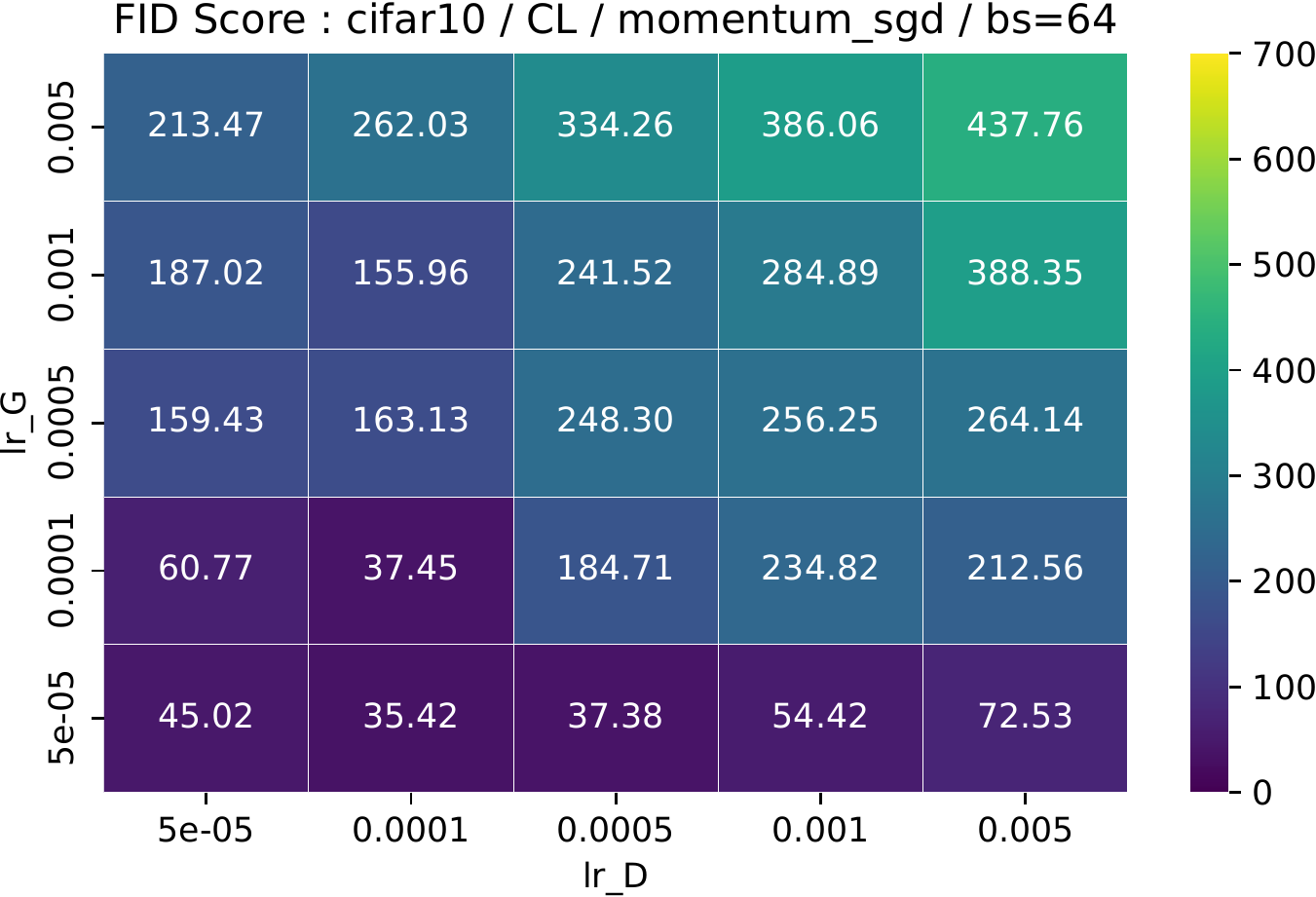}
 \includegraphics[width=0.32\linewidth]{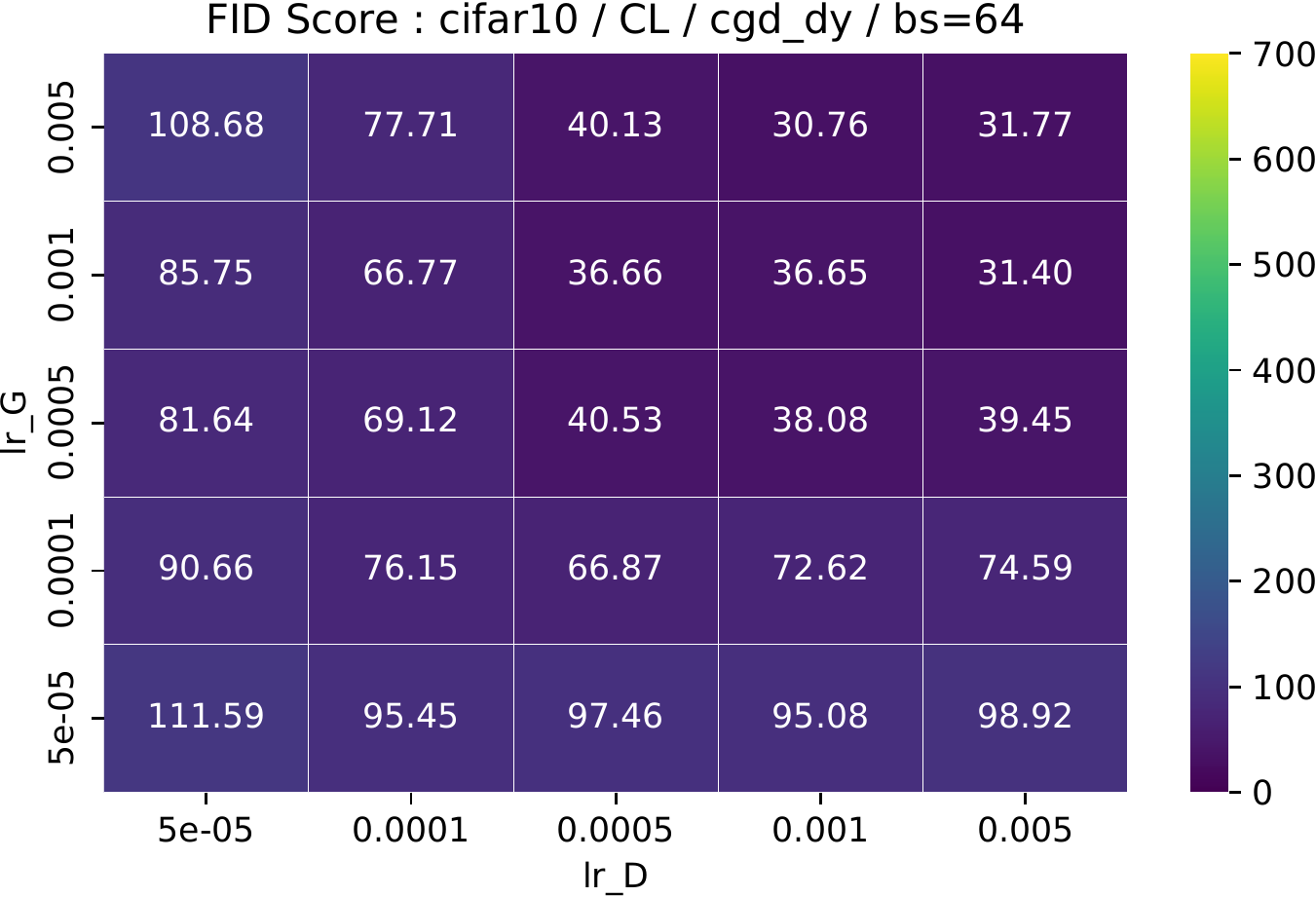}
 \includegraphics[width=0.32\linewidth]{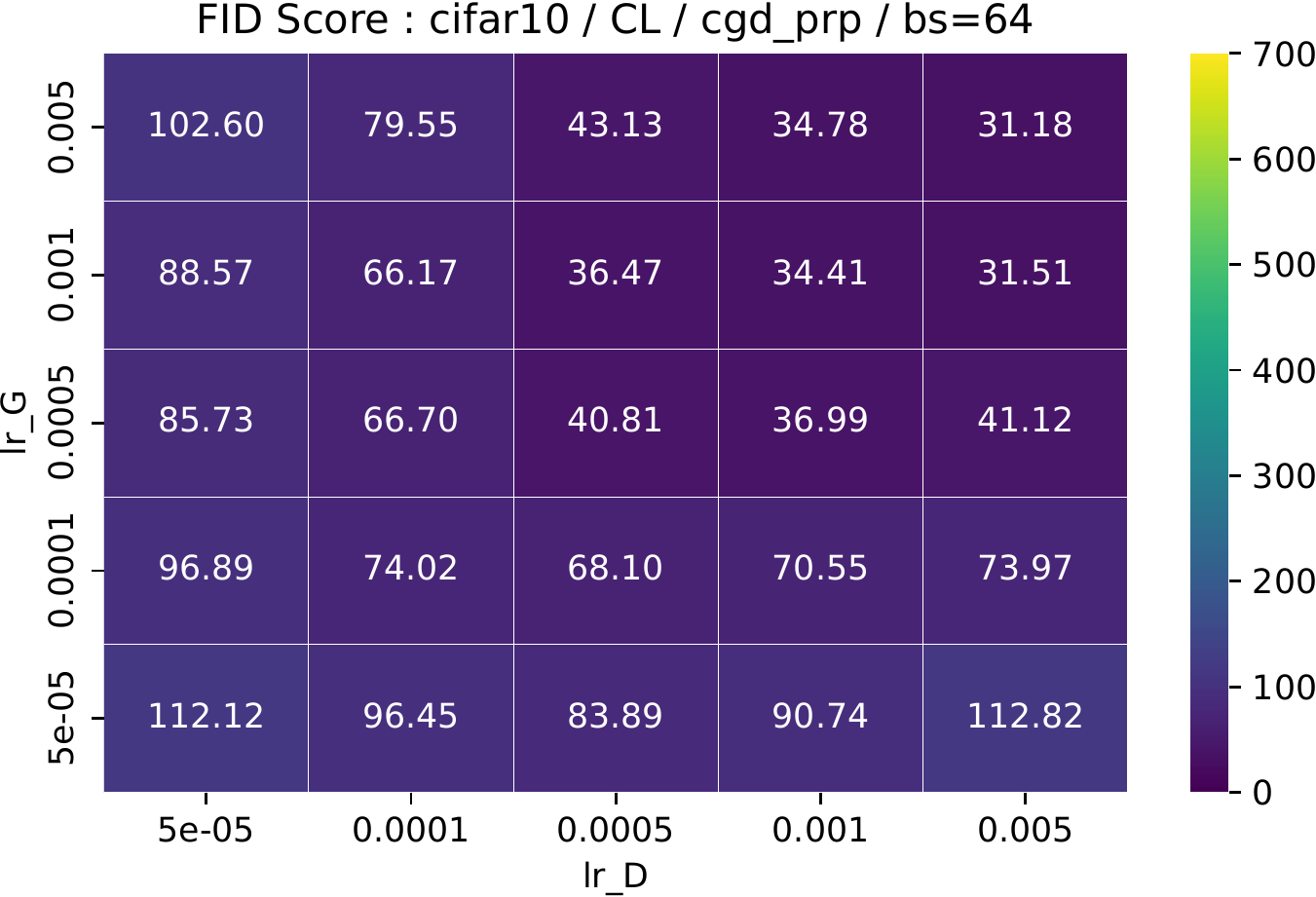}
 \includegraphics[width=0.32\linewidth]{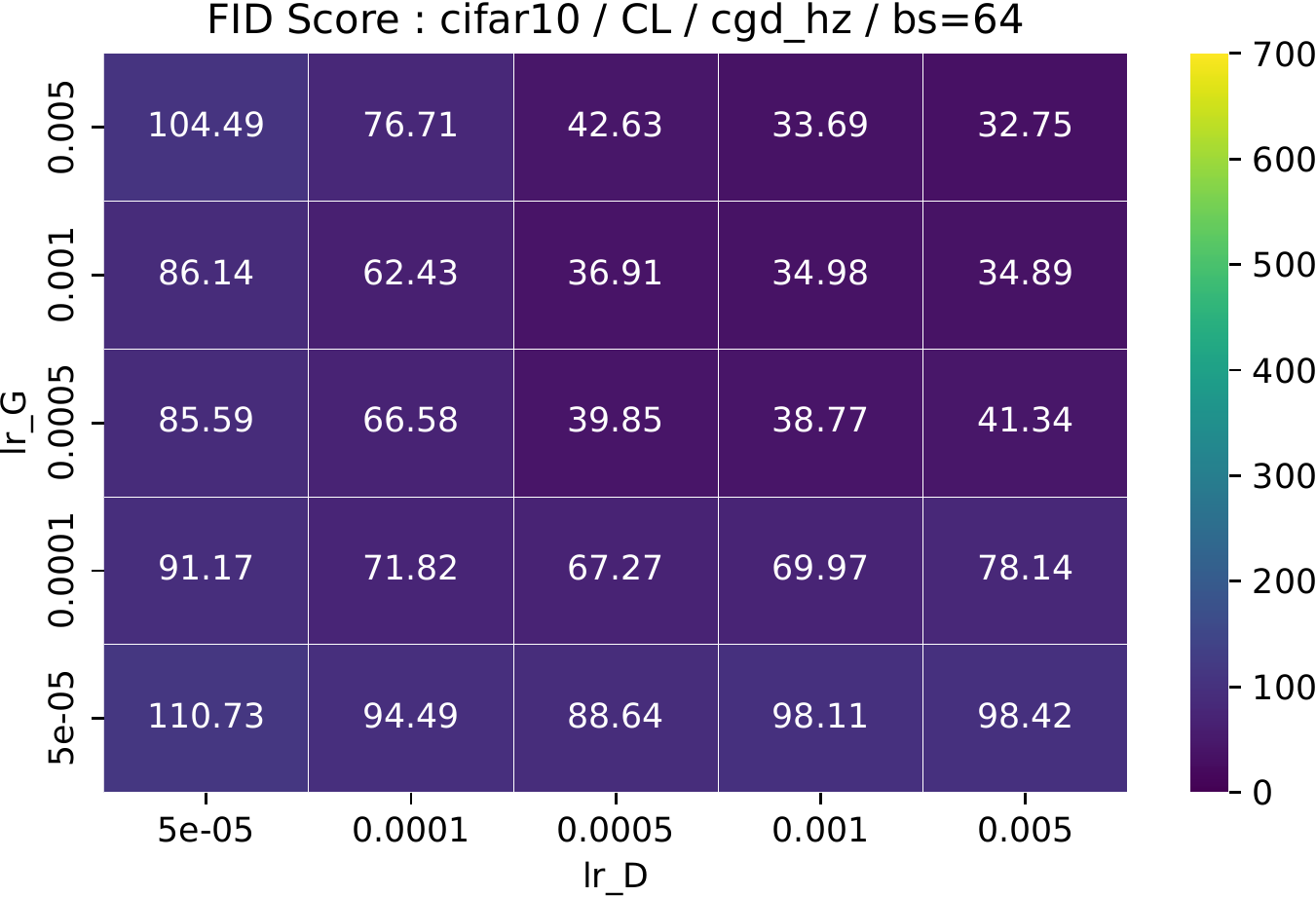}
 \includegraphics[width=0.32\linewidth]{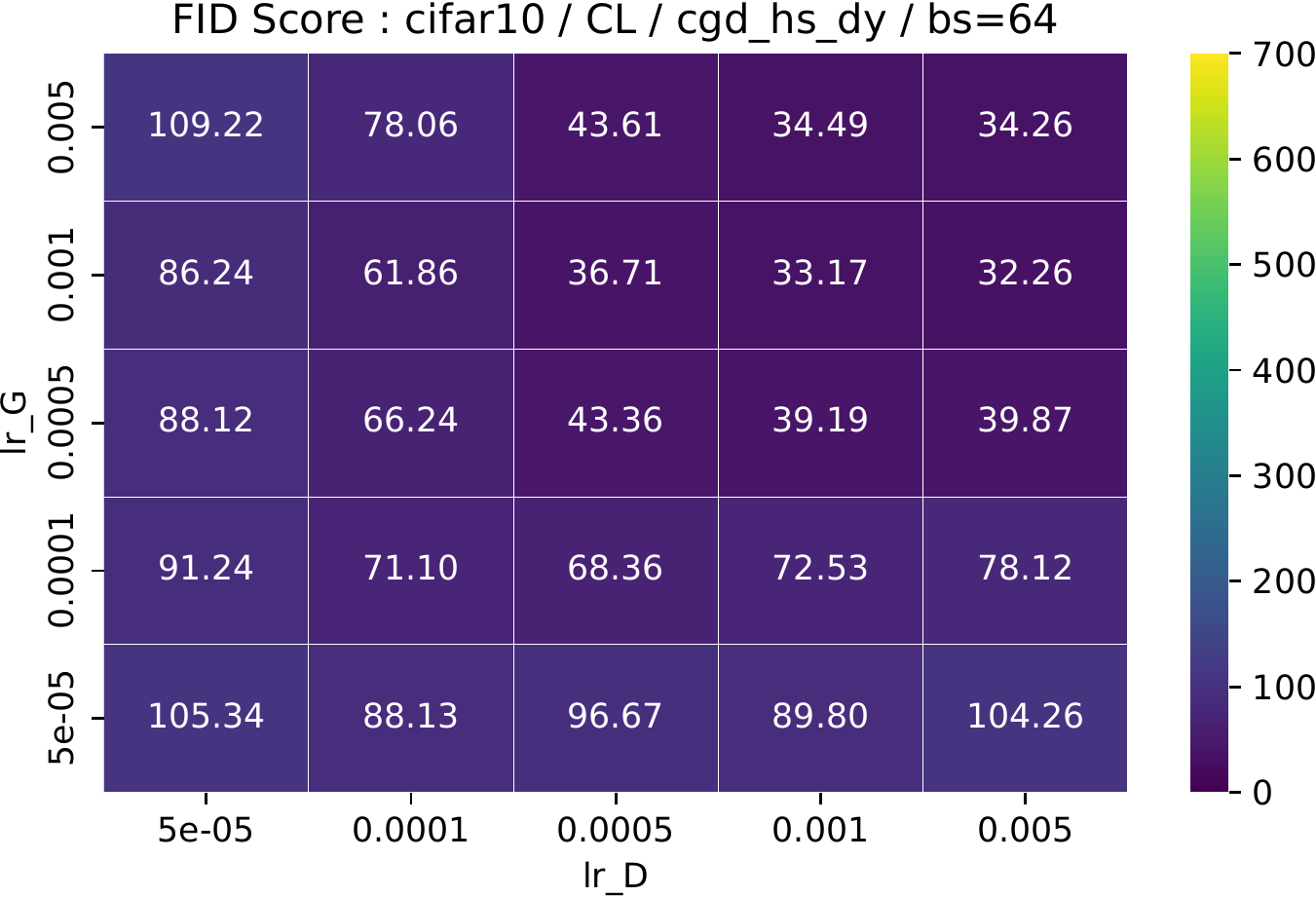}
 \includegraphics[width=0.32\linewidth]{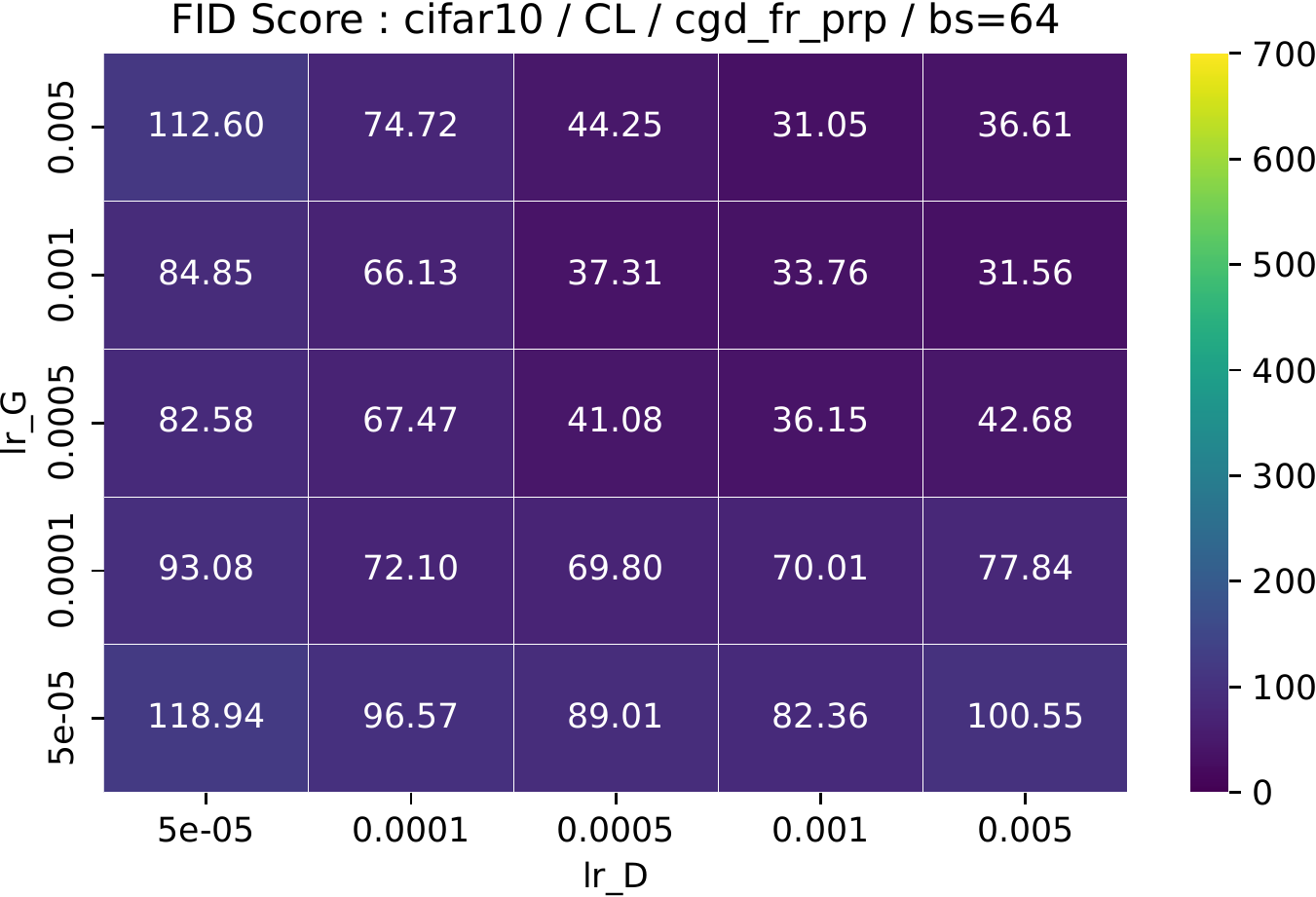}
 \caption{Analysis of the dependence of FID value on LR (CIFAR10, constant learning rate, batch size = 64): LR of the generator on the vertical axis and LR of discriminator on the horizontal axis. The heatmap colors denote the FID scores: the darker the blue, the lower the FID, meaning that the training of the generator succeeded.}
\label{additional_fig:res-lr_analysis_constlr_cifar10_bs64}
\vspace{-3mm}
\end{figure}

\newpage
\subsection{Diminishing learning rate rule}
\vspace{-3mm}
{
For dataset $\times$ optimizer combinations, we performed a grid search of 16 hyperparameter combinations, including five combinations each for LR\_G (initial learning rate for generator) and LR\_D (initial learning rate for discriminator). To evaluate the optimizer's performance on LR\_G and LR\_D, we created a heat map of FID scores.
}

\begin{figure}[ht]
 \centering
\includegraphics[width=0.32\linewidth]{figs/v4/grid/cifar10/64-adam.pdf}
 \includegraphics[width=0.32\linewidth]{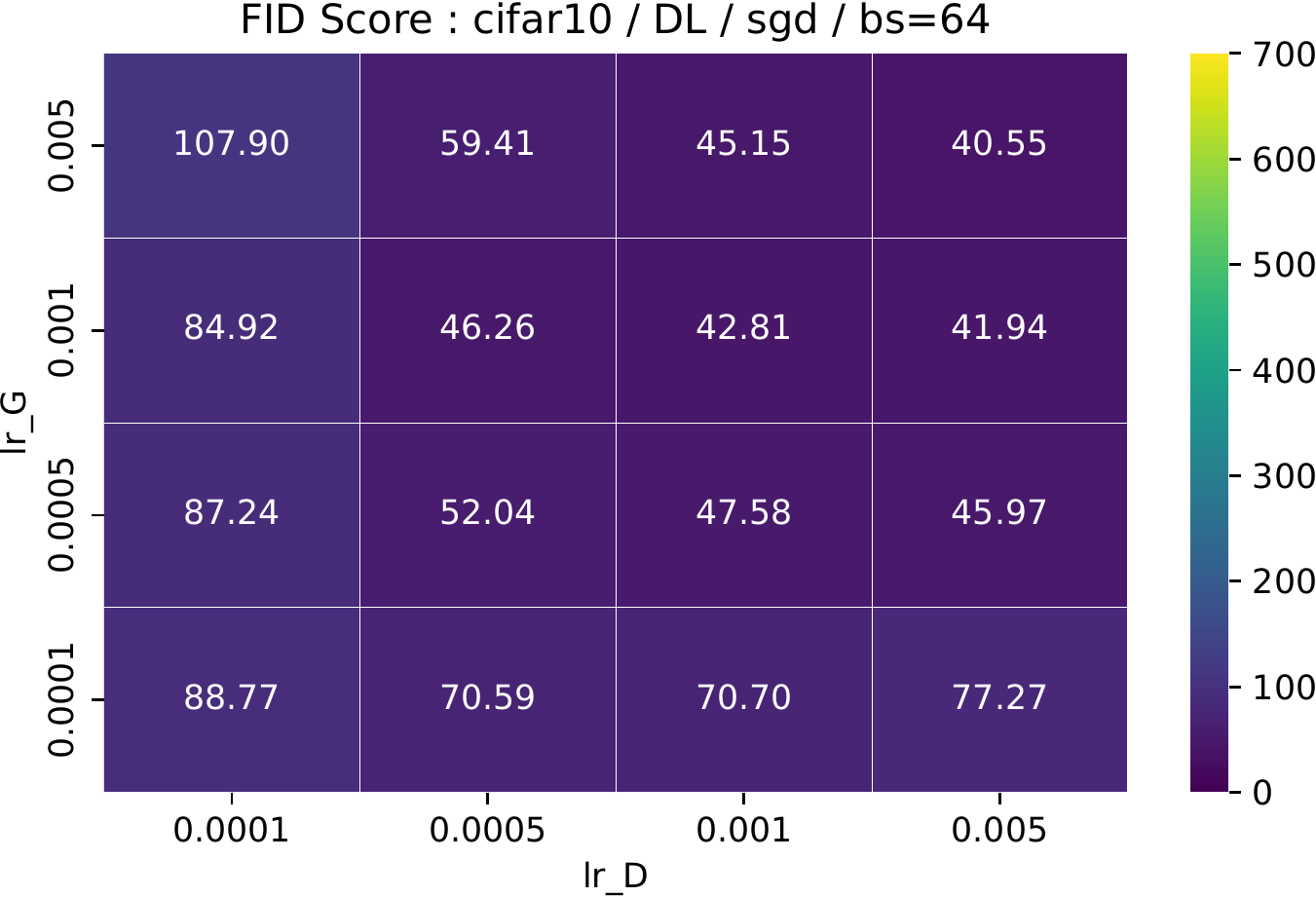}
 \includegraphics[width=0.32\linewidth]{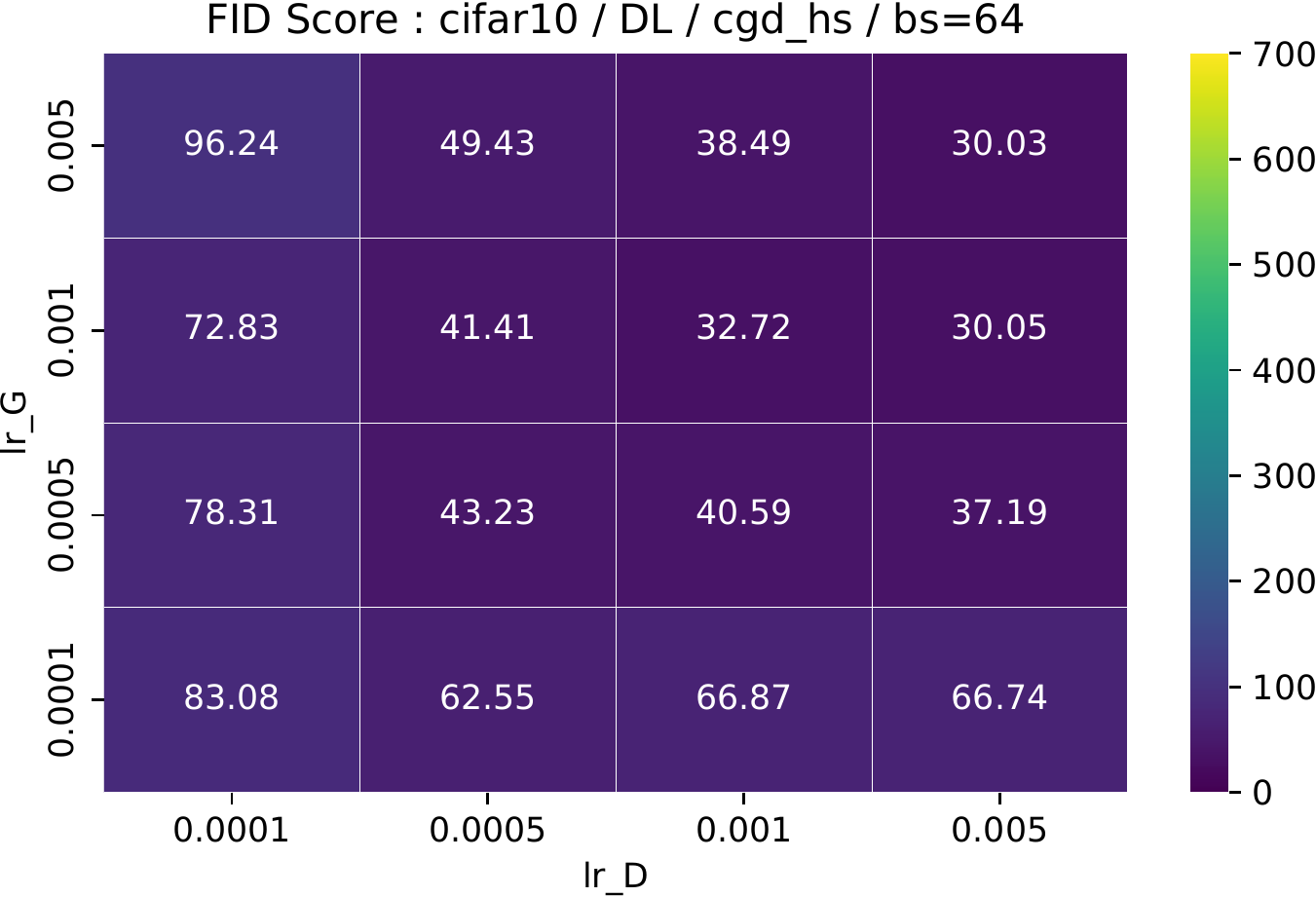}
 \includegraphics[width=0.32\linewidth]{figs/v4/grid/cifar10/64-cgd_fr.pdf}
 \includegraphics[width=0.32\linewidth]{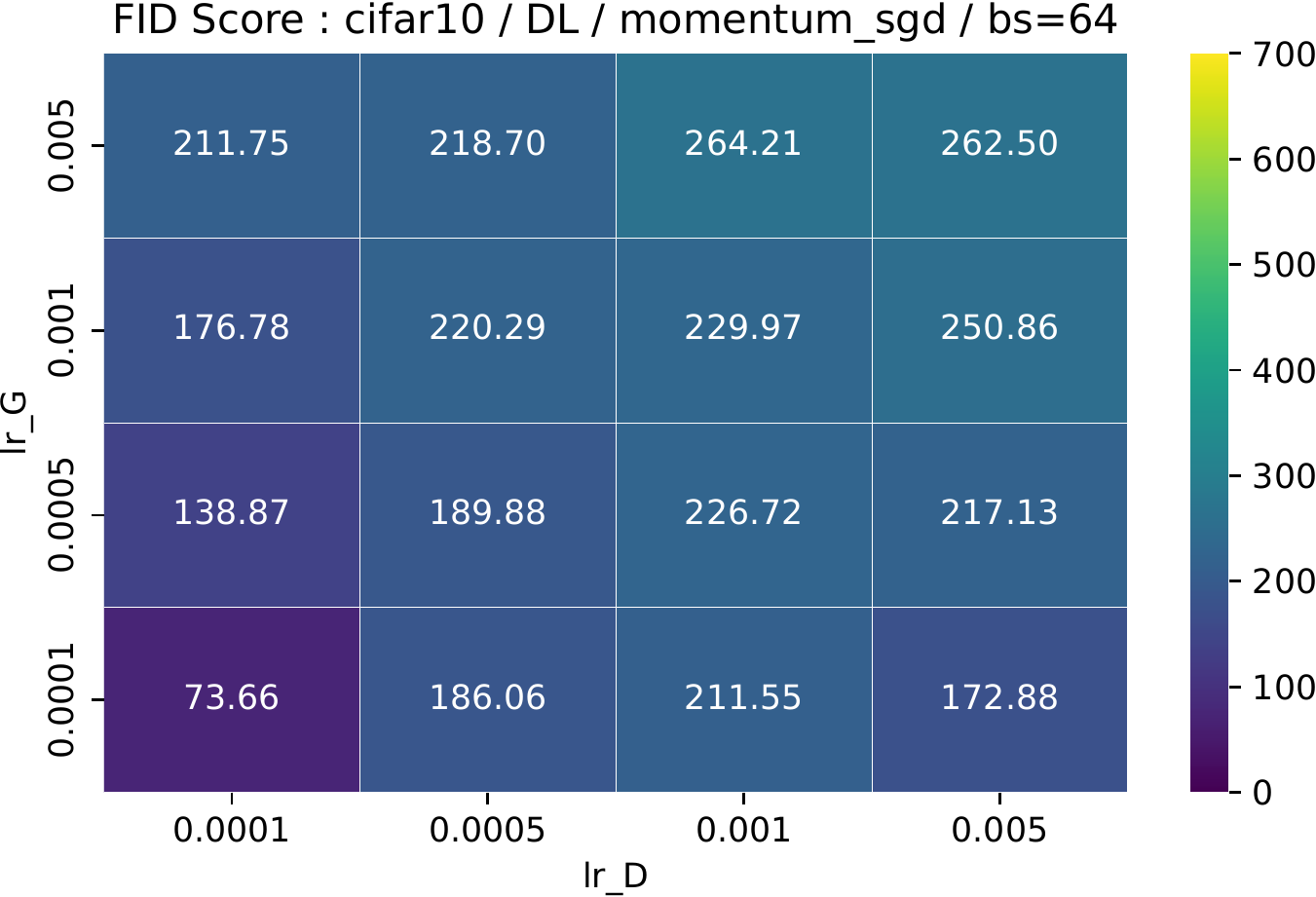}
 \includegraphics[width=0.32\linewidth]{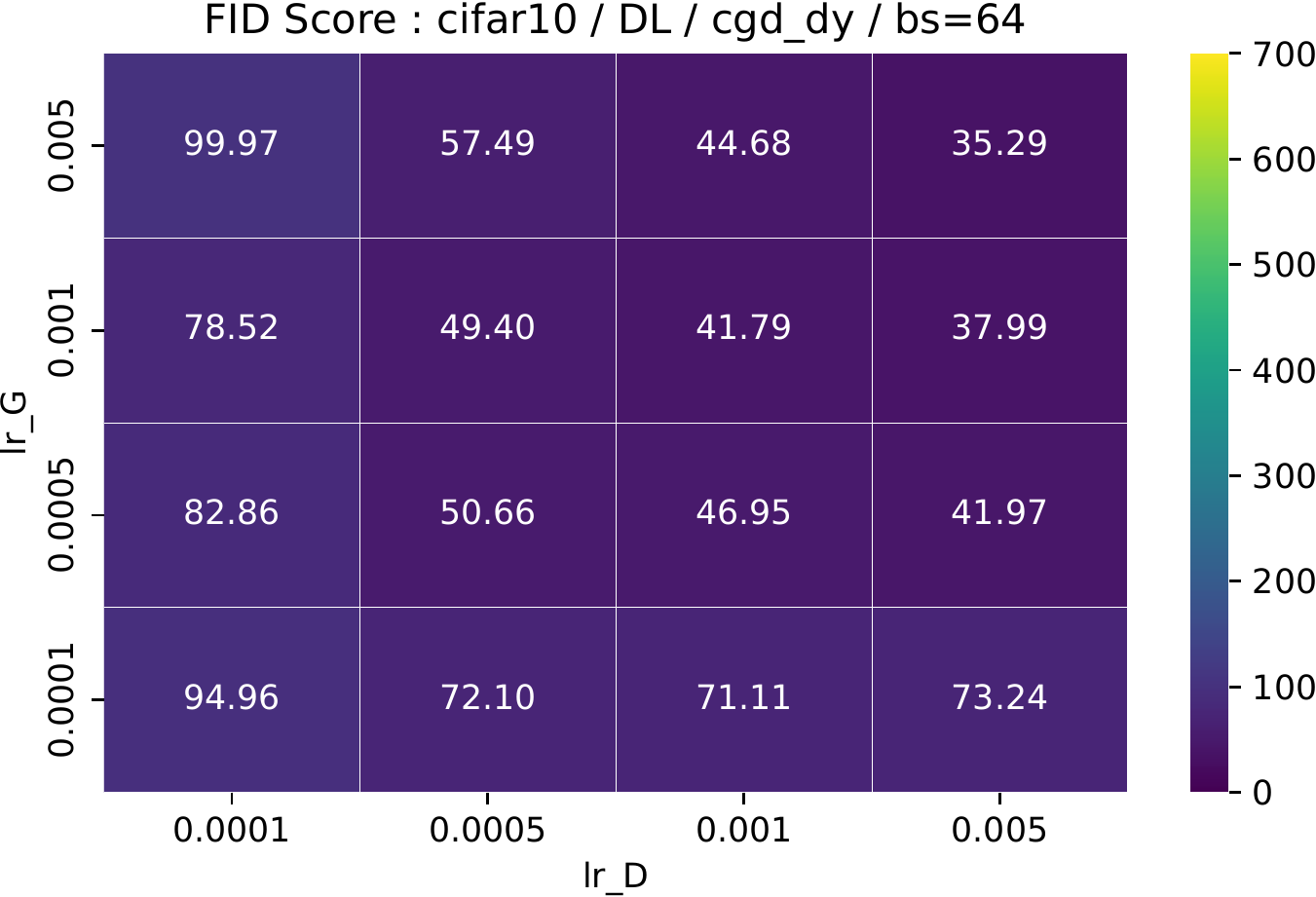}
 \includegraphics[width=0.32\linewidth]{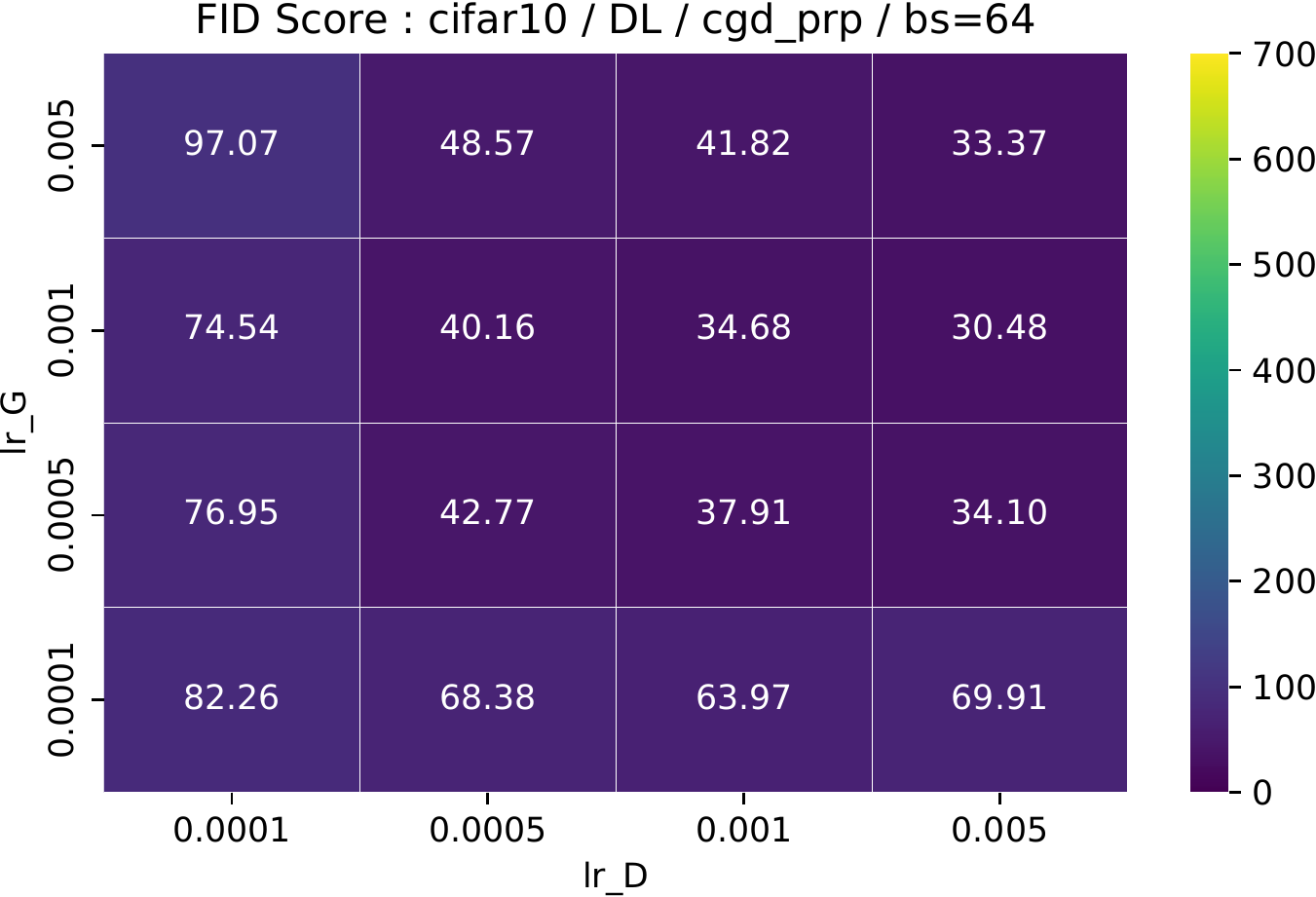}
 \includegraphics[width=0.32\linewidth]{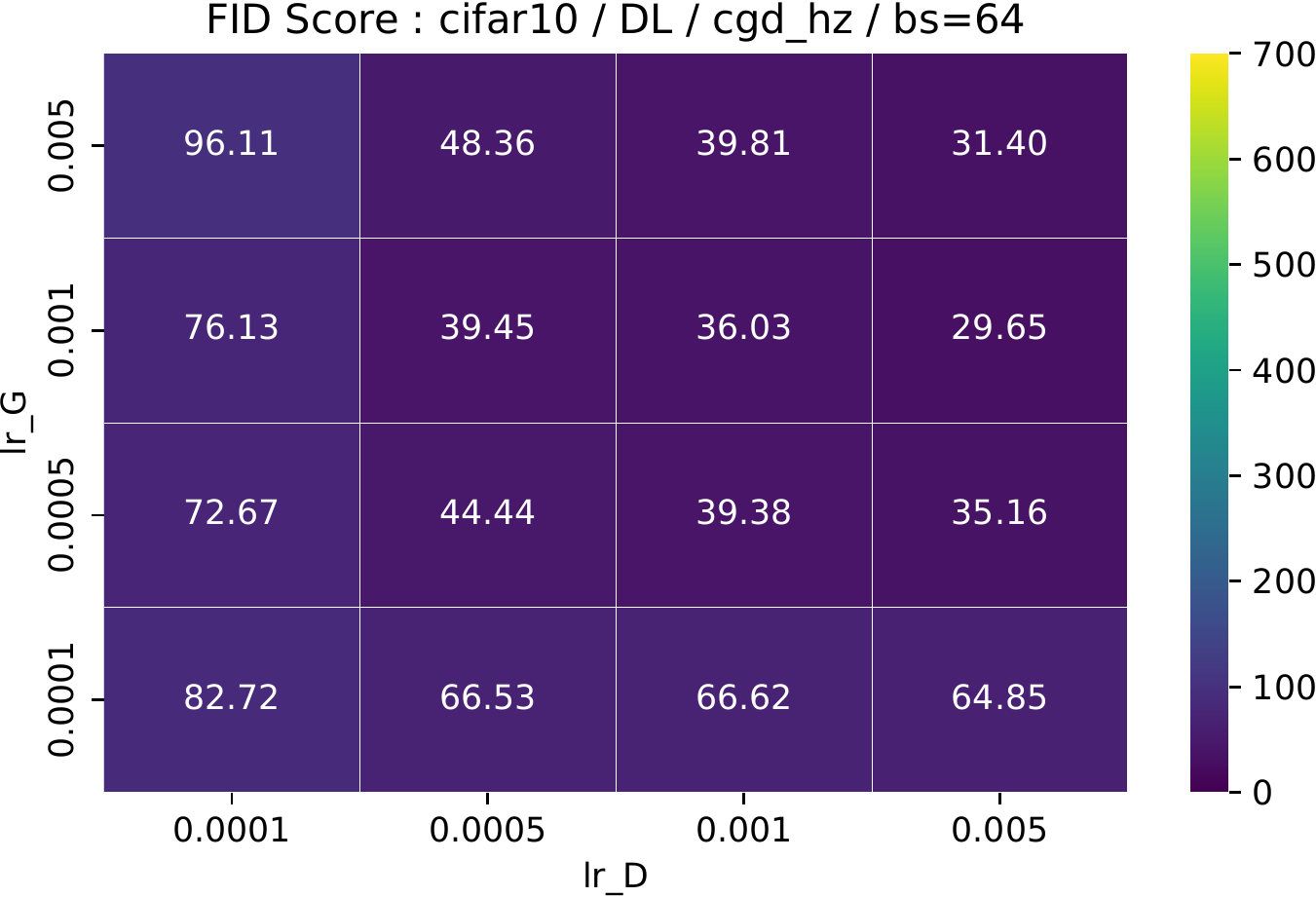}
 \includegraphics[width=0.32\linewidth]{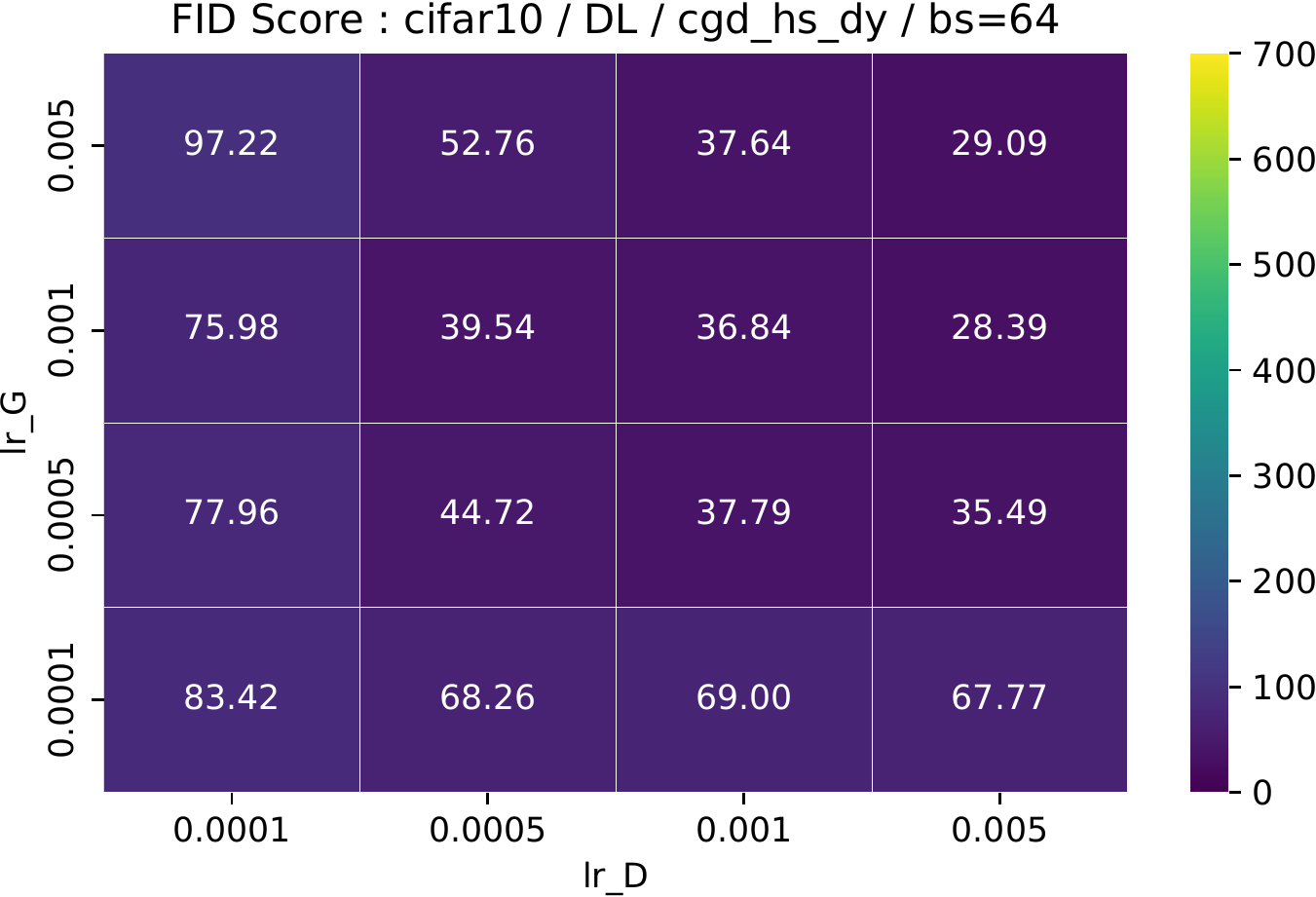}
 \includegraphics[width=0.32\linewidth]{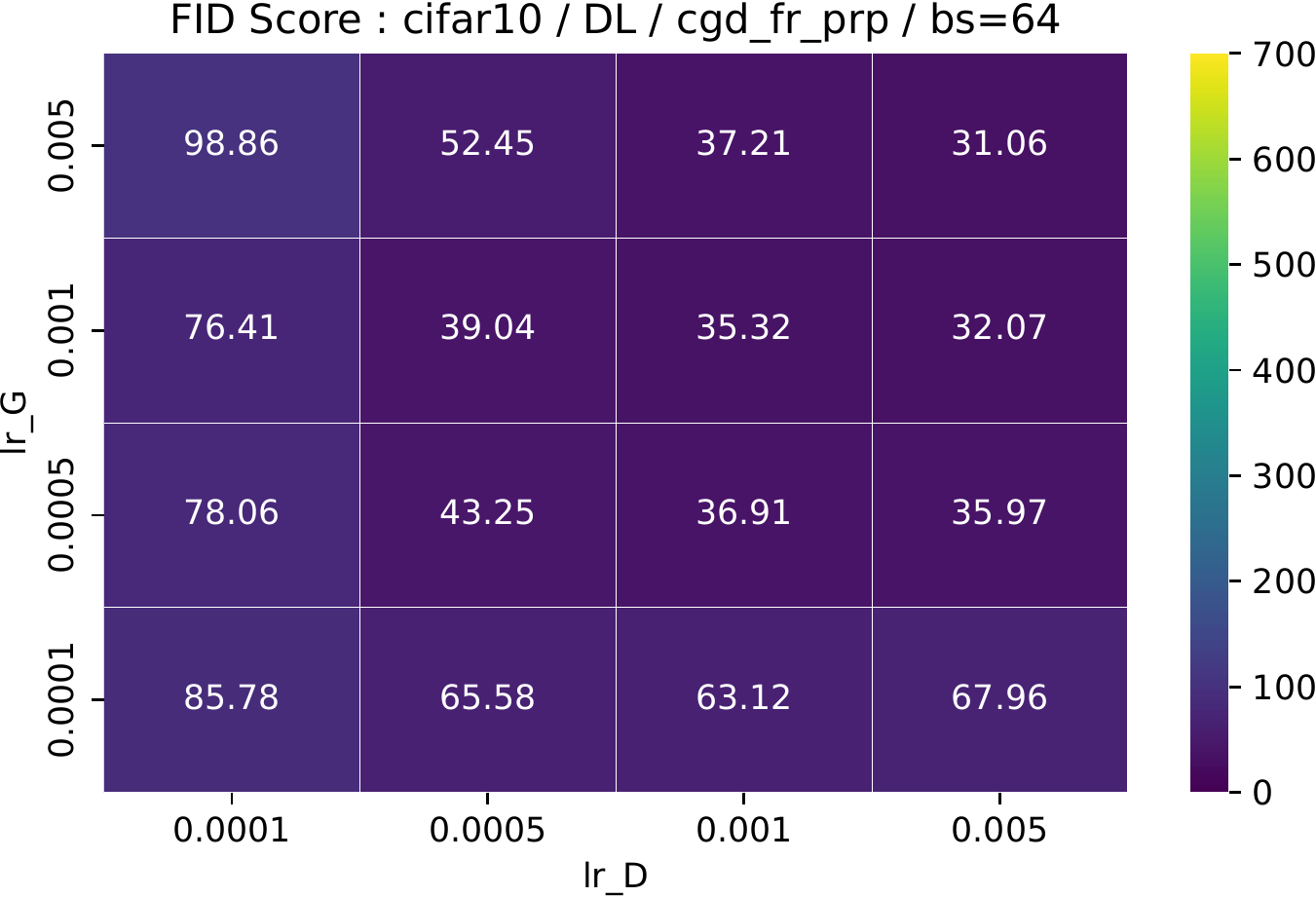}
 \caption{Analysis of the dependence of FID value on LR (CIFAR10, diminishing learning rate, batch size = 64): LR of the generator on the vertical axis and LR of discriminator on the horizontal axis. The heatmap colors denote the FID scores: the darker the blue, the lower the FID, meaning that the training of the generator succeeded.}
\label{additional_fig:res-lr_analysis_diminishing_cifar10_bs64}
\vspace{-3mm}
\end{figure}

\end{appendices}

\end{document}